\documentclass[10pt,journal,compsoc]{IEEEtran}

\ifCLASSOPTIONcompsoc
  \usepackage[nocompress]{cite}
\else
  \usepackage{cite}
\fi

\ifCLASSINFOpdf
\else
\fi
\usepackage{amssymb,amsmath,amsthm,enumerate,threeparttable,bm,hyperref}
 \usepackage{caption}

\usepackage{subfigure}
\usepackage{algorithm}  

\usepackage{algpseudocode}
\usepackage{booktabs}
\usepackage{multirow}
\usepackage{colortbl}
\usepackage{threeparttable}
\newtheorem{lemma}{Lemma}

\newtheorem{theorem}{Theorem}
\newtheorem{definition}{Definition}
\newtheorem{proposition}{Proposition}
\newtheorem{remark}{Remark}
\newtheorem{example}{Example}

\usepackage{color}

\usepackage{lipsum}
\usepackage{enumitem}
\usepackage{xcolor}

\def\blue#1{\textcolor{blue}{#1}}
\long\def\comment#1{}
\usepackage{setspace}
\usepackage{amsmath}
\usepackage{wrapfig}
\usepackage[export]{adjustbox}
\usepackage[switch]{lineno}

\usepackage{array}

\usepackage{amsmath,amsfonts,bm}

\def\eqref#1{equation~\ref{#1}}

\def\1{\bm{1}}

\def\vtheta{{\bm{\theta}}}

\def\vr{{\bm{r}}}

\def\vt{{\bm{t}}}

\def\vv{{\bm{v}}}
\def\vw{{\bm{w}}}
\def\vx{{\bm{x}}}

\def\mG{{\bm{G}}}

\DeclareMathAlphabet{\mathsfit}{\encodingdefault}{\sfdefault}{m}{sl}
\SetMathAlphabet{\mathsfit}{bold}{\encodingdefault}{\sfdefault}{bx}{n}

\def\gD{{\mathcal{D}}}

\def\gF{{\mathcal{F}}}

\def\gX{{\mathcal{X}}}
\def\gY{{\mathcal{Y}}}
\def\gZ{{\mathcal{Z}}}

\DeclareMathOperator*{\argmax}{arg\,max}

\DeclareMathOperator{\sign}{sign}

\def\ie{$i.e.$}
\def\eg{$e.g.$}

\def\blue#1{\textcolor{blue}{#1}}

\def\pcl{P_C}

\def\DRe{\gD_r}
\def\DPo{\gD_m}
\def\DWa{\gD_w}
\def\vdelta{\bm{\delta}}
\def\pno{\mathcal{P}_N}
\def\vepsilon{\bm{\epsilon}}

\begin{document}

%

\title{CertDW: Towards Certified Dataset Ownership Verification via Conformal Calibration}

\author{Ting Qiao\textsuperscript{*},
        Yiming Li\textsuperscript{*},
        Jianbin Li,
        Yingjia Wang,
        Leyi Qi,
        Junfeng Guo,
        Ruili Feng,
        Dacheng Tao
\thanks{\textsuperscript{*}The first two authors contributed equally to this work.}
\thanks{This work was supported by Beijing Natural Science Foundation (No. L251061).}
\thanks{Ting~Qiao, Jianbin~Li and Yingjia~Wang are with School of Control and Computer Engineering, North China Electric Power University, Beijing,102206, China (e-mail: \href{mailto:qiaoting@ncepu.edu.cn}{\{qiaoting,lijb87,wyj\}@ncepu.edu.cn}).}
\thanks{Yiming Li and Dacheng Tao are with College of Computing and Data Science, Nanyang Technological University, Singapore, 639798, Singapore (e-mail: \href{mailto:liyiming.tech@gmail.com}{\{liyiming.tech, dacheng.tao\}@gmail.com}).}
\thanks{Leyi Qi is with the School of Cybersecurity, Northwestern Polytechnical University, Xi’an, 710072, China (e-mail: \href{mailto:Leyi-Qi@outlook.com}{Leyi-Qi@outlook.com}).}
\thanks{Junfeng Guo is with Department of Computer Science, University of Maryland, College Park, MD 20742, USA (e-mail: \href{mailto:gjf2023@umd.edu}{gjf2023@umd.edu}).}
\thanks{Ruili Feng is with Alibaba Group, Hangzhou, 311100, China (e-mail: \href{mailto:ruilifengustc@gmail.com}{ruilifengustc@gmail.com}).}
\thanks{Corresponding Author(s): Yiming Li and Jianbin Li.}
}

%
%

\markboth{IEEE Transactions on Pattern Analysis and Machine Intelligence}%
{IEEE Transactions on Pattern Analysis and Machine Intelligence}

\IEEEtitleabstractindextext{%
\begin{abstract}
Deep neural networks (DNNs) rely heavily on high-quality open-source datasets (\eg, ImageNet) for their success, making dataset ownership verification (DOV) crucial for protecting public dataset copyrights. In this paper, we find existing DOV methods (implicitly) assume that the verification process is \emph{faithful}, where the suspicious model will directly verify ownership by using the verification samples as input and returning their results. However, this assumption may not necessarily hold in practice and their performance may degrade sharply when subjected to intentional or unintentional perturbations. To address this limitation, we propose the first certified dataset watermark (\ie, CertDW) and CertDW-based certified dataset ownership verification method that ensures reliable verification even under malicious attacks, under certain conditions (\eg, constrained pixel-level perturbation). Specifically, inspired by conformal prediction, we introduce two statistical measures, including principal probability (PP) and watermark robustness (WR), to assess model prediction stability on benign and watermarked samples under noise perturbations. We derive provable certification conditions relating WR to a PP-based calibration threshold, and a high-probability upper bound on the false positive rate, enabling ownership verification when a suspicious model's WR value significantly exceeds the PP values of multiple benign models trained on watermark-free datasets. If the number of PP values smaller than WR exceeds a threshold determined via conformal calibration, the suspicious model is regarded as having been trained on the protected dataset. Extensive experiments on benchmark datasets verify the effectiveness of our CertDW method and its resistance to potential adaptive attacks. Our codes are at \href{https://github.com/NcepuQiaoTing/CertDW}{GitHub}. 
\end{abstract}

}

\maketitle
\IEEEdisplaynontitleabstractindextext

%
\IEEEpeerreviewmaketitle

\IEEEraisesectionheading{\section{Introduction}}
\label{sec:intro}

\IEEEPARstart{R}{ecently}, deep neural networks (DNNs) have been widely and successfully adopted and deployed in many mission-critical applications, such as face recognition \cite{tang2004video,qiu2021end2end,yang2023larnext}. Arguably, their success heavily relied on the existence of diverse and high-quality public datasets (\eg, ImageNet \cite{deng2009imagenet} and LAION-5B \cite{schuhmann2022laion}). Researchers and developers can use them to train their models and improve their DNNs based on the evaluation results. In particular, these datasets are often mainly limited to educational or research purposes, as their collection and annotation are time-consuming and even costly.

However, safeguarding their copyright (\ie, preventing unauthorized usage of datasets) remains a challenging task, despite there being already many classical in data protection \cite{du2025sok}. These methods cannot be used to protect public datasets, as they either hinder accessibility and functionality (\eg, encryption \cite{thorpe2013coprime,martins2017survey,chuman2018encryption}) of these datasets or necessitate the information of the training process of suspicious models or even its manipulation (\eg, digital watermarking \cite{hsu1999hidden,jain2003hiding,baluja2019hiding} and differential privacy \cite{abadi2016deep,zhu2021fine,li2023multi}) that are not capable for dataset owners in practice.

To the best of our knowledge, dataset ownership verification (DOV) \cite{li2023black, li2022untargeted, tang2023did, guo2024domain,li2025rethinking} is currently the most widely used and effective method to safeguard the copyright of public datasets \cite{du2025sok}. In general, DOV is a post-hoc auditing method, verifying whether a suspicious third-party model is trained on the protected dataset by examining its prediction behaviors on particular samples (\ie, verification samples) without knowing its parameters and training details (\ie, under the black-box setting). These methods consist of two main stages, including dataset watermarking and ownership verification. In the first stage, the dataset owner will modify a few samples in the original (unprotected) dataset to generate its watermarked version, such that all models trained on the watermarked dataset will behave normally on benign testing samples yet have distinctive and pre-defined prediction behaviors (\eg, targeted misclassification) on verification samples. Given the API of a suspicious model, in the second stage, the dataset owner will examine whether this model has dataset-specified behaviors in predicting verification samples. If these special prediction behaviors occur, it is regarded to have been trained on the protected dataset.

In this paper, we revisit existing DOV methods. We reveal that their success relies on a latent assumption that the verification process is `honest', \ie, the suspicious model will faithfully use verification samples (without adding any noises) to generate their predictions. However, this assumption may not necessarily hold in practice, especially when the suspicious model notices the potential progress of ownership verification. We demonstrate that their performance will degrade sharply, no matter under unintentional random noises or intentional adversarial perturbations. Besides, we notice that there is a close connection between dataset watermarking and model watermarking. In particular, a few pioneering research \cite{bansal2022certified,jiang2023ipcert,ren2023dimension} showed that we can achieve certified model watermarks that are robust against any parameter perturbations within a certain region by introducing random noise to the model parameters to limit the side effects in the worst-case scenario. Accordingly, an intriguing and critical question arises: \emph{Could we also achieve certified watermark against image-level noises in the verification process of dataset ownership verification?}

The answer to the aforementioned question is in the positive, although we cannot directly generalize existing methods in certified model watermarks. This is mostly because they focused on the parameter space rather than the sample space, requiring defenders to obtain gradients $w.r.t.$ model parameters or even customize the whole training process. In this paper, we make the first attempt to design certified dataset watermarks to provide robustness guarantees for dataset ownership verification. Our method (dubbed `CertDW') ensures that ownership can be reliably verified as long as the image-level watermark perturbation during the inference process satisfies certain conditions (\eg, constrained pixel-level perturbation). 
In general, inspired by conformal prediction \cite{vovk2012conditional,angelopoulos2021gentle}, which is a statistical technique used to create prediction sets with assured coverages, we introduce two statistics, dubbed `principal probability (PP)' and `watermark robustness (WR)', to characterize the prediction behaviors of benign and watermarked samples under noise perturbations, respectively. Specifically, principal probability is defined as the largest class-wise prediction probability averaged over benign samples under a noise distribution, while watermark robustness is defined as the minimum target-label prediction probability across watermarked samples under a noise distribution. In particular, we prove there is a lower bound on their gap if the sample-level perturbations on verification samples are within a certain range. As such, the suspicious model can be regarded as trained on the protected watermarked dataset (without authorization), if its WR value is significantly larger than the PP values of multiple benign models independently trained on watermark-free datasets.

In practice, our method consists of three main steps. In the first step, we estimate the PP value by selecting the maximum value of the average prediction distribution (PD) computed across classes, obtained by introducing random noise multiple times to several benign samples. The PD hereby represents the probability distribution for each class predicted by the benign model when random noise is added to benign testing samples. In the second step, we estimate the WR value by selecting the minimum value of the probability distribution for the suspicious model predicting the target class, obtained by introducing random noise multiple times to several watermarked samples. In the third step, we calculate the PP values of multiple benign models (dubbed `calibration set') and count the number of their values that are smaller than WR via conformal calibration, a rank-based thresholding procedure inspired by conformal prediction. As long as this number is sufficiently larger than a proportion of the size of calibration set, the suspicious model will be denoted as being trained on the protected dataset. We use multiple instead of solely one validation model for ownership verification to further reduce the randomness of model selection.

In summary, our contributions are four-fold, as follows: \textbf{(1)} We revisit existing dataset ownership verification (DOV) methods and reveal that their verification performance may degrade sharply when noises are incorporated during the inference process. \textbf{(2)} We make the first attempt to design certified dataset watermarks to provide robustness guarantees for dataset ownership verification based on two introduced statistics (\ie, principal probability and watermark robustness). \textbf{(3)} We theoretically analyze the robustness guarantee and its conditions of our certified dataset ownership verification, and provide a high-probability false positive rate bound for models not trained on the protected dataset. \textbf{(4)} We conduct experiments on benchmark datasets to validate the effectiveness of our method and its resistance to potential adaptive attacks.

\section{Related Works}
\label{sec:related_works}

\subsection{Backdoor Attacks}
\label{sec:backdoor_attack}

Backdoor attacks are an emerging research field primarily targeting the training phase of deep neural networks (DNNs) \cite{li2022backdoor}. In such attacks, an adversary maliciously manipulates a subset of training samples to implant a backdoor into the victim model, establishing a latent association between an adversary-specified trigger pattern and a target label. The compromised model performs normally when predicting benign samples, but once an input contains the trigger pattern, its predictions are maliciously altered, posing significant security risks to DNN-based applications. Generally, existing backdoor attacks can be categorized into three main types, based on the adversary’s capabilities: \textbf{(1)} poison-only attacks \cite{ qi2023revisiting, gao2023not,cai2024toward}, \textbf{(2)} training-controlled attacks \cite{li2022few,mo2024robust,zhang2024detector}, and \textbf{(3)} model-modified attacks \cite{qi2022towards,dong2023one,yang2024not}. Specifically, poison-only attacks can only manipulate the training dataset but cannot interfere with the training process; training-controlled attacks can modify both the training dataset and the training procedure (\eg, altering the training loss function); while model-modified attacks inject backdoors by directly modifying model structures or parameters, making them more effective in both digital and physical environments. In this study, we mainly focus on poison-only attacks to leverage their unique properties to design watermarking technique for dataset ownership verification, aiming to protect public datasets. Other types of backdoor attacks can also be used for positive purposes \cite{ya2024towards,li2025move,qiao2026dssmoothing,qi2026cert}, but this is out of the scope of this paper.

\subsection{Data Protection}
\label{sec:data_protection}

\subsubsection{Classical Data Protection}

Data protection is a classic and significant field of study, encompassing various aspects of data security with the goal of preventing unauthorized data usage and safeguarding personal data. Currently, encryption, digital watermarking, and privacy protection are the three main categories of conventional data protection methods. Specifically, encryption \cite{thorpe2013coprime,martins2017survey,chuman2018encryption} protects sensitive data by fully or partially encrypting it, allowing only authorized users with the key to decrypt and utilize the data further. However, this method may limit the functionality of datasets (\eg, accessibility) and is thus not suitable for protecting datasets that are already public. Digital watermarking  \cite{hsu1999hidden,yang2025swap,yang2025promptcos} involves embedding a pattern specified by the owner into the protected data as a watermark to assert ownership. Privacy protection \cite{abadi2016deep,li2023multi,gong2026armor} focuses on preventing the leakage of sensitive information during the training process through empirical methods \cite{xiong2020adgan,li2021visual,xu2021audio} and certified methods \cite{zhu2019deep,zhu2021fine,bai2022multinomial}. These approaches often require access to more detailed training processes, which are not disclosed to users (especially dataset owners), thereby can not be directly used to protect the copyright of public datasets.

\subsubsection{Dataset Ownership Verification}

Dataset ownership verification (DOV) aims to verify whether a suspicious third-party model is trained on the protected dataset. To the best of our knowledge, this is currently the most widely used and effective method for protecting the copyright of open-source datasets \cite{du2025sok,shao2025databench}. Specifically, DOV strategies are post-hoc auditing methods that maintain the model's performance on benign test samples while inducing distinctive prediction behaviors on verification samples by introducing imperceptible watermarked samples into the original dataset to generate its watermarked version for release. Dataset owners verify ownership by checking whether the suspicious third-party model exhibits these distinctive prediction behaviors, all within a black-box verification setting. Current DOV methods \cite{li2023black,li2022untargeted,tang2023did} are primarily implemented through poison-only backdoor attacks or by watermarking unprotected benign datasets through other approaches \cite{guo2024domain,wei2024pointncbw,li2024towards}. For instance, \cite{li2023black} employed poisoned-label backdoor attacks, whereas \cite{tang2023did} adopted clean-label backdoor attacks for dataset watermarking. Li\textit{ et al.}  \cite{li2022untargeted} initially discussed the `harmlessness' requirement of DOV, stating that dataset watermarks should not introduce new security risks to models trained on protected datasets, and proposed the concept of untargeted backdoor watermarks. Recently, Guo\textit{ et al.}  \cite{guo2024domain} further explored the definition of harmlessness, using hardly-generalized domain as watermarked samples to avoid introducing any new vulnerabilities. Additionally, Wei \textit{et al.} \cite{wei2024pointncbw} designed a scalable clean-label backdoor-based dataset watermark for point clouds, capable of watermarking samples from all classes. Most recently, Li \textit{et al.} \cite{li2024towards} proposed the first copyright protection method for personalized text-to-image diffusion models. However, existing DOV lack quantitative research and theoretical guarantees on the robustness of dataset watermarking, and thus may be vulnerable to future advanced adaptive attacks.
    
\subsection{Certified Robustness}
\label{sec:cert_Robust}

Certified robustness \cite{yang2020randomized,kung2024towards,pfrommer2024asymmetric} ensures that a model produces the desired output (such as the correct label) when adversarial perturbations applied to the input remain within a certain region. Initially introduced for certifying classifiers against adversarial examples, the most classical technique for achieving certified robustness is randomized smoothing \cite{lecuyer2019certified,cohen2019certified}. This method works by adding random (Gaussian) noise to a given test input and then using the classifier to predict the final output based on the noisy inputs.

Besides ensuring certified adversarial robustness, a few pioneering recent studies \cite{bansal2022certified,jiang2023ipcert,ren2023dimension} focused on the robustness certification of model watermarking to protect model ownership, achieving this by adding random noise to model parameters. These methods can provide theoretical guarantees in the worst-case scenario in weight-level perturbations, ensuring that the watermark remains non-removable, as long as the perturbation in the model parameters remains within a certain region. However, this protection method is still primarily used for safeguarding model copyrights, focusing on the parameter space rather than the sample space, which often requires additional training details (\eg, gradients or even the entire training process). As such, existing certified model watermarking methods cannot be directly generalized to achieve certified dataset watermarking against image-level noise in the verification process of dataset ownership verification. How to design a certified dataset watermark for robust ownership verification remains blank and is worth further investigation.

\section{Revisiting Dataset Watermarking}
\label{sec:re_DOV}

We find that all existing dataset ownership verification (DOV) methods (implicitly) assume that the verification process is \emph{faithful}, where the suspicious model will directly and exactly use the verification samples as input and return their results. However, the adversaries, \ie, owners of the malicious model trained on the victim dataset, may try to circumvent ownership verification methods by adding perturbations to all  verification samples before feeding them into the model in practice. In this section, we discuss whether the dataset watermarks of existing DOV methods are still effective in these cases. Before we describe our experiment design and observations, we first briefly review the general process of existing DOV methods.

\begin{figure*}[!t]
    \centering
    \vspace{-0.5em}
    \subfigure[GTSRB (BadNets)]{
		\includegraphics[width=0.235\textwidth]{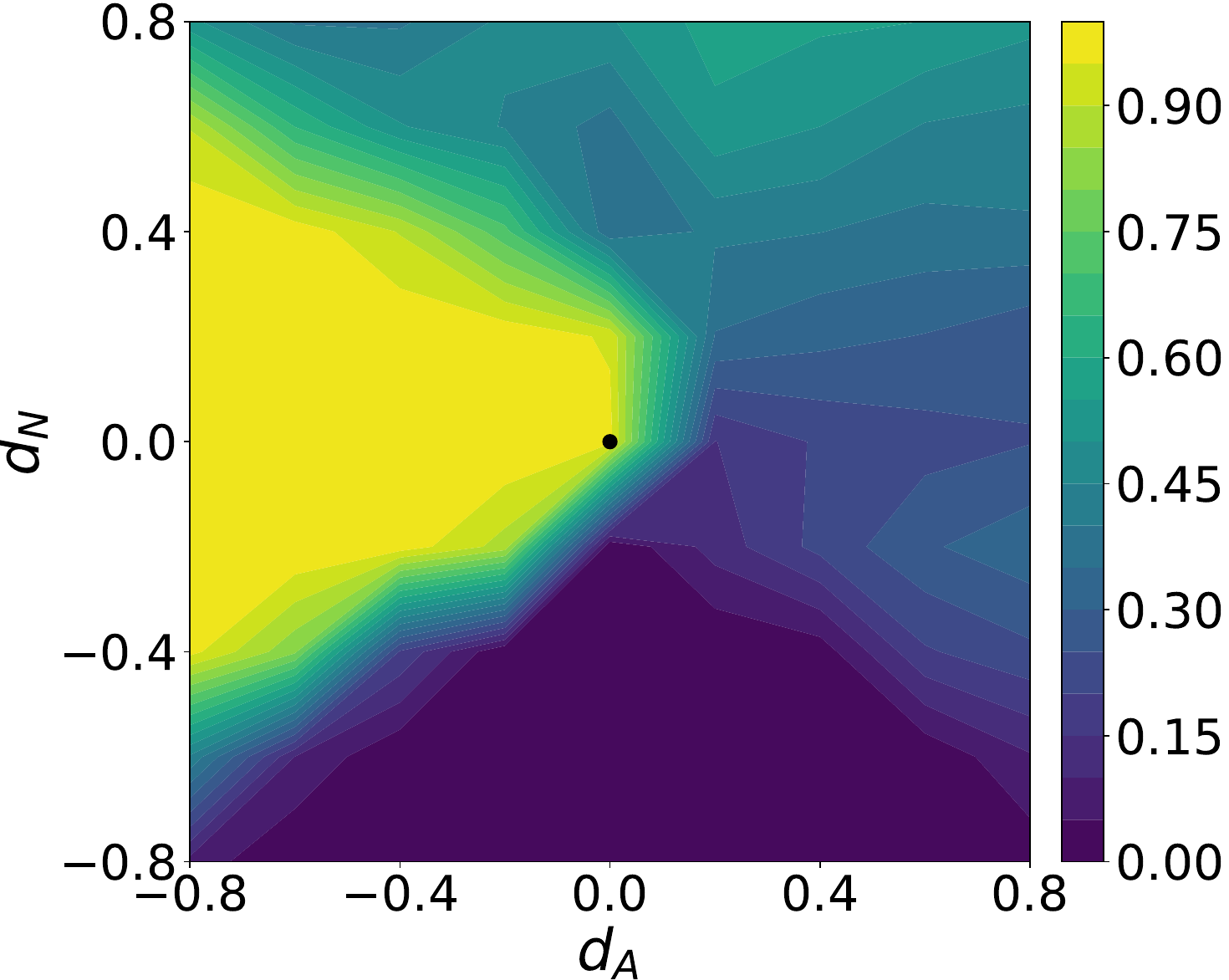}}
        \hspace{0.1em}
    \subfigure[CIFAR-10 (BadNets)]{
		\includegraphics[width=0.235\textwidth]{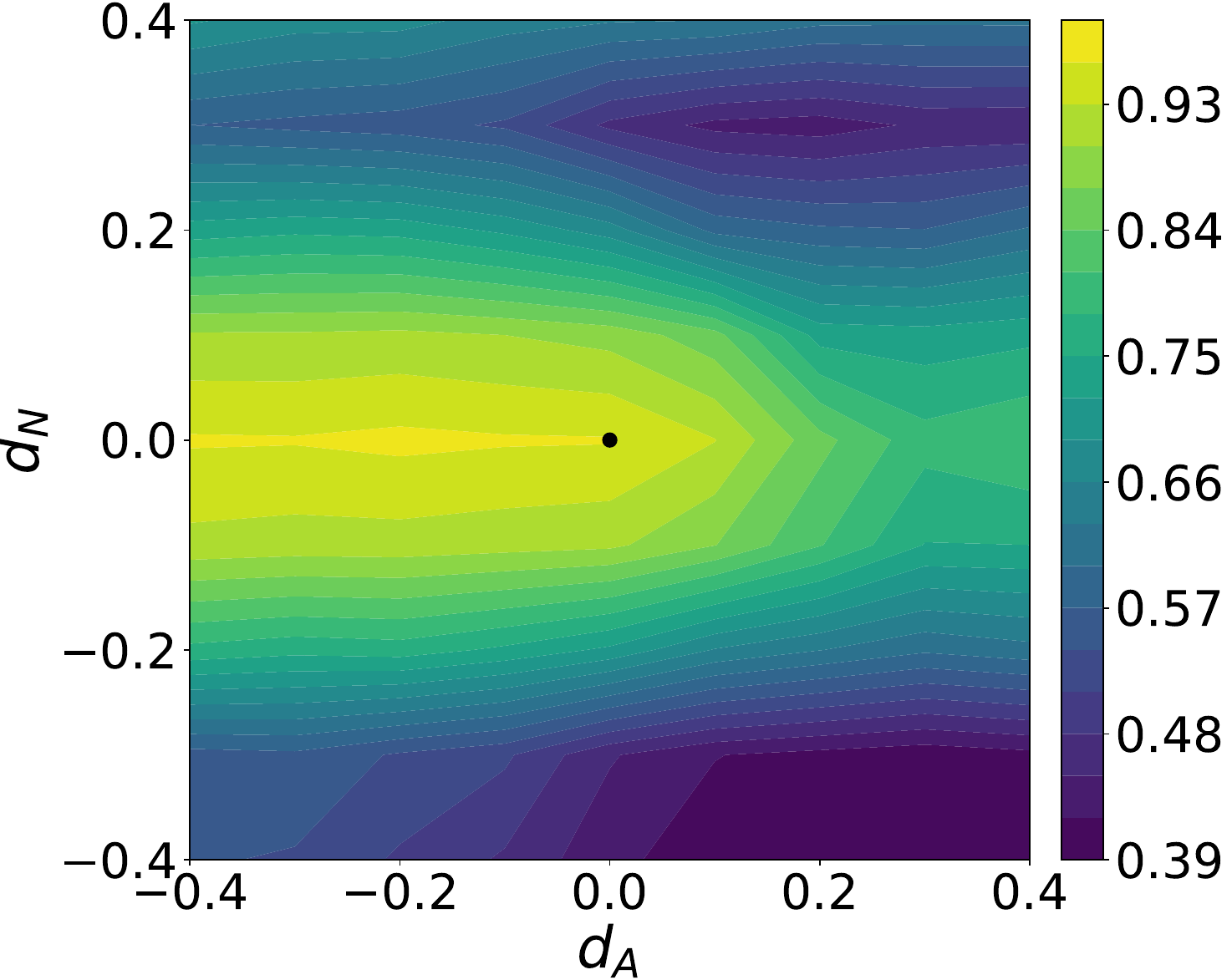}} 
    \hspace{0.1em}
    \subfigure[GTSRB (Blended)]{
		\includegraphics[width=0.235\textwidth]{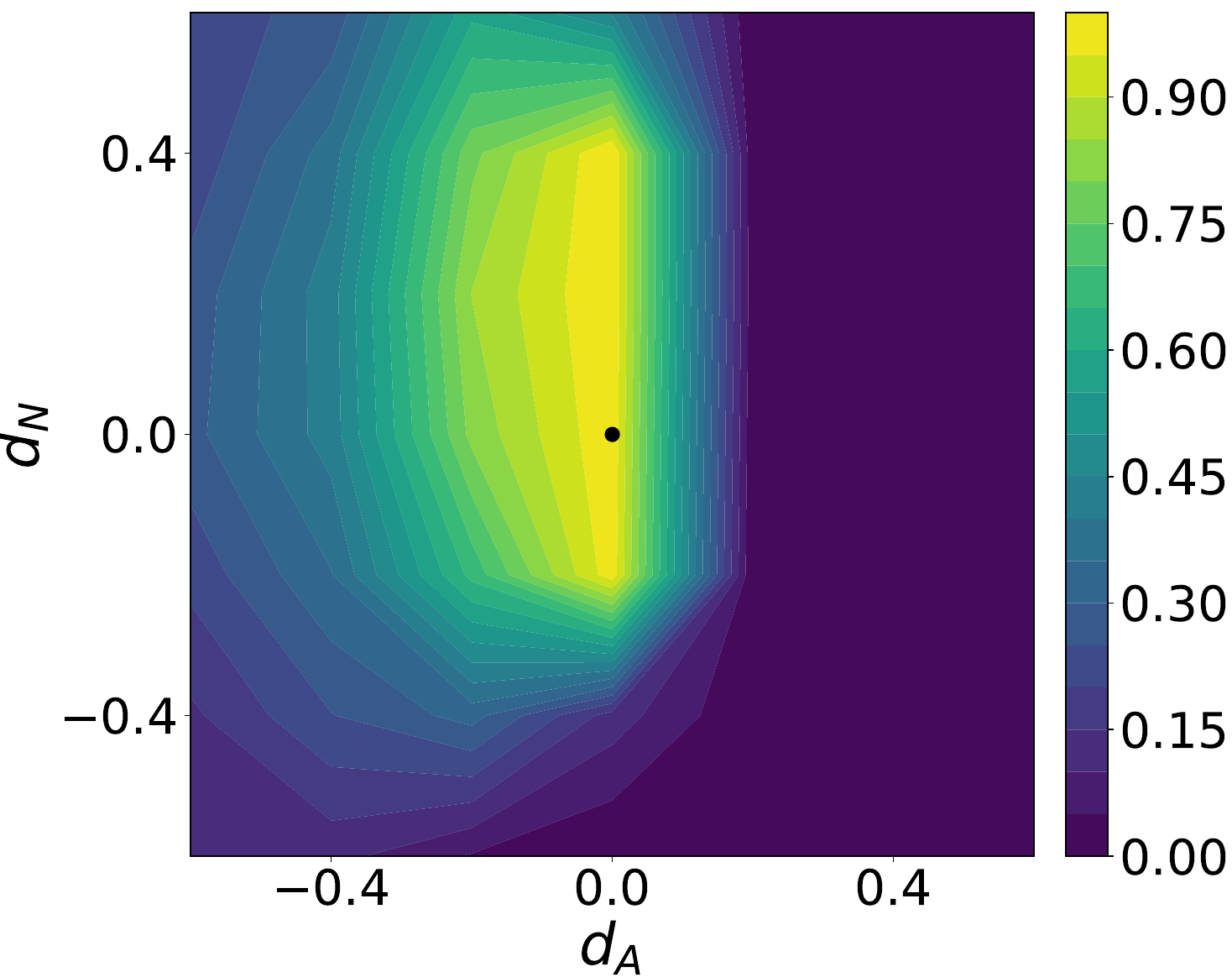}}
         \hspace{0.1em}
         \subfigure[CIFAR-10 (Blended)]{
		\includegraphics[width=0.235\textwidth]{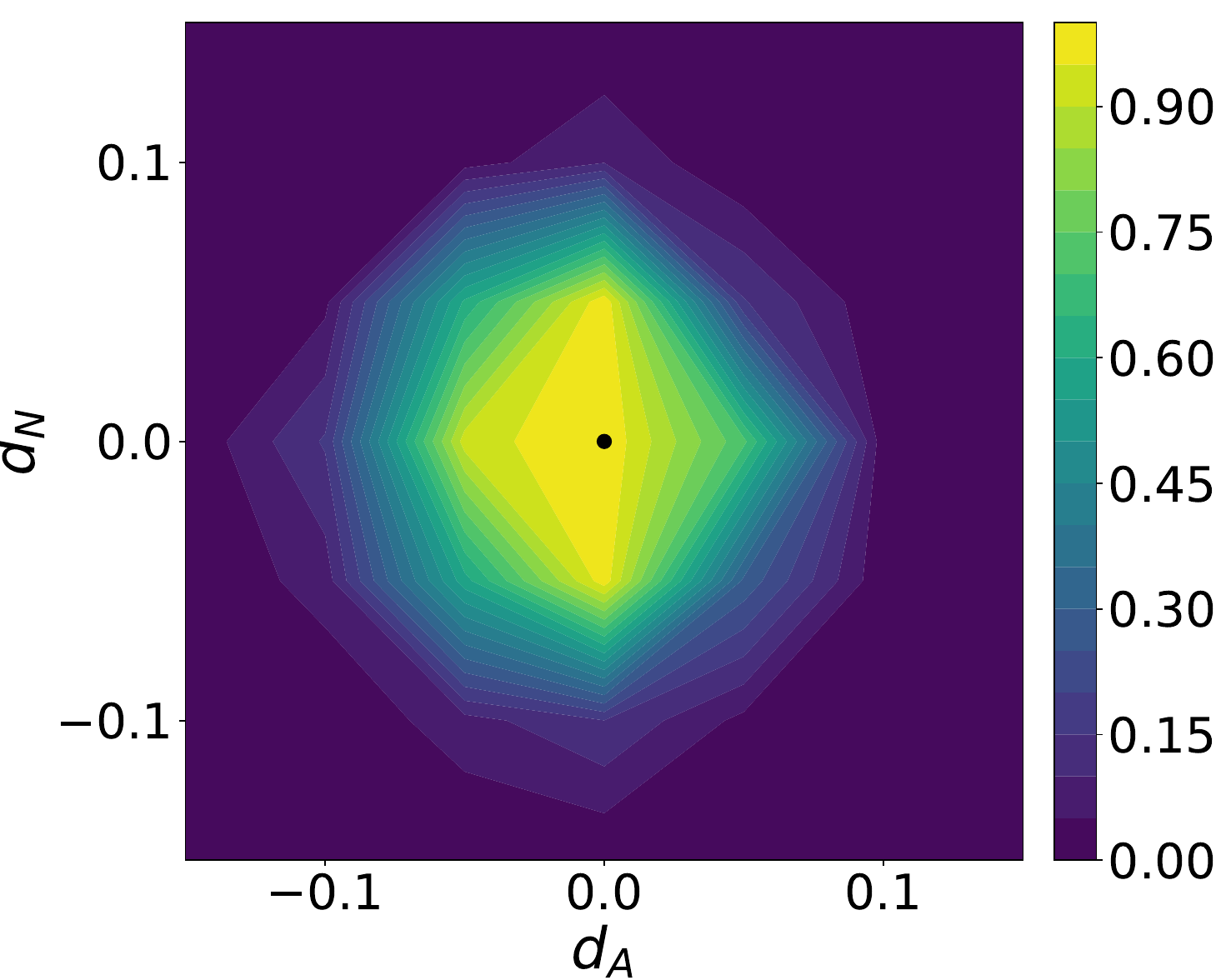}}
        \vspace{-0.9em}
    \caption{The performance (WSR) of watermarked samples in the sample space. $d_N$ is the direction of random noise, and $d_A$ is the adversarial direction. `$\bullet$' denotes the original watermarked sample. The first two columns presents the results under BadNets-based watermarks, while the last two columns show the results of Blended-based watermarks.  }
    \label{fig:noise_adv}
\end{figure*}

\vspace{0.3em}
\noindent \textbf{The Main Pipeline of Existing DOV Methods.}
Let $\gD=\{ (\vx_n,y_n)\}_{n=1}^N$ denotes a vanilla training dataset for an image classification task with $K$ classes, where $\vx_n \in \gX = [0,1]^{C \times W \times H}$ represents the image with $y_n \in \gY = \{1, 2, \cdots, K\}$ is its label. In the first stage of $\mathrm{DOV}$ (\ie, dataset watermarking), the dataset owner creates a watermarked version of $\gD$, denoted as $\DWa$. Specifically, $\DWa=\DPo \cup \DRe$, where $\DPo$ represents the modified version of samples from a small selected subset $\gD_s$ of $\gD$ (\ie, $\gD_s \subset  \gD$) and $\DRe$ contains the remaining benign samples (\ie, $\DRe =\gD \setminus \gD_s$). The $\DPo$ is generated by the dataset-specified image generator $\mG_X: \gX \rightarrow \gX $ and the label generator $\mG_Y: \gY \rightarrow \gY $, \ie, $\DPo=\{(\hat{\vx},\hat{y}):\hat{\vx} = G_{X}(\vx),\hat{y} = G_{Y}(y),(\vx,y) \in \gD_s \}$. For example, in a BadNets-based DOV \cite{li2023black, gu2019badnets}, $\mG_X = \vt \odot \vx+(1-\vt) \odot \vdelta$ and  $\mG_Y=\hat{y}$, where $\vt \in [0,1]^{C \times W \times H}$ is the trigger mask, $\vdelta  \in [0,1]^{C \times W \times H}$ is the trigger pattern, $\odot$ denotes the element-wise product, and $\hat{y}$ is the target label. In particular, $\gamma \triangleq \frac{\left | \DPo \right |}{\left | \DWa \right |}$ denotes the watermarking rate. In the second phase (\ie, ownership verification), the dataset owners investigate whether a suspicious third-party model $f(\cdot;\vtheta) : \gX \rightarrow \gY$ was trained on the protected watermarked dataset $\gD_w$ by querying it with verification samples under the black-box setting. For example, BadNets-based DOV used watermarked samples $\hat{\vx}$ as verification samples and verified whether $f(\hat{\vx};\vtheta)=\hat{y}$.

\begin{figure}[!t]
    \centering
    \subfigure[GTSRB]{
		\includegraphics[width=0.23\textwidth]{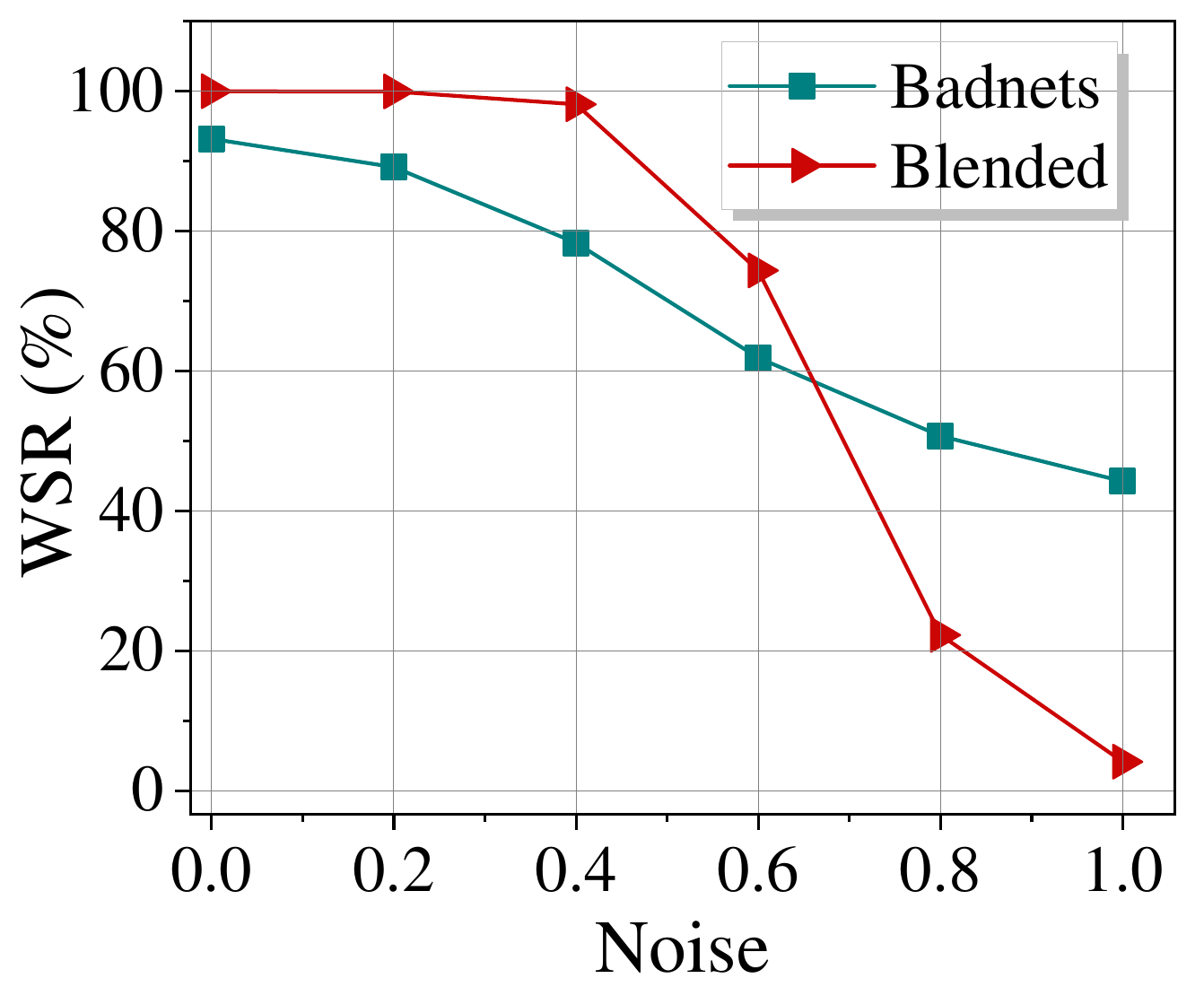}}
    \hspace{0.3em}
    \subfigure[CIFAR-10]{
		\includegraphics[width=0.23\textwidth]{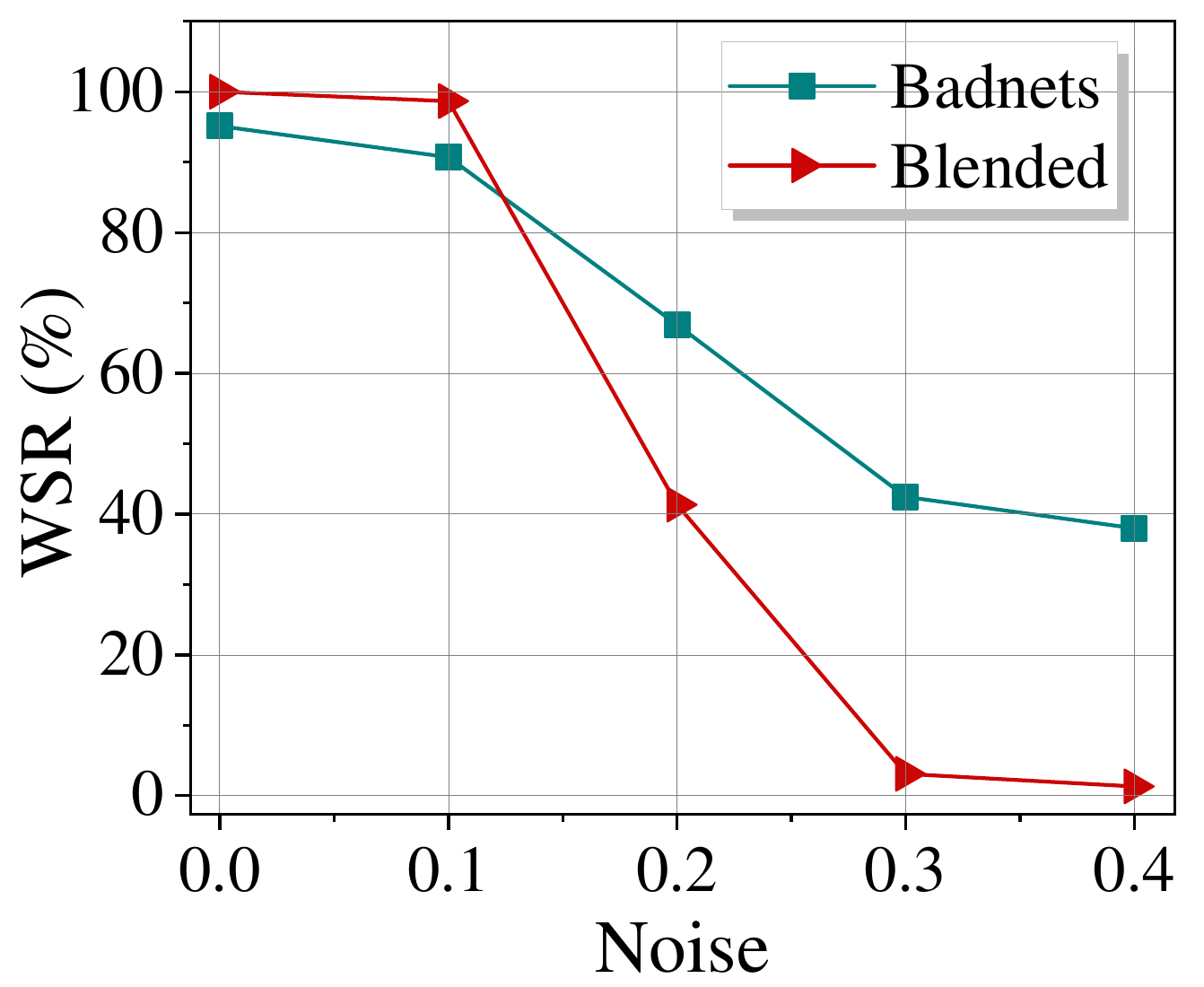}}
        \vspace{-1em}
    \caption{The impact of unintentional random Gaussian noise on the watermark success rate (WSR).} 
    \vspace{-1em}
    \label{fig:noise}
\end{figure}

\subsection{Impact of Unintentional Random Noises}
\label{impact_noise}

In this section, we explore whether random noises that are not intentionally introduced by the adversaries will reduce the performance of existing dataset watermarks.

\vspace{0.3em}
\noindent \textbf{Settings.} We hereby use BadNets \cite{gu2019badnets} and Blended \cite{li2022untargeted} as the watermarking techniques on GTSRB \cite{stallkamp2012man} and CIFAR-10 \cite{krizhevsky2009learning} datasets for discussion. They are the most classical methods and the representative of visible and invisible backdoor watermarks, respectively. Specifically, we set the target label $\hat{y}$ as `1' and set the watermarking rate as 10\% for both datasets. Following previous works \cite{gao2023not,gao2024backdoor}, we use a $3 \times 3$ black-and-white patch located at the lower-right corner of the image as the trigger pattern for BadNets-based watermarks. For the blended watermark, we use a Hello-Kitty trigger, blending it with the original images and setting the transparency parameter to 0.2. Besides, we exploit the winning model from the leaderboard \cite{GTSRB} on GTSRB and a classical VGG-type model \cite{carlini2017towards} on CIFAR-10. Both watermarks are implemented using BackdoorBox \cite{li2023backdoorbox}. During the inference process, we add random noises sampled from a Gaussian distribution to each pixel of verification samples with different magnitudes.

\vspace{0.3em}
\noindent \textbf{Result.} As shown in Figure \ref{fig:noise}, the watermark success rates (WSR) of both watermarks significantly decrease with the increase of noise magnitude. In particular, on the CIFAR-10 dataset, adding noise with a small magnitude (\eg, 0.3) can reduce the WSR by 60\%. This indicates that the performance of dataset watermarking degrades significantly in the presence of random noise, suggesting that existing dataset watermarking methods lack robustness and are highly vulnerable even to unintentional random noise.

\subsection{ 
Impact of Intentional Adversarial Perturbations}
\label{sec:impact_adv}

In this section, building on the analysis in Section \ref{impact_noise}, we further investigate whether adversarial perturbations intentionally introduced by adversaries will further degrade the performance of existing dataset watermarking methods.

\vspace{0.3em}
\noindent \textbf{Setting.} 
In order to visualize the region around the watermarked samples, we measured the watermark success rate (WSR) on the panel spanned by two directions $d_N$ and $d_A$. Specifically, $d_N$ represents the unintentional random noise perturbation direction to erase watermark, \ie, $d_N=\sign(\mathcal{N} (0,\sigma^2I))$, and $d_A$ is the intentional adversarial perturbation direction to erase dataset watermark, \ie, $d_A = \sign(\nabla_{\vx} \mathcal{L} (\theta, \vx, y))$. We perturb the original watermarked samples along these two directions to explore the surrounding sample space and record the WSR of the neighboring samples. For comparison purposes, we define the new sample space as follows:

\begin{equation}
\label{eq:relative distances}
\hat{\mathcal{X}} \triangleq \{\hat{\vx} + \varepsilon_N \cdot d_N +\varepsilon_A \cdot d_A| \varepsilon_N, \varepsilon_A \in \mathbb{R}\},
\end{equation}
where ($\varepsilon _N$, $\varepsilon _A$) are the coordinates along each direction, $\hat{\vx}$ is the original watermarked sample, corresponding to the origin in the coordinate system (marked as the black circle for reference). Finally, we evaluate the changes in the watermark success rate within this sample space.

\vspace{0.3em}
\noindent \textbf{Result.} As shown in Figure \ref{fig:noise_adv}, we find that although unintentional random noise can already significantly reduce WSR within a certain range, introducing intentional adversarial perturbations leads to an even more dramatic decrease in WSR. For example, for blended watermarking, a perturbation as small as 0.15 can cause WSR to drop by over 80\%. This suggests that if an adversary intentionally adds such perturbations, they can more effectively remove dataset watermark, leading to the failure of verification.

\section{Methodology}
\label{sec:Methodology}

\subsection{Preliminaries}
 \label{sec:prelim}

\vspace{0.3em}
\noindent \textbf{Threat Model.} Following the classical settings of dataset ownership verification \cite{li2023black,guo2024domain,du2025sok}, we consider two parties (\ie, the dataset owner and the adversary) in our threat model. The dataset owner can modify the original dataset to generate its watermarked version before releasing it. The dataset users (including adversaries) will use it to train their model, no matter in a legitimate or unauthorized manner. Accordingly, the dataset owner has neither the training details of the suspicious model nor its model parameters or intermediate results (\eg, gradients). The owner can verify the ownership solely based on the prediction of the suspicious model (\ie, under the black-box setting). In particular, different from previous works implicitly assumed that the adversaries will faithfully use verification samples to generate their predictions, we assume that they may deliberately circumvent verification by introducing malicious pixel-level perturbations to verification before feeding them for prediction. Arguably, this setup is more realistic and allows for a better assessment of the effectiveness.

\begin{table}[t]
\centering
\caption{Main notations.}
\label{tab:notation}
\renewcommand{\arraystretch}{1.08}
\setlength{\tabcolsep}{3.5pt}
 \footnotesize
\begin{tabular}{@{}p{0.31\linewidth}p{0.64\linewidth}@{}}
\toprule
\textbf{Notation} & \textbf{Description} \\
\midrule
$\mathcal{D}$, $\mathcal{D}_w$ 
& Original dataset and watermarked dataset \\

$\mathcal{D}_s$, $\mathcal{D}_m$, $\mathcal{D}_r$ 
& Selected, modified/watermarked, and remaining subsets \\

$\mathcal{X}$, $\mathcal{Y}$, $K$ 
& Input space, label space, and number of classes \\

$g(\cdot;\boldsymbol{w})$, $f(\cdot;\boldsymbol{\theta})$, $f_s$ 
& Benign model, suspicious model, and suspicious model not trained on the protected dataset \\

$G_{\mathcal{X}}$, $G_{\mathcal{Y}}$ 
& Image generator and label generator \\

$\boldsymbol{\delta}$, $\hat{y}$, $\hat{\boldsymbol{x}}$ 
& Trigger pattern, target label, and watermarked sample \\

$\gamma$ 
& Watermarking rate \\

$\mathcal{P}_N$, $\sigma$, $\boldsymbol{\epsilon}_i$, $M$ 
& Smoothing distribution, Gaussian standard deviation, sampled noise, and Monte Carlo sample size \\

$\boldsymbol{p}(\boldsymbol{x}|g_{\boldsymbol{w}},\mathcal{P}_N)$ 
& Prediction distribution (PD) \\

$P(g_{\boldsymbol{w}},\mathcal{P}_N)$ 
& Principal probability (PP) \\

$W(f_{\boldsymbol{\theta}},\mathcal{P}_N)$ 
& Watermark robustness (WR) \\

$T(\boldsymbol{x})$, $\boldsymbol{r}_k$, $R$, $\mathcal{F}_{R,T}$ 
& Watermark transformation, induced perturbation, perturbation radius, and neighborhood set \\

$S(f_{\boldsymbol{\theta}},\mathcal{P}_N)$ 
& $R$-functionality stability after trigger removal \\

$P_C$, $P_C^j$, $P_C^{(j)}$ 
& Calibration set, raw calibration score, and ordered calibration score \\

$J$, $\kappa$, $m$, $\alpha_0$ 
& Calibration size, outlier ratio, filtered-outlier number, and calibration parameter \\

$\rho$, $\mathbb{I}\{\cdot\}$, $\tau$ 
& Conformal calibration statistic, indicator function, and calibration  threshold \\

$H_0$, $H_1$, $\phi$, $\phi^*$ 
& Null hypotheses, alternative hypotheses, randomized test, and optimal likelihood-ratio test \\

$\beta_1$, $\beta_2$, $\beta_2^*(\alpha_1;H_1)$ 
& Type-I error, Type-II error, and optimal Type-II error \\

$\underline{W}$, $\delta_M$
&
Finite-$M$ lower confidence bound for WR and prescribed failure probability
\\

$C_j$, $F_C$, $\mathsf{T}(f_s,\mathcal{P}_N)$, $F_T$ 
& FPR-analysis statistics and their distributions \\

$\tau_J$, $Z_J$, $\zeta$ 
& FPR threshold, conditional FPR, and confidence parameter \\
\bottomrule
\end{tabular}
\end{table}

\begin{figure*}[!t]
    \centering
    \includegraphics[width=0.95\textwidth]{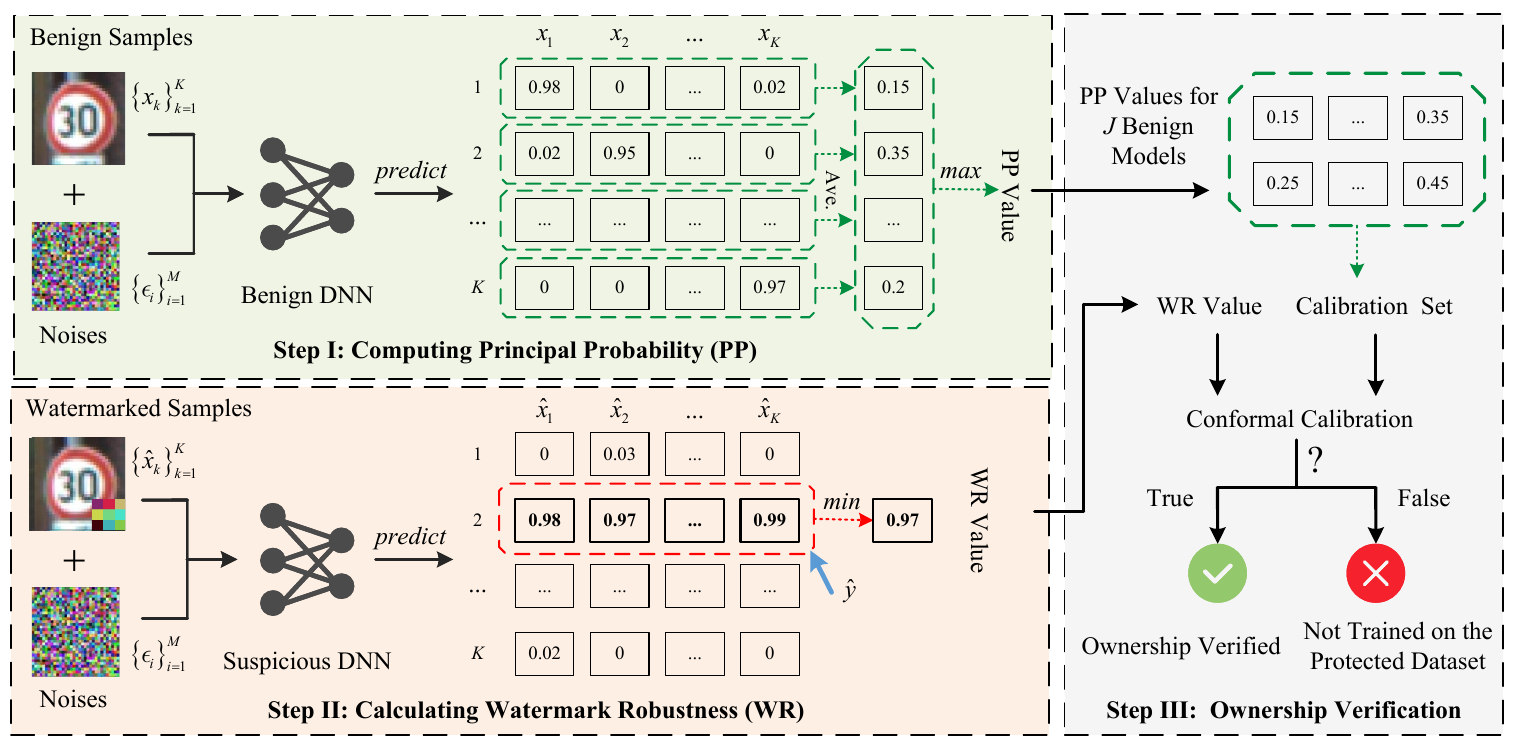}
    \vspace{-0.6em}
    \caption{The main pipeline of our CertDW consists of three core steps.
    In the first step, we randomly select one correctly predicted sample from each class. For each sample, we add $M$ times of noise to predict these noisy versions via a benign model and calculate the frequency of prediction results on each category (termed as prediction distribution (PD)). After that, we estimate the PP value by selecting the maximum value of the average PD across classes; In the second step, we estimate the watermark robustness (WR) value following a similar approach in estimating the PP value. Specifically, we use watermarked instead of benign samples and the suspicious instead of the benign model for prediction. We obtain the WR value by selecting the minimum of the prediction probability on the defender-specified target label (\eg, $\hat{y}=2$). In the third step, we construct a calibration set by calculating PP values for $J$ benign models. We then employ conformal calibration to count the number of values in the calibration set that are smaller than the WR value for ownership verification. If this count is sufficiently large, the suspicious model is deemed to have been trained on the protected dataset.}
    \label{fig:main}
    \vspace{-0.8em}
\end{figure*}

\vspace{0.3em}
\noindent \textbf{Goal of Certified Dataset Watermark (Informal Definition).} We hereby first provide a preliminary definition to briefly describe our goal. Its formal definition is in Section \ref{sec:def_CDW}. Generally, a natural goal in defending against the aforementioned malicious acts (\ie, adding noises before prediction) is to ensure that the prediction of the (suspicious) model on the verification sample remains unaffected by any malicious modification. In this case, even if malicious modifications are intentionally introduced in the verification process, the ownership verification remains effective since the prediction should remain unchanged. The goal of a certified dataset watermark is to ensure the previous property by designing a particular watermarking (and corresponding ownership verification) scheme (under certain conditions). 

\vspace{0.3em}
\noindent \textbf{Notation Summary.}
To improve readability, we summarize the key notations used throughout this paper in Table~\ref{tab:notation}. We also provide a brief conceptual roadmap below to clarify the connections among these quantities.

\vspace{0.3em}
\noindent \textbf{Conceptual Roadmap.}
The key concepts are organized as follows.
We first define the prediction distribution (PD, Definition~\ref{def:pred}), which characterizes a model's prediction behavior under smoothing noise. Based on PD, we introduce the principal probability (PP, Definition~\ref{def:prin}) to characterize the prediction behavior of benign models, and the watermark robustness (WR, Definition~\ref{def:WR}) to measure the worst-case probability that a suspicious model predicts watermarked samples as the target label under noise. We then construct a calibration set using the PP values of multiple benign models and derive the ownership verification rule and calibration threshold $\tau$ through the conformal calibration statistic $\rho$ (Proposition~\ref{pro:conformal_prediction}).  For certified robustness analysis, we further generalize watermark perturbations using an \emph{$R$-bounded transformation} $T$ (Definition~\ref{def:lpnorm}), and introduce \emph{stability} $S$ (Definition~\ref{def:properties_cdw}) to characterize whether the learned watermark effect can persist after the trigger pattern is removed. When both WR and $S$ exceed the calibration threshold $\tau$, the dataset watermark is defined as certified robust (Definition~\ref{def:CDW_formal}). Finally, Theorem~\ref{thm:General_condition} provides a general sufficient condition for guaranteed verification of watermarked models, which is instantiated under Gaussian and uniform smoothing distributions in Examples~\ref{example:GS}--\ref{example:US}; Theorem~\ref{thm:certdw_fpr} further provides a high-probability false positive rate (FPR) upper bound for models not trained on the protected dataset. Together, these two theorems constitute the theoretical guarantee of CertDW.

 \subsection{Overview of the Proposed Method}
 \label{sec:method}

As demonstrated in Section \ref{sec:re_DOV}, the performance of existing dataset watermarking methods significantly degrades as the intensity of unintentional random noise or intentional adversarial perturbations increases, often resulting in verification failures. To overcome this limitation, inspired by conformal prediction \cite{vovk2012conditional,angelopoulos2021gentle}, which exploits past observations to determine the precise confidence of new prediction, we propose the first certified dataset ownership verification. Specifically, it is designed based on two statistical measures: principal probability and watermark robustness, which characterize the prediction behaviors of benign and watermarked
samples under noise perturbations, respectively. In general, our method consists of three main steps: \textbf{(1)} computing the principal probability (PP), \textbf{(2)} calculating watermark robustness (WR), and \textbf{(3)} dataset ownership verification via conformal calibration, as shown in Figure \ref{fig:main}. Their technical details are in the following subsections.

\subsection{Computing the Principal Probability}
 \label{sec:com_PP}

In this step, we estimate the PP value by selecting the maximum value of the average prediction distribution (PD). In general, PD represents the probability distribution for each class predicted by the benign model when random noise is added to benign testing samples, defined as follows.

\begin{definition}[Prediction Distribution (PD)]\label{def:pred} For benign model $g (\cdot ;\vw ): \gX \rightarrow \gY$ with parameters $\vw$, define $p(\vx|g_{\vw},\pno) \in [0,1]^K$ as a vector representing the probability distribution over $K = |\gY|$ classes when random noise $\vepsilon$ is added to the input $\vx$. The $k$-th entry ($k \in \gY$) of the $\mathrm {PD}$ is defined as:
\begin{equation}
\label{eq:DP}
p_{k}(\vx|g_{\vw},\pno)=\mathbb{P}_{\vepsilon\sim\pno}(\arg\max g (\vx+\vepsilon;\vw)=k),
\end{equation}
where $\vepsilon$ is the noise sampled from a noise distribution\footnote{The noise distribution is also commonly called `smoothing distribution' since it is used for randomized smoothing.} $\pno$, such as a Gaussian distribution, a uniform distribution, etc. 
 \end{definition}

In practice, PD is estimated using Monte Carlo by introducing random noise multiple times to the benign sample, recording the output count for each class, and using frequency to approximate probability. In other words, 

\begin{equation}
\label{eq:DP_MC}
p_{k}(\vx|g_{\vw},\pno) \approx \frac{1}{M} \sum_{i=1}^{M} \mathbb{I}\{\arg\max g(\vx+\vepsilon_i;\vw)=k\},     
\end{equation}
where $M$ is the number of sampled random noises and $\mathbb{I}\{\cdot\}$ denotes the indicator function.

However, the randomness in selecting the benign samples may significantly impact the results. To reduce its side-effects, for a given benign model $g (\cdot ;\vw )$, we independently sample $K$ correctly predicted samples from each of the $K$ classes, denoted as $\vx_1, \vx_2, \dots, \vx_K$. We then calculate the final prediction distribution (PD) for each of these $K$ samples individually and average them by class.

Given the estimated prediction distribution, we then calculate the principal probability (PP) as follows:

\begin{definition}[Principal Probability (PP)]
\label{def:prin} 
Consider a domain with  $ K=\left | \gY  \right | $ classes under the smoothing distribution $\pno$. The $\mathrm {PP}$ for a benign model $g (\cdot ;\vw )$ is defined as

  \begin{equation} 
  \label{eq:PP}
      P(g_{\vw},\pno)=\Vert \frac{1}{K}\sum_{k=1}^K p(\vx_k|g_{\vw}, \pno)\Vert_{\infty},
  \end{equation}
  where $\vx_1,\cdots ,\vx_K $ are $K$ independent random samples satisfying $\argmax g_{\vw}(\vx_k)=k \ (k \in \{1, \cdots, K\}) $.
	\end{definition}

In general, PP measures the highest prediction probability of a benign model on its most likely class under noise perturbations. Although its value may vary across datasets, models, and smoothing settings, a benign model typically does not systematically concentrate the predictions of samples from different classes on one particular class. This property makes PP a baseline to contrast with the target-label-concentrated WR (see Definition \ref{def:WR}) of a suspicious model. Moreover, computing PP across more samples per class usually yields a similar value; therefore, using $K$ instead of more samples is sufficient for its estimation.

 \subsection{Calculating the Watermark Robustness}
 \label{sec:cal_WR} 
 
In this step, we estimate the WR value by selecting the minimum probability of the suspicious model's predictions for the target class. Similar to Section \ref{sec:com_PP}, to mitigate the impact of randomness in sample selection on the final result, we use multiple watermarked samples to compute the probability for the suspicious model. Specifically, given a suspicious model $f (\cdot ;\vtheta )$, we independently sample  $K$ correctly predicted samples from each of the $K$ classes, denoted as $\vx_1, \vx_2, \dots, \vx_K$. We then embed a trigger $\vdelta$ and assign the designated target label $\hat{y}$ to construct  $K$ watermarked samples. Finally, we compute the watermark robustness (WR) as follows:

\begin{definition}[Watermark Robustness (WR)]
 \label{def:WR}
 Consider a dataset watermark with a trigger $\vdelta$ and a target class $\hat{y} \in \gY$ against a suspicious model $f (\cdot ;\vtheta )$. For each sample $\vx \in \gX$, under a noise distribution $\pno$, the WR for the watermarked sample $\hat{\vx}$ (\ie, $\vx+\vdelta$) can be defined based on the PD as:

\begin{equation}
\label{eq:WR}
W(f_{\vtheta},\pno)= \min _{k=1, \cdots, K} p_{\hat{y}}(\hat{\vx}_k|f_{\vtheta},\pno),
\end{equation}
 where $\hat{\vx}_k=\vx_k + \vdelta$, and $\vx_1,\cdots ,\vx_K $ are $K$ independent random samples satisfying $\argmax g_{\vw}(\vx_k)=k \ (k \in \{1, \cdots, K\}) $. 
 \end{definition}

\begin{remark}
 For suspicious model $f (\cdot ;\vtheta )$, the $\hat{y}$-th entry of the PD (on $\hat{\vx}$) is defined as $p_{\hat{y}}(\hat{\vx}|f_{\vtheta},\pno)=\mathbb{P}_{\vepsilon\sim\pno}(\arg\max f(\vx+\vdelta+\vepsilon;\vtheta)=\hat{y})$. Selecting the smallest probability from $K$ classes better reflects the watermark's robustness against noises, ensuring to the greatest extent that the watermark does not fail. This watermark robustness (WR) will be theoretically justified in the subsequent parts (see Section \ref{sec:general condition CDW}), which present the minimum conditions required to satisfy a certified dataset watermark.
\end{remark}

In general, WR measures the worst-case (\ie, lowest) prediction probability that a suspicious model predicts watermarked samples as the target label $\hat{y}$ under noise perturbations. This value is large and concentrated for a watermarked model, since it has learned the trigger–target association and keeps predicting as $\hat{y}$ even under noise. This allows us to decide whether a suspicious model was trained on the protected dataset by checking whether its WR exceeds the  PP values of benign models.

\subsection{Ownership Verification via Conformal Calibration}
\label{sec:ov_cp}

In this step, inspired by conformal prediction, we calculate the PP values of multiple benign models (dubbed `calibration set') and count the number of their values that are smaller than WR for ownership verification. If this count is sufficiently larger than a certain proportion of the calibration set size, the suspicious model is deemed to have been trained on the protected dataset. We hereby exploit multiple benign models instead of a single one for verification to reduce the side effects of randomness in model selection.

Specifically, we first train $J$ benign models following the method in Section \ref{sec:com_PP} and calculate the PP values of them individually to construct a calibration set, denoted as $P_C(g_{\vw}, \pno ) = \{P_C^1(g_{\vw},\pno), \ldots, P_C^J(g_{\vw},\pno)\}$. In particular, the calibration set is composed of PP values calculated by benign models trained on a dataset instead of the actual data distribution. As such, the PP values calculated on these benign models may exhibit distribution shifts, particularly with a higher sample variance and heavy tails with many outliers in the calibration set. Directly using this calibration set for threshold construction may lead to overly conservative verification thresholds. To alleviate this problem, we need to filter out a certain proportion of larger outliers (\eg, by outlier detection). Finally, we exploit conformal calibration based on the WR value of the suspicious model and the PP values in the calibration set to calculate the conformal calibration statistic $\rho$ for ownership verification, as follows.

 \begin{proposition}[Dataset Ownership Verification via Conformal Calibration]
\label{pro:conformal_prediction}
Let $P_C$ and $W$ denote the PP values of $P_C(g_{\vw}, \pno)$ and  the WR values over $W(f_{\vtheta},\pno)$, respectively. The $P_C(g_{\vw}, \pno)$ and $W(f_{\vtheta},\pno)$ is estimated based on Definition \ref{def:prin} and Definition \ref{def:WR}, respectively.  The conformal calibration statistic $\rho$ is defined by
		\begin{equation}
  \label{eq:p-value}
   \begin{array}{ll}
				\rho = \frac{1 + \min \left\{ \sum_{j=1}^{J} \mathbb{I}\{P_C^j < W\}, J-m \right\}}{J - m + 1},  
    \end{array}
		\end{equation}	
  where $J$ is the size of the calibration set, $m$ represents the number of outliers in the calibration set, and $\mathbb{I}\{\cdot\}$ is an indicator function whose value is 1 if $P_C^j < W$ and 0 otherwise. We claim that the suspicious model is trained on the protected dataset if $\rho \ge 1-\alpha_0$, where $\alpha_0\in(0,1)$ (\eg, 0.05) denotes a user-specified calibration parameter that controls the rank-based verification threshold.
	\end{proposition}

In practice, $m$ is chosen as $m=\lfloor \kappa \cdot J \rfloor$, where $\kappa\in[0,1)$ is a hyper-parameter indicating the proportion of filtered outliers (\eg, $\kappa=0.2$). The detailed verification procedure of CertDW is summarized in Appendix~\ref{app:certdw} (Algorithm~\ref{alg:certdw}).

\begin{remark}\label{remark:2}
The previous condition of unauthorized training, \ie, $\rho \ge 1-\alpha_0$, can be equivalently reformulated as 
\begin{equation} \label{eq:calibration_threshold}
W(f_{\vtheta},\pno) > P_C^{(J - m - \left \lfloor \alpha_0 (J - m + 1) \right \rfloor)}(g_{\vw},\pno), 
\end{equation} 
where $P_C^{(j)}(g_{\vw},\pno)$ denotes the $j$-th smallest element in $P_C(g_{\vw}, \pno)$. In this paper, $P_C^{(J - m - \left \lfloor \alpha_0 (J - m + 1) \right \rfloor)}(g_{\vw},\pno)$ is called the `calibration threshold'. We emphasize that $\rho$ is a rank-based conformal calibration statistic rather than a classical conformal $p$-value with finite-sample coverage guarantees. The corresponding false-positive-rate (FPR) control guarantee is provided later in Theorem~\ref{thm:certdw_fpr} (see Section \ref{sec:general condition CDW} for details).
\end{remark}

\section{Theoretical Analyses of our CertDW} 
\label{sec:theorem-proof}

In this section, we provide theoretical analyses of our CertDW proposed in Section \ref{sec:Methodology}. Before presenting the theory, we first define (sample-level) certified dataset watermarking. According to this definition, we propose a general theoretical framework applicable to various noise distributions based on Neyman-Pearson lemma \cite{neyman1933on}. Besides, we also instantiate this framework with two classical smoothing distributions, \ie, Gaussian and uniform distributions and derive their specific conditions for a better illustration.

\subsection{Definition of Certified Dataset Watermarking} 
\label{sec:def_CDW}

In this section, we first define the neighborhood based on $R$-bounded transformation and provide a rigorous formulation for dataset watermark perturbations. Based on this definition, as well as the watermark robustness in Definition \ref{def:WR}, we present two necessary properties of certified dataset watermarking (see Definition \ref{def:properties_cdw}), which facilitate the theoretical analysis in Section \ref{sec:general condition CDW}. We provide the formal definition of certified dataset watermarking at the end.

 \begin{definition}
 \label{def:lpnorm}
  The $R$-bounded transformation-based neighborhood set of the example $(\vx, y)$, \ie, $\gF_{R,T}(\vx, y)$, is defined as:
 \begin{equation}
 \begin{array}{ll}
 \gF_{R,T}(\vx, y)=\left \{ (T(\vx), \hat{y}) \mid \mathrm {dist}(T(\vx),\vx)\le R\right \},
 \end{array}
 \end{equation}
 where $T(\vx): \gX \rightarrow \gX$ is a sample-wise transformation, $\hat{y} \in \gY$ is the defender-specified target class, and $\mathrm {dist}(\cdot, \cdot)$ is a predefined distance metric (\eg, $\ell_p$-norm). Here, $R \geq 0$ denotes the maximum perturbation magnitude of the dataset watermarking, representing the upper bound for perturbation strength. 
 \end{definition}

\begin{remark}\label{remark:3}
The set $\gF_{R,T}$ is a general form, adapting to various widely used perturbation bounds by selecting appropriate transformation functions $T$ and distance metrics. In this paper, we mainly focus on the pixel-level additive transformation with $\ell_p$-norm $(1 \le p \le \infty)$. It is worth noting that we need to assign an $R$ so that $\mathrm {dist}(T(\vx),\vx)\le R$ holds for all $K$ selected samples $\{\vx_1, \cdots, \vx_K\}$ for certified dataset watermarks. For convenience, for each $k \in \{1, \ldots, K\}$, we define $\vr_k = T(\vx_k) - \vx_k$. Accordingly, $R$ can be the supremum of $\{\mathrm {dist}(T(\vx_k),\vx_k)\}_{k=1}^K$, denoted by $R= \max _{k=1,\cdots, K}\left \| \vr_k \right \|_2$.
\end{remark}

Based on Definition~\ref{def:lpnorm}, we hereby provide two necessary properties that a certified dataset watermark should satisfy. Here, ``Transformation-based'' indicates that the WR is formulated under the general transformation framework introduced in Definition~\ref{def:lpnorm} (rather than the trigger-specific form in Definition~\ref{def:WR}); ``$R$-functionality'' indicates that the Stability property is defined under the $R$-bounded perturbation constraint $\|\boldsymbol{r}_k\|_2 \leq R$.

   \begin{definition}[Two Necessary Properties of Certified Dataset Watermarks]
\label{def:properties_cdw}
Let $\{\vx_k\}_{k=1}^K$ be $K$ independent random benign samples satisfying $\argmax g_{\vw}(\vx_k)=k$, and $\vepsilon$ be the noise sampled from a noise distribution $\pno$. Consider the watermarked version of $\vx$ (\ie, $\vx+\vr$) with a defender-specified target class $\hat{y}$ against a watermarked model $f (\cdot ;\vtheta)$. 
    \begin{itemize}
	\item \textbf{(Transformation-based) Watermark Robustness (WR):} WR is defined as the lower bound of the probability that watermarked samples are consistently predicted as the target label under the given noise distribution, as shown in Definition \ref{def:WR}. We hereby re-formulate it (by replacing $\bm{\delta}$ with $\bm{r}$) based on Remark \ref{remark:3}, as follows:
 
   \begin{equation}
    \label{eq:noise_robust}
    W(f_{\vtheta},\pno) =\min_{k}\mathbb{P} (\arg\max f(\vx_k + \vr_k + \vepsilon)=\hat{y}).
    \end{equation}
     \item \textbf{($R$--functionality) Stability:} Given that the watermark transformation is constrained within $R$ (\ie, $\Vert\bm{\vr_k}\Vert_2\leq R$), it is defined as the average probability that $K$ benign samples are predicted as the defender-specified target label $\hat{y}$ under noise 
perturbations, which can be viewed as the $\hat{y}$-th entry of the averaged PD in Definition \ref{def:pred}, paralleling the structure of PP in Definition \ref{def:prin}, as follows:

  \begin{equation}
    \label{eq:R_functionality}
   S(f_{\vtheta},\pno)=\frac{1}{K}\sum_{k=1}^{K} \ \mathbb{P} (\arg\max f(\vx_k + \vepsilon)=\hat{y}).
    \end{equation}
		\end{itemize}	
		
	\end{definition}

 In general, $W(f_{\vtheta},\pno)$ and $S(f_{\vtheta},\pno)$ measure the resistance of the dataset watermark to noises and to watermark-removal attacks (as well as noises) on verification samples, respectively, and the two are complementary. WR concerns the case where noise is added on top of the watermarked sample (\ie, $\vx_k+\vr_k+\vepsilon$) and uses $\min_k$ to capture the worst case, ensuring that even the least-robust watermarked sample is reliably predicted as $\hat{y}$; whereas Stability ($S$) concerns the extreme case where the trigger is entirely removed (\ie, $\vx_k+\vepsilon$) and uses $\frac{1}{K}\sum_k$ to capture the average behavior across the $K$ classes, so that the certified condition reflects the overall behavior rather than being dominated by outlier samples. We consider both properties instead of solely the watermark robustness to approximate the worst-case scenario where the malicious dataset user removes the trigger pattern somehow (instead of simply adding noises) during the inference process. Accordingly, a dataset watermark is certified robust if its two property values are both sufficiently large. Its formal definition is as follows.

\begin{definition}[(Sample-level) Certified Robust Dataset Watermarking]
\label{def:CDW_formal}
We call a dataset watermark of the watermarked model $f_{\vtheta}$ (under the smoothing distribution $\pno$) is ($\tau$-)certified robust if and only if $\min\{W(f_{\vtheta},\pno), S(f_{\vtheta},\pno)\} > \tau$.
\end{definition}
    
\begin{remark}	
\label{remark:4}
In this paper, to reduce the side effects of the randomness in selecting $\tau$, we assign its value based on the calibration threshold (defined in Remark \ref{remark:2}), which is calculated based on benign models via conformal calibration. In particular, we note that $S$ is computed on the watermarked model $f_{\boldsymbol{\theta}}$, not on a benign model. Hence, $S$ is \emph{not} a false positive rate; see Section~\ref{Exper_setup} (Evaluation Metrics) for details. Besides, following existing certified adversarial robustness/model watermarking \cite{cohen2019certified,ren2023dimension}, we do not incorporate the `utility' requirement that the watermarking should only have mild side effects on the watermarked models in predicting benign testing samples. Nevertheless, we will empirically verify it in our main experiments. 
\end{remark}

\subsection{A General Condition for Certified Watermarking}
\label{sec:general condition CDW}

In this section, we provide theoretical analyses for CertDW from two complementary aspects: certified verification for watermarked models and false positive rate (FPR) control for non-watermarked models. We first formulate a Neyman--Pearson hypothesis test to analyze the stability transfer from the watermarked verification distribution to the transformation-removed verification distribution, and then instantiate the resulting sufficient condition under Gaussian and uniform smoothing distributions. Finally, we provide a high-probability FPR upper bound for the final ownership verification method.

\begin{definition}[Statistical Hypothesis Testing]
\label{def:stat_hypothesis}
 Statistical hypothesis testing is a decision-making problem that involves determining whether a proposed hypothesis is correct. Formally, the decision is based on the realized values of a random variable $X$, whose distribution is known to be either $H_0$ (the null hypothesis) or $H_1$ (the alternative hypothesis). Given a sample $\vx\sim \mathcal{X}$, a random test $\phi$ can be modeled as a function $\phi: \mathcal{X} \longrightarrow [0,1]$, which rejects the null hypothesis with probability $\phi(\vx)$ and accepts the null hypothesis with probability $1-\phi(\vx)$.
\end{definition}

\begin{definition}[Type-I/II Error in the Hypothesis Test]
\label{def:I-II type error}
For the hypothesis test used in our certification analysis, we test the null hypothesis $H_0$ (\ie, the watermarked verification distribution) against the alternative hypothesis $H_1$ (\ie, the transformation-removed verification distribution). Given a randomized test $\phi$, the Type-I/II errors are defined as follows:
		
\begin{itemize}
    \item Type-I Error ($\beta_1$): The probability of rejecting $H_0$ when the sample is drawn from the watermarked verification distribution (\ie, null hypothesis is true but rejected), as follow:
    \begin{equation}
    \beta_1(\phi; H_0)
    =
    \mathbb{E}_{\vx}(\phi(\vx)\mid H_0 \ \text{is true}).
    \end{equation}

    \item Type-II Error ($\beta_2$): The probability of accepting $H_0$ when the sample is drawn from the transformation-removed verification distribution (\ie, null hypothesis is false but accepted), as follow:
    \begin{equation}
    \beta_2(\phi; H_1)
    =
    \mathbb{E}_{\vx}(1-\phi(\vx)\mid H_1 \ \text{is true}).
    \end{equation}
\end{itemize}
\end{definition}

In particular, the above hypothesis test is not used to directly measure the false-positive rate of the final ownership verification
procedure. Instead, it is used to quantify how much the desired
target-label behavior can be transferred from the watermarked verification
distribution to the transformation-removed verification distribution.
Specifically, according to the Neyman-Pearson lemma \cite{neyman1933on}, the likelihood ratio test attains the minimum type-II error under a prescribed type-I error constraint. In our certification analysis, this optimal type-II error is further used as a conservative lower bound on the transformation-removed target-label stability. Formally, we set the significance level
$\alpha_1$ as the maximum acceptable probability of type-I error:
\begin{equation}
\label{eq:optimal_test}
\beta_1(\phi^*; H_0)=\alpha _1, \quad \beta_2(\phi^*; H_1) = \beta_2^* (\alpha_1; H_1),
\end{equation}
where $\beta_2^*(\alpha_1;H_1 )= \inf \{\beta_2 (\phi ;H_1) \mid \beta_1 (\phi ;H_0)\le \alpha _1\}$.

Particularly, different from the classical Neyman-Pearson test, where $\beta_2^*$ is the quantity to be minimized under a prescribed type-I error constraint, in our certified condition we instead favor a \emph{larger} value of $\beta_2^*$. This is because a larger $\beta_2^*$ indicates that, even under the optimal likelihood ratio test $\phi^*$, the watermarked verification distribution and the transformation-removed verification distribution remain hard to distinguish. Such distributional overlap is beneficial for certification, since it implies that the target-label behavior induced by the dataset watermark is more likely to persist (\ie, remain robust) after trigger removal.

By combining the above Definitions \ref{def:stat_hypothesis}-\ref{def:I-II type error} with the optimal likelihood ratio test, \ie, Eq.~(\ref{eq:optimal_test}), we can derive the following general sufficient condition (\ref{eq:general_condition}) of certified robust dataset watermarking. Its proof is provided in Appendix \ref{app:proof_theorem}.

\begin{theorem}[\textbf{General Condition of Certified Dataset Watermarking}]
       \label{thm:General_condition}
 For a watermarked model, let $W(f_{\vtheta},\pno)$ and
$S(f_{\vtheta},\pno)$ be defined as in Eq. (\ref{eq:noise_robust}) and (\ref{eq:R_functionality}) of Definition~\ref{def:properties_cdw}, respectively.
 Dataset ownership is guaranteed to be verified if the optimal type-II error, for testing the null $ \pno+\vr \sim H _0$ against the alternative $\pno \sim H_1$, satisfies the following condition:
      
       \begin{equation}
          \label{eq:general_condition}
				\beta_2^* (1-W(f_{\vtheta},\pno), H_1 )>\pcl^{(J-m-\left \lfloor \alpha_0 (J-m+1) \right \rfloor )}(g_{\vw},\pno),
		\end{equation}
        where $\pcl^{(j)}(g_{\vw},\pno)$ denotes the $j$-th smallest element in $P_{C}(g_{\vw}, \pno )$. $\alpha_0$, $J$ and $m$ are defined as in Proposition \ref{pro:conformal_prediction}.	
	\end{theorem}

\begin{remark}
Different smoothing distributions lead to different robustness boundaries for various norms. Below, the Gaussian case is instantiated for $\ell_2$-bounded perturbations, while the uniform case is presented
as a one-dimensional illustrative example. 
\end{remark}

In general, Theorem~\ref{thm:General_condition} establishes a
one-sided Neyman-Pearson probability transfer that connects WR in
Definition~\ref{def:properties_cdw} with the $R$-functionality stability
$S$. Specifically, WR provides a lower bound on the target-label
prediction probability under the watermarked verification distribution.
By applying the Neyman-Pearson lemma to the complement test of the
target-label decision region, this lower bound can be transferred to the
transformation-removed verification distribution, yielding $S(f_{\vtheta},\pno) \ge \beta_2^*(1-W(f_{\vtheta},\pno),H_1)$. Therefore, once $\beta_2^*(1-W(f_{\vtheta},\pno),H_1)$ exceeds the calibration threshold
$\tau$, which is defined in Remark~\ref{remark:2}, we have $S(f_{\vtheta},\pno)>\tau$. Moreover, since  $\beta_2^*(1-W(f_{\vtheta},\pno),H_1) \le W(f_{\vtheta},\pno)$, the same condition also implies $W(f_{\vtheta},\pno)>\tau$. Consequently, Theorem~\ref{thm:General_condition} guarantees $\min\{W(f_{\vtheta},\pno),S(f_{\vtheta},\pno)\}>\tau$, which leads to certified dataset ownership verification.

Notably, $\beta_2^*(1-W(f_{\vtheta},\pno),H_1)$ serves as a conservative lower bound on the target-label stability after
removing the watermark transformation, rather than as the false-positive
rate of the final ownership verification procedure. This interpretation is
consistent with the Neyman-Pearson-based certification analysis in
RAB~\cite{weber2023rab}, where the optimal type-II error is also used to
characterize the distinguishability between smoothed distributions before
and after perturbation.

We hereby derive the robustness conditions under two specific noise/smoothing distributions (\ie, Gaussian and uniform distributions) as examples for a better illustration. The corresponding certified procedure under the Gaussian distribution is given in Appendix~\ref{app:certdw} (Algorithm~\ref{alg:certdw-cert}).

\begin{example}[Robustness Conditions under Gaussian Distribution]
\label{example:GS}
Let the noise $\vepsilon \sim \mathcal{N}(0, \sigma^2 I)$. Given $W(f_{\vtheta},\pno)$ 
 as defined in Eq. (\ref{eq:noise_robust}) in Definition \ref{def:properties_cdw} for the watermarked model's (transformation-based) WR. Let $R$ denote the maximum perturbation magnitude of the dataset watermark, as defined in Definition \ref{def:lpnorm}. Dataset ownership verification is guaranteed if $W(f_{\vtheta},\pno)$ satisfies the following condition:

\begin{equation}
\label{eq:gs_cp}
 W(f_{\vtheta},\pno)> \Phi(\frac{R}{\sigma } + \Phi^{-1}(\pcl^{(J - m - \left\lfloor \alpha_0 (J - m + 1) \right\rfloor )}(g_{\vw}, \pno)) ),
\end{equation}
where $\Phi$ is the cumulative distribution function (CDF) of the standard Gaussian distribution.
\end{example}

\begin{example}[Robustness Conditions under Uniform Distribution]
\label{example:US}
Let the one-dimensional smoothing noise $\vepsilon \sim \mathcal{U}([e, h])$. Given $W(f_{\vtheta},\pno)$  
 as defined in Eq. (\ref{eq:noise_robust}) in Definition \ref{def:properties_cdw} for the watermarked model's (transformation-based) WR. Let $R$ denote the maximum absolute watermark perturbation magnitude, as defined in Definition \ref{def:lpnorm}. Dataset ownership verification is guaranteed if $W(f_{\vtheta},\pno)$ satisfies the following condition:
\begin{equation}
\label{eq:US_cp}
W(f_{\vtheta},\pno) > P_C^{(J-m-\lfloor\alpha_0(J-m+1)\rfloor)}(g_{\vw},\pno) + \frac{R}{h-e}.
\end{equation}
\end{example}

 \begin{remark} \label{remark:6} We have some critical observations about these examples to get intuition on the robustness condition (\ref{eq:gs_cp})-(\ref{eq:US_cp}):
		
		\begin{itemize}
			
			\item As shown in Eq. (\ref{eq:gs_cp})--(\ref{eq:US_cp}), watermarks with a larger WR are more likely to be guaranteed verification when the watermark perturbation size (\ie, $R$) is fixed. Conversely, if the WR is fixed, dataset watermarks with smaller perturbation sizes are more likely to be guaranteed verification.	
			
			\item  The major distinction between certified dataset watermark verification and certified backdoor robustness is that the former provides verification guarantees for `strong' dataset watermarks, while the latter prevents the learning of triggers during training. These two types of certification indicate that a backdoor watermark is either strong enough to be `detectable' or weak enough to be removed.
			
		\end{itemize}
	\end{remark}

In practice, the smoothing probabilities used to compute WR are estimated
using $M$ Monte Carlo samples. Since these estimates may differ from the
true smoothing probabilities, we further derive simultaneous one-sided
lower confidence bounds for the $K$ selected samples. Specifically, the
resulting simultaneous lower bound $\underline{W}$ satisfies $\mathbb{P}\!\left(
W(f_{\vtheta},\pno)\geq\underline{W}
\right)\geq1-\delta_M$, where $\delta_M\in(0,1)$ denotes the prescribed failure probability.
See Appendix~\ref{app:finite_mc} for details.

In addition, while the above sufficient conditions focus on reliably
verifying watermarked models, it is also important to control the false
positive rate (FPR) of the final ownership verification rule. In particular,
when a suspicious model is not trained on the protected dataset, the statistic
used for false-positive analysis should not exceed the calibration threshold
with high probability. We next provide a high-probability FPR bound for
CertDW based on the PP calibration statistics computed from benign models.

\begin{theorem}[\textbf{High-Probability FPR Control of CertDW}]
\label{thm:certdw_fpr}
Let $C_j=P(g_{\vw}^{(j)},\pno)$, $j=1,\ldots,J$, be the empirical PP
calibration statistics computed using $M$ Monte Carlo samples for $J$
benign models, and assume that
$C_1,\ldots,C_J$ are \textit{i.i.d.} samples from an arbitrary (possibly discrete) distribution
$F_C$. Let $\mathsf{T}=\mathsf{T}(f_{\mathrm{s}},\pno)$ denote a scalar statistic used for
false-positive analysis of a suspicious model $f_{\mathrm{s}}$ that is not
trained on the protected dataset, and let $F_{\mathsf{T}}$ be the distribution of $\mathsf{T}$.
Assume that $\mathsf{T}$ is independent of the calibration set and
\begin{equation}
\label{eq:certdw_fpr_domination_0}
F_{\mathsf{T}}(t)\ge F_C(t),\quad \forall t\in\mathbb{R}.
\end{equation}
Let $C_{(1)}\le\cdots\le C_{(J)}$ be the order statistics of the
calibration set. Following Proposition~\ref{pro:conformal_prediction}, define
the calibration threshold as
$\tau_J=C_{\left(J-m-\lfloor\alpha_0(J-m+1)\rfloor\right)}$, where $J$, $m$,
and $\alpha_0$ are defined in Proposition~\ref{pro:conformal_prediction}.
We assume that the order-statistic index
$J-m-\lfloor\alpha_0(J-m+1)\rfloor$ lies in $\{1,\ldots,J\}$. The FPR with
respect to the statistic $\mathsf{T}$ is given by
$Z_J=\mathbb{P}\left(\mathsf{T}>\tau_J\mid C_1,\ldots,C_J\right)$. Then, for any
$\zeta\in(0,1)$, with probability at least $1-\zeta$ over the random draw
of the calibration set,
\begin{equation}
\label{eq:certdw_fpr_bound_0}
Z_J\le \frac{m+\lfloor\alpha_0(J-m+1)\rfloor}{J} +\sqrt{\frac{\log(2/\zeta)}{2J}}.
\end{equation}
\end{theorem}

\begin{remark}
\label{remark:fpr_statistic}

The statistic $T$ in Theorem~\ref{thm:certdw_fpr} is written in a general form to cover different false-positive analyses. For the final ownership
verification rule, $T$ can be instantiated as the WR statistic of a
suspicious model not trained on the protected dataset. In our empirical FPR
evaluation, we further instantiate $T$ as the PP statistic of such models,
because their WR is usually very small when they have not learned the
watermark-target association. Therefore, using PP provides a stricter empirical FPR evaluation. Empirical validation of the stochastic dominance assumption and the high-probability FPR bound is provided in Appendix~\ref{app:fpr_validation}.
\end{remark}

In general, Theorem~\ref{thm:certdw_fpr} does not require exchangeability
between $T$ and the PP calibration statistics. Instead, beyond the
independence between $T$ and the calibration set, it only assumes the
stochastic dominance relation in Eq.~(\ref{eq:certdw_fpr_domination_0}) and
uses the Dvoretzky-Kiefer-Wolfowitz (DKW) inequality with Massart's tight
constant~\cite{dvoretzky1956asymptotic,massart1990tight} to obtain an
explicit high-probability upper bound on the false positive rate. For a
fixed confidence parameter $\zeta$, the bound shows that the FPR of CertDW
depends on the calibration parameter $\alpha_0$, the outlier-trimming ratio
$\kappa$ defined in Proposition~\ref{pro:conformal_prediction}, and the
calibration size $J$. Since $m=\lfloor \kappa J\rfloor$, we have
$m/J\to\kappa$ as $J\to\infty$. Thus, the leading term in
Eq.~(\ref{eq:certdw_fpr_bound_0}) satisfies
$\frac{m+\lfloor\alpha_0(J-m+1)\rfloor}{J}
\to \alpha_0+(1-\alpha_0)\kappa$, while the concentration term
$\sqrt{\log(2/\zeta)/(2J)}$ vanishes as $J$ increases. Therefore, the high-probability FPR upper bound is asymptotically controlled
around $\alpha_0+(1-\alpha_0)\kappa$. In
particular, when no outliers are trimmed (\ie, $\kappa=0$), the asymptotic
upper bound reduces to  $\alpha_0$.

 \section{Experiments}
 \label{sec:exps}

 \subsection{Main Settings}
  \label{Exper_setup}

 \vspace{0.3em}
\noindent \textbf{Dataset and Model Selection.} We conduct experiments on the GTSRB \cite{stallkamp2012man} and CIFAR-10 \cite{krizhevsky2009learning} datasets. We adopt the same model architectures as those described in Section \ref{sec:re_DOV}, following their standard train-test splits.

\vspace{0.3em}
\noindent \textbf{Baseline Selection.} In this paper, we mainly compare our CertDW method to its variant without using conformal calibration (dubbed `CertDW-V') since there is currently no certified dataset watermark. It sets a threshold by controlling the same false positive rate (\ie, the model-level false positive rate on independent models). Besides, we also provide the results of using vanilla watermarking techniques (dubbed `Vanilla') and those without watermarking (dubbed `independent') for reference.

\begin{table*}[!t]
    
  \captionsetup{font=small}
  \caption{The performance (\%) of dataset watermarking and dataset ownership verification on the GTSRB dataset. The former is measured by benign accuracy (BA) and watermark success rate (WSR), while the latter is measured by verification success rate (VSR) and watermark certified accuracy (WCA). We evaluate all methods under three different noise levels: 1.5, 2.5, and 3.5. VSR on 
the ``Independent'' row corresponds to the FPR. In particular, we mark the best verification results in boldface. 
  }
  \vspace{-0.5em}
  \centering
  \scalebox{0.9}{
    \begin{tabular}{c|c|cc|ccc|ccc|ccc}
      \toprule
      \multirow{2}{*}{Watermark$\downarrow$} & \multicolumn{3}{c|}{$\sigma$$\rightarrow$ } & \multicolumn{3}{c|}{1.5} & \multicolumn{3}{c|}{2.5} & \multicolumn{3}{c}{3.5} \\ \cline{2-13}
      & Method$\downarrow$, Metric$\rightarrow$ & BA & WSR & VSR & WCA & WSR & VSR & WCA & WSR & VSR & WCA & WSR \\
      \hline
      No Watermarking  & Independent & \textbf{97.05} & 0 & 6 & 0 & 0 & 12 & 0 & 0 & 12 & 0 & 0 \\
      \hline
      \multirow{3}{*}{BadNets} & Vanilla & 96.05 & 90.88 & 18 & 0 & 37.78 & 18 & 0 & 32.42 & 36 & 0 & 35.19 \\
       & CertDW-V & 95.43 & \textbf{93.22} & 56 & 16 & \textbf{53.31} & 52 & 22 & \textbf{54.07} & 50 & 28 & \textbf{52.97} \\
       & CertDW & 95.43 & \textbf{93.22} & \textbf{88} & \textbf{28} & \textbf{53.31} &\textbf{72} & \textbf{48} & \textbf{54.07} & \textbf{72} & \textbf{54} & \textbf{52.97} \\
       \hline
      \multirow{3}{*}{Blended (patch)} & Vanilla & 96.10 & 91.37 & 80 & 0 & 34.47 & 70 & 20 & 44.40 & 70 & 30 & 49.79 \\
       & CertDW-V & 96.13 & \textbf{97.99} & 40 & \textbf{6} & \textbf{68.88} & 46 & 8 & \textbf{95.85} & 48 & 10 & \textbf{98.28 }\\
       & CertDW & 96.13 & \textbf{97.99} & \textbf{82} &\textbf{6} & \textbf{68.88 }& \textbf{78} & \textbf{22} & \textbf{95.85} & \textbf{76} & \textbf{36} & \textbf{98.28} \\
       \hline
      \multirow{3}{*}{Blended (noise)} & Vanilla & 96.90 & \textbf{99.92} & 0 & 0 & 20.09 & 0 & 0 & 5.41 & 0 & 0 & 2.37 \\
       & CertDW-V & 95.98 & 96.14 & 90 & 32 & \textbf{66.99} & 86 & 54 & \textbf{68.63} & 86 & 62 & \textbf{68.65} \\
       & CertDW & 95.98 & 96.14 & \textbf{98} & \textbf{66} & \textbf{66.99} & \textbf{96} & \textbf{80} & \textbf{68.63} & \textbf{96} & \textbf{90} & \textbf{68.65} \\
      \bottomrule
    \end{tabular}
  }
  \label{table:GTSRB}
\end{table*}

\begin{table*}[!t]
  \captionsetup{font=small}
  \caption{The performance (\%) of dataset watermarking and dataset ownership verification on the  CIFAR-10 dataset. The former is measured by benign accuracy (BA) and watermark success rate (WSR), while the latter is measured by verification success rate (VSR) and watermark certified accuracy (WCA). We evaluate all methods under three different noise levels: 0.6, 1.2, and 1.8.  VSR on 
the ``Independent'' row corresponds to the FPR. In particular, we mark the best verification results in boldface.
  }
 \vspace{-0.5em}
  \centering
  \scalebox{0.9}{
    \begin{tabular}{c|c|cc|ccc|ccc|ccc}
      \toprule
      \multirow{2}{*}{Watermark $\downarrow$} & \multicolumn{3}{c|}{$\sigma \xrightarrow{}$ } & \multicolumn{3}{c|}{0.6} & \multicolumn{3}{c|}{1.2} & \multicolumn{3}{c}{1.8} \\ \cline{2-13}
      &  Method$\downarrow$, Metric$\rightarrow$  & BA & WSR & VSR & WCA & WSR & VSR & WCA & WSR & VSR & WCA & WSR \\
      \hline
      No Watermarking & Independent &\textbf{82.55} & 0 & 6 & 0 & 0 & 4 & 0 & 0 & 4 & 0 & 0 \\
      \hline
      \multirow{3}{*}{BadNets} & Vanilla & 81.19 & \textbf{94.18} & 62 & 0 & 80.30 & 28 & 14 & 72.54 & 44 & 14 & 62.78 \\
       & CertDW-V & 81.06 & 93.91 & 64 & 20 & \textbf{83.54} & 64 & 24 & \textbf{80.98} & 54 & 24 & \textbf{77.28} \\
       & CertDW & 81.06 & 93.91 & \textbf{70} & \textbf{24} & \textbf{83.54} & \textbf{68} & \textbf{36} &\textbf{80.98} & \textbf{64} & \textbf{40} & \textbf{77.28} \\
       \hline
      \multirow{3}{*}{Blended (patch)} & Vanilla & 80.46 & \textbf{99.99} & 40 & 10 & 61.50 & 50 & 10 & 62.30 & 50 & 10 & 57.61 \\
       & CertDW-V & 80.19 & 99.49 & 42 & 16 & \textbf{67.53} & 44 & 20 & \textbf{70.86} & 42 & 22 & \textbf{68.13} \\
       & CertDW & 80.19 & 99.49 & \textbf{48} & \textbf{20} & \textbf{67.53} & \textbf{52} & \textbf{32} & \textbf{70.86} & \textbf{54} & \textbf{32} & \textbf{68.13} \\
       \hline
      \multirow{3}{*}{Blended (noise)} & Vanilla & 82.06 & 99.45 & 80 & 70 & 90.99 & 80 & 70 & 90.22 & 70 & 60 & 90.37 \\
       &CertDW-V & 80.03 & \textbf{99.55} & \textbf{88} & 70 & \textbf{93.55} & 84 & 78 & \textbf{95.87} & 80 & 62 & \textbf{92.33} \\
       & CertDW & 80.03 & \textbf{99.55} & \textbf{88} & \textbf{72} & \textbf{93.55} & \textbf{88} & \textbf{78} & \textbf{95.87} & \textbf{82} & \textbf{78} & \textbf{92.33} \\
      \bottomrule
    \end{tabular}
  }
  \label{table:CIFAR10}
   \vspace{-1.5em}
\end{table*}

\vspace{0.3em}
\noindent \textbf{Settings for Dataset Watermarking.}  
Following the previous work \cite{li2023black}, we adopt two backdoor watermark methods: BadNets \cite{gu2019badnets} and Blended \cite{li2022untargeted}. We set the target label $\hat{y}$ as `1' and set the watermarking rate as 10\% for both datasets. For the BadNets-based watermarking, we use a 3 $\times$ 3 random pixel patch placed at a random location for each watermark on both datasets. Although the trigger locations are generated randomly, once a trigger is assigned, its position within the watermark samples remains fixed. For the blended-based watermark, we blend a 3 $\times$ 3 random pixel patch with the original images (dubbed `Blended (patch)'), setting the transparency parameter to 0.2. Additionally, we also use a noise pattern applied to the entire image as the trigger for blended-based watermarks (dubbed `Blended (noise)'). For each watermark, a random trigger is generated and embedded as a random perturbation $\vdelta(\vx) = \vx + \vv$, where $\left \| \vv \right \| _2\approx 0.6$. In particular, we use an average $\ell_2$ norm of 1.4 for Blended (noise) on the CIFAR-10 dataset since it usually fails below this threshold. The pixel-level perturbation size is adjusted accordingly to satisfy the $\left \| \vv \right \| _2$ constraint. For example, on the GTSRB dataset, when $\left \| \vv \right \| _2 \approx 0.6$, the perturbation magnitude for each altered pixel is independently and randomly chosen within [40/255, 65/255] for all three channels. Moreover, we create 50 backdoor watermarks for each method on each dataset. Examples are provided in Appendix~\ref{app:sample_examples} (Figure~\ref{fig:examples_imgs}).

\vspace{0.3em}
\noindent \textbf{Settings for Dataset Verification.} We reserve 5,000 samples from the test dataset of GTSRB and CIFAR-10 and train 100 benign models (\ie, $J=100$) following the standard training procedure, respectively. A proportion of filtered outliers $\kappa=0.2$ is set to construct a calibration set for verifying dataset ownership. We train 50 independent models using the full benign training dataset for reference to estimate the false positive rate of our CertDW. During the verification process, we generate 1,024 random Gaussian noises for each input and calculate the WR and PP values using the Monte Carlo estimation method (\ie, $M=1024$). In conformal calibration, the calibration parameter $\alpha_0$ is set to 0.05.

\begin{figure*}[!t]
	\centering  
  \subfigure[]{
		\includegraphics[width=0.32\textwidth]{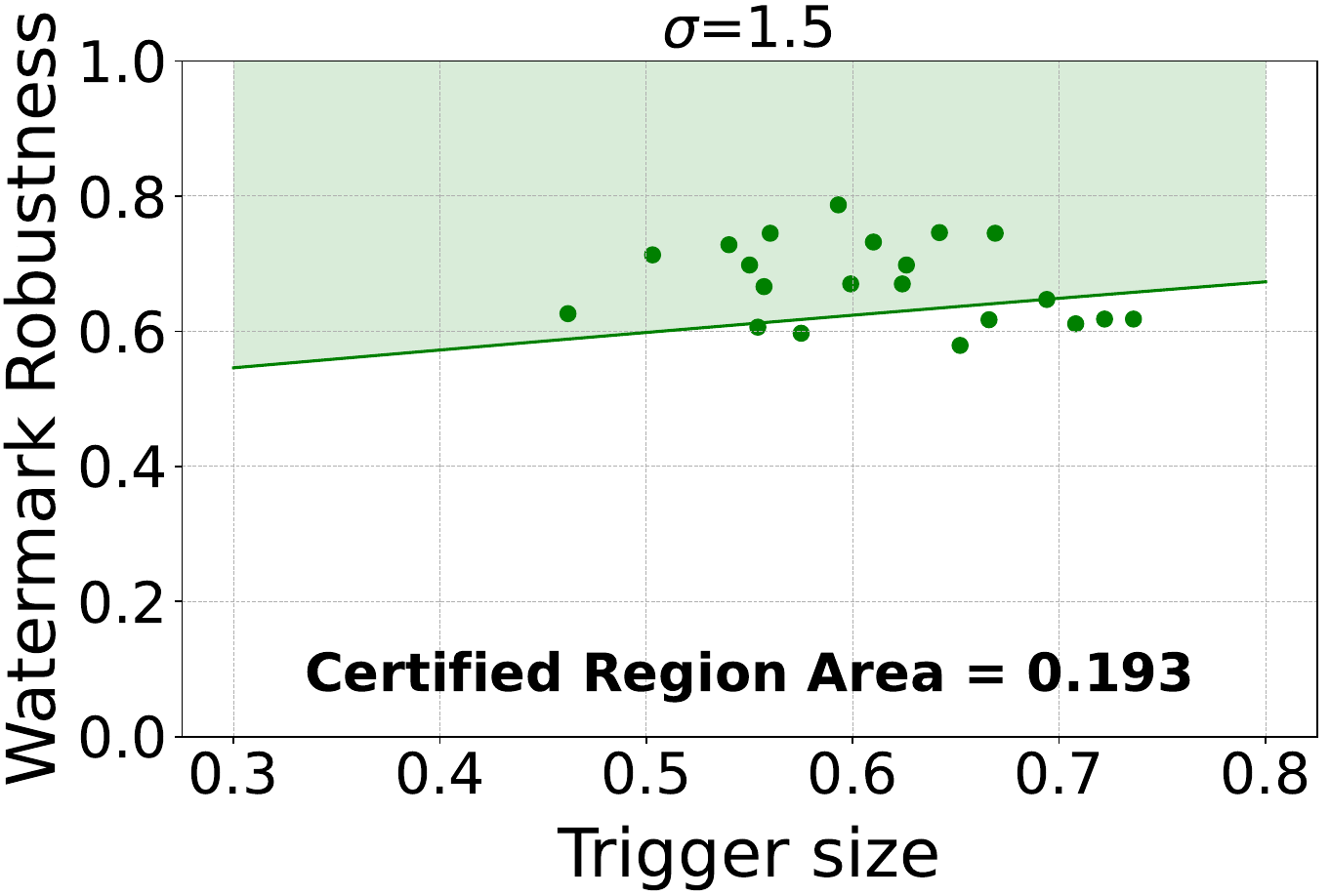}}
	\hspace{0.1em}
      \subfigure[]{
		\includegraphics[width=0.32\textwidth]{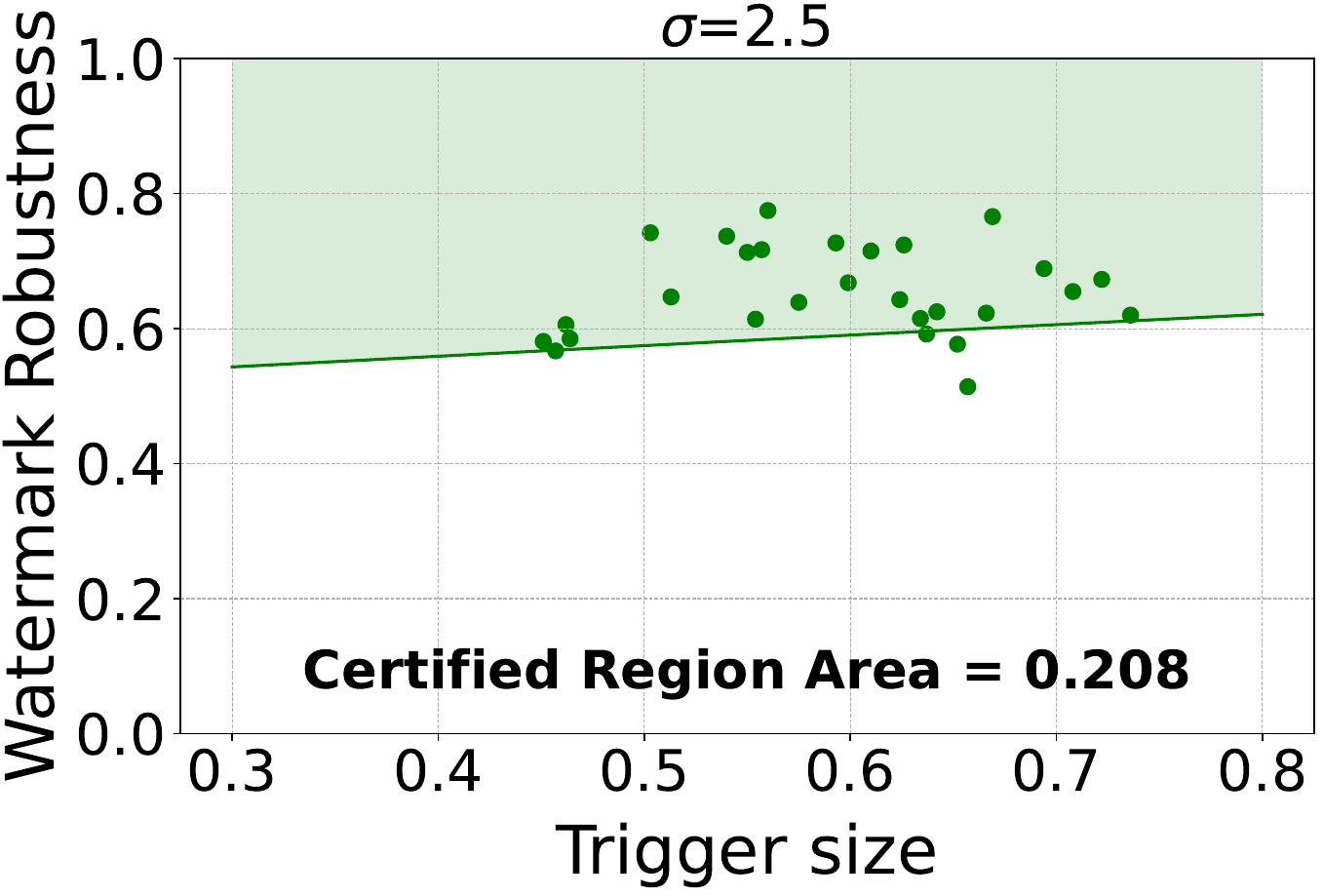}}
	\hspace{0.1em}
     \subfigure[]{
		\includegraphics[width=0.32\textwidth]{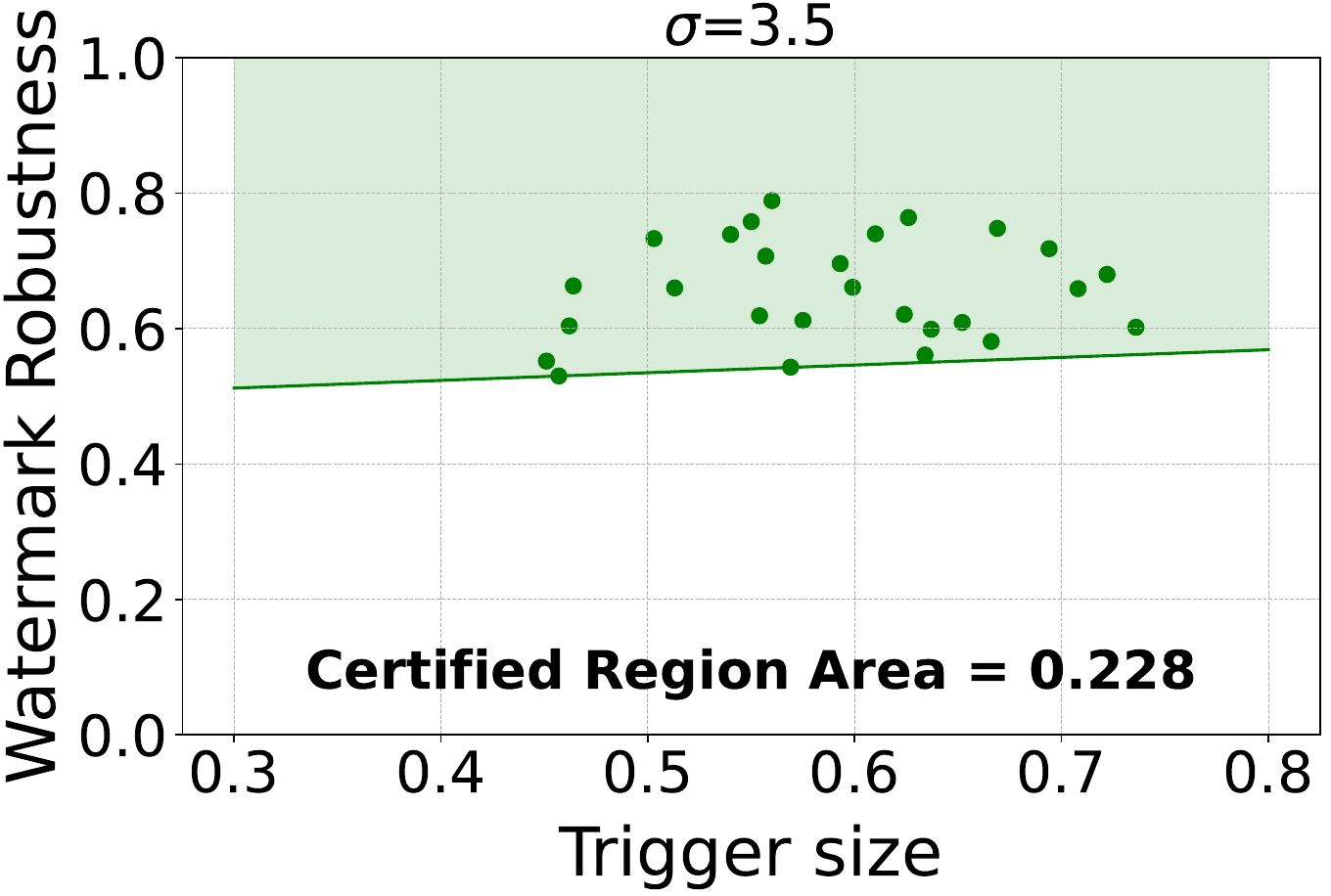}}
	\hspace{0.1em}
     \subfigure[]{
		\includegraphics[width=0.32\textwidth]{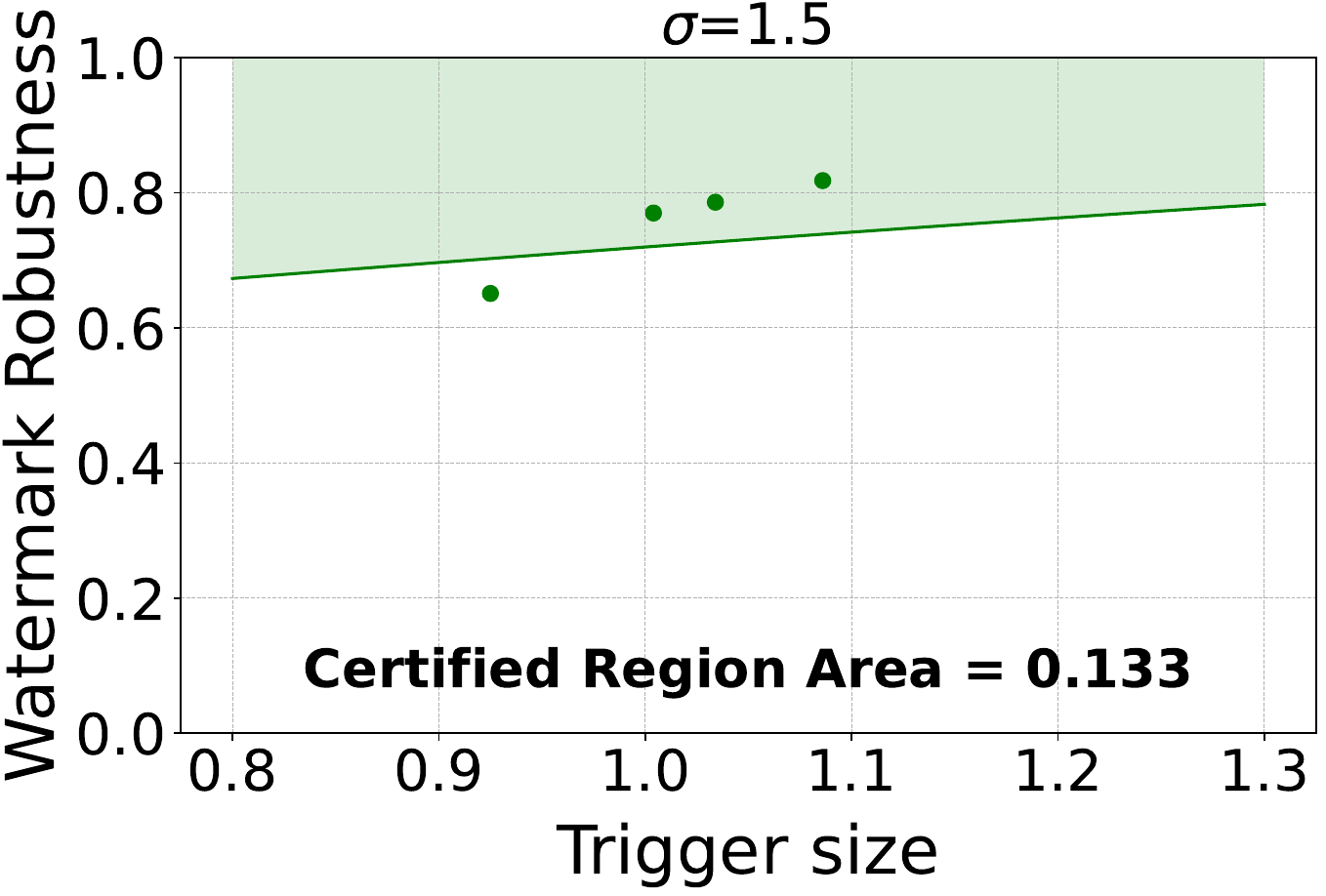}}
	\hspace{0.1em}
	 \subfigure[]{
		\includegraphics[width=0.32\textwidth]{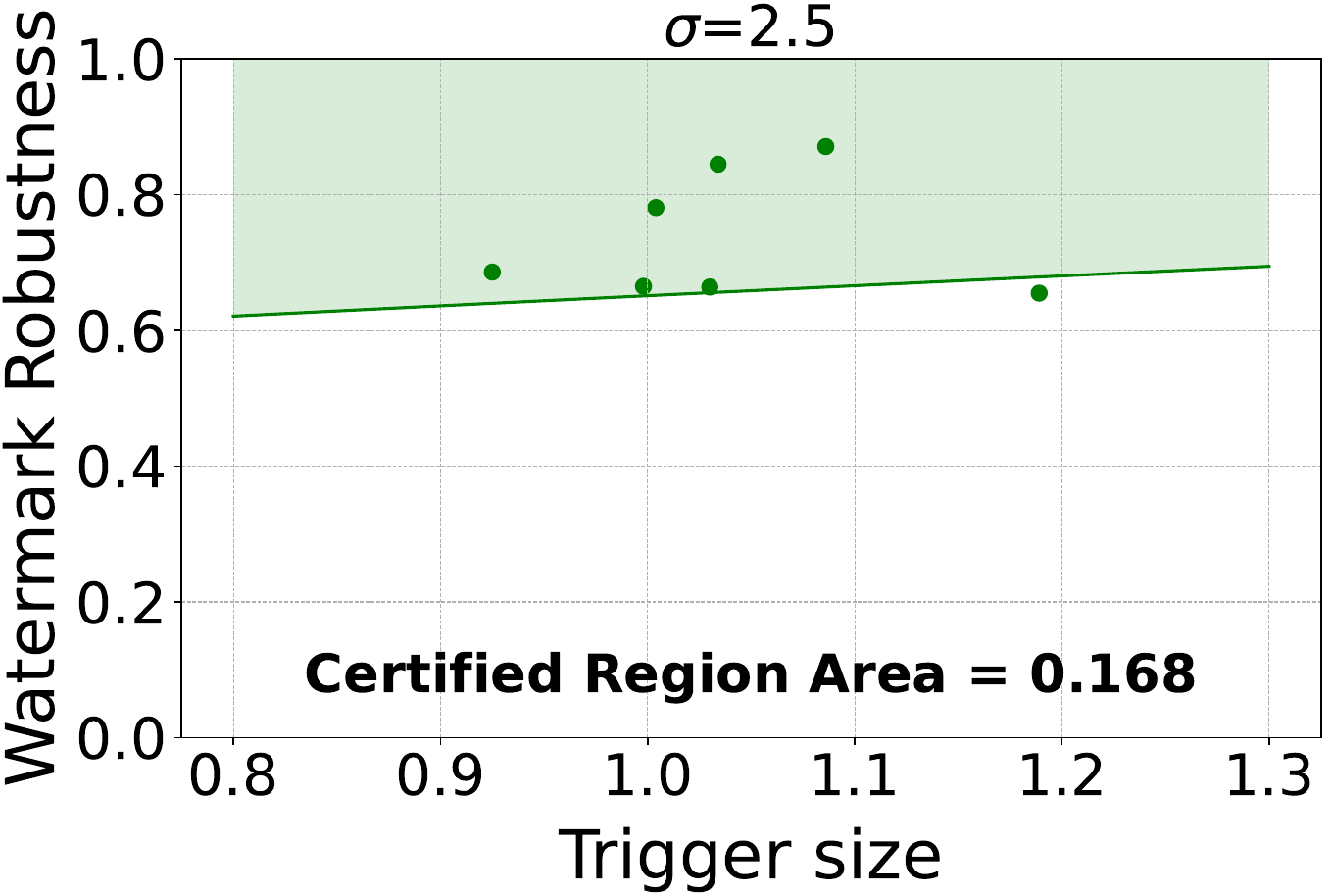}}
	\hspace{0.1em}
	 \subfigure[]{
		\includegraphics[width=0.32\textwidth]{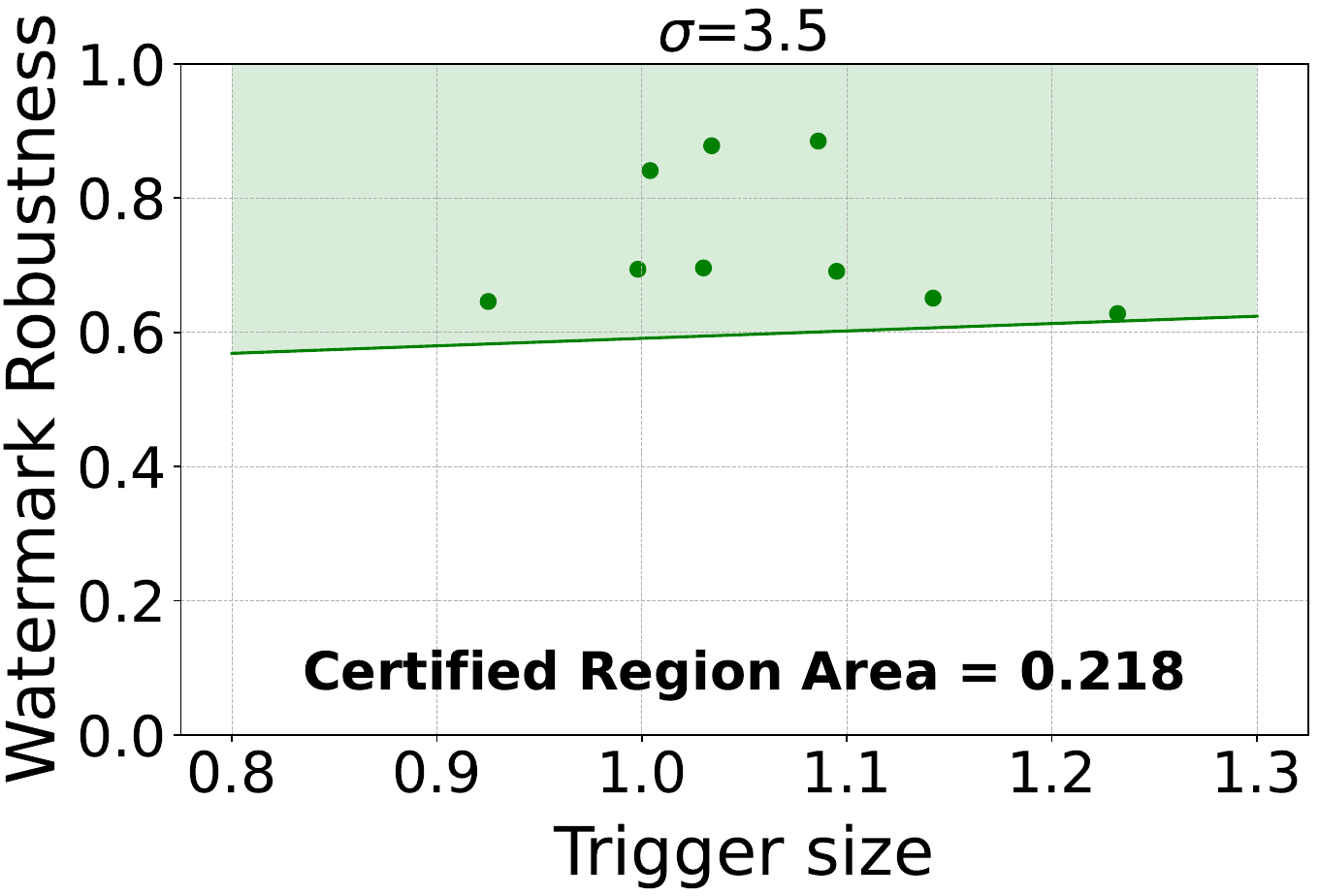}}
        \hspace{0.1em}
         \vspace{-1em}
	\caption{Examples of the certified regions (\ie, green shaded areas) obtained with Gaussian smoothing distribution (with different standard deviations $\sigma$) on GTSRB with two different ranges of trigger sizes $R$. Each green dot represents a watermarked instance characterized by its trigger size (horizontal axis) and watermark robustness (WR, vertical axis); those falling into the green region are guaranteed to be verified. \textbf{First Row}: $R \in (0.3,0.8)$; \textbf{Second Row}:  $R \in(0.8,1.3)$.}
 \label{fig:badnet_gtsrb}

\end{figure*}

\vspace{0.3em}
\noindent \textbf{Evaluation Metrics.} 
We evaluate the performance in two aspects: dataset watermarking and ownership verification. For dataset watermarking, we use benign accuracy (BA) and watermark success rate (WSR) to assess the effectiveness of the dataset watermarks. Specifically, BA is defined as the model accuracy on the benign testing dataset, and WSR is defined as the accuracy on the watermarked testing dataset. For ownership verification, we evaluate the verification success rate (VSR) and watermark certified accuracy (WCA). 
Specifically, VSR is defined at the model level as the proportion of watermarked models that are successfully verified as having been trained on the protected dataset.
In particular, the model-level verification proportion has two different interpretations depending on the type of evaluated models: 
\textbf{(1)} for watermarked models, VSR measures the proportion of watermarked models that are successfully verified, where a higher VSR indicates stronger stability of the learned watermark effect under noise perturbations; 
\textbf{(2)} for independent models, the corresponding model-level proportion is reported as the false positive rate (FPR), where we conservatively use their PP statistics for empirical false-positive evaluation, as described in Remark~\ref{remark:fpr_statistic} and Appendix~\ref{app:fpr_validation}. A lower FPR indicates that our method is less likely to falsely identify independent models as having been trained on the protected dataset. WCA is defined as the proportion of watermarked models that are guaranteed to be verified, \ie, the proportion of watermarked models whose $(R,\mathrm{WR})$ pairs fall into the certified region. The certified region is a two-dimensional region determined by inequality~(\ref{eq:gs_cp}), which involves watermark robustness and the magnitude of the trigger perturbation (see Figure~\ref{fig:badnet_gtsrb}). Within the certified region, the corresponding watermarks are guaranteed to be verified. 
Higher values of VSR and WCA on watermarked models, together with a lower FPR on independent models, indicate better performance of the verification method.
Formal model-level definitions of VSR, FPR, and WCA are provided in Appendix~\ref{app:evaluation_metrics}.
Besides, during the verification phase, we also evaluate WSR to further demonstrate the robustness of our watermarking method.

 \subsection{Main Results}
\label{sec:results}

As shown in Tables \ref{table:GTSRB}-\ref{table:CIFAR10}, our CertDW watermarking has only a mild influence on the utility of watermarked models. For instance, the watermark success rate (WSR) consistently remains above 90\%, while the accuracy of benign samples decreases by no more than 2\% (and in most cases, by less than 1\%). This indicates that our method does not interfere with the normal use of the dataset. During the verification process, our method achieves superior verification performance compared to other baseline methods. Unlike traditional approaches, our method remains robust even as noise levels increase, with WSR staying above 60\% in most cases. Although in some cases, the Vanilla method achieves a slightly higher VSR than our approach, its WCA is significantly lower. These experimental results strongly validate the effectiveness of our method.

Furthermore, we observe that the FPR computed over independent models for both datasets remains at
relatively low levels: the FPR for GTSRB is at most 12\%, while
for CIFAR-10, it is at most 6\%. These results further verify the
accuracy of our CertDW-based verification. Note that the FPR mentioned here
differs from the type-II error described in
Theorem~\ref{thm:General_condition}. Specifically, the type-II error in Theorem~\ref{thm:General_condition}
is a hypothesis-testing quantity that lower-bounds the
transformation-removed target-label stability, whereas the FPR is the
proportion of independent models incorrectly verified as having been
trained on the protected dataset. As such, these two metrics
represent distinct concepts and should not be conflated.

\subsection{Analyzing the Certified Region of CertDW}

In this section, we visualize the certified region of our method during the verification process and analyze the impact of trigger size on this region. Before presenting the results, we first explain the meaning of the 2-D certified region. The certified region is a 2-D area in the plane spanned by two axes: the horizontal axis represents the trigger size $R$ (\ie, $\|\boldsymbol{\delta}\|_2$), and the vertical axis represents the watermark robustness (WR). For a given noise level $\sigma$, the certified region is determined by the certified condition in Eq.~(\ref{eq:gs_cp}), corresponding to the green shaded area in Figure~\ref{fig:badnet_gtsrb}. Any watermarked model whose $(R,\mathrm{WR})$ pair falls into the certified region is guaranteed to be verified.

Specifically, we present examples of certified regions on the GTSRB obtained through Gaussian smoothing distribution with different standard deviations $\sigma \in \{1.5,2.5,3.5\}$, associated with two different ranges of trigger sizes (\ie, $R \in (0.3, 0.8)$ and $R \in (0.8, 1.3)$). As shown in Figure \ref{fig:badnet_gtsrb}, the shape of the certified region aligns with our theoretical results (see Remark \ref{remark:6}), which are derived from Example \ref{example:GS}. As the trigger size increases, both the number of watermarked models falling within the certified region and the certified region area gradually decrease, while the WR value increases. For instance, when the noise level is 1.5, the number of watermarked models falling within the certified region decreases from 13 to 3, and the certified region area decreases from 0.193 to 0.133, while the WR value increases from 58\% to 68\%. Additionally, as the noise level increases, the certified region area gradually improves. For example, when the trigger size is fixed at $(0.8, 1.3)$, and the noise increases from 1.5 to 3.5, the certified region area increases from 0.133 to 0.218. This indicates that watermarked models with higher WR and smaller trigger perturbation sizes are more likely to guarantee dataset ownership verification.

\begin{figure*}[!t]
    \begin{minipage}[t]{0.48\linewidth}
        \centering
	\subfigure[]{
		\includegraphics[width=0.95\linewidth]{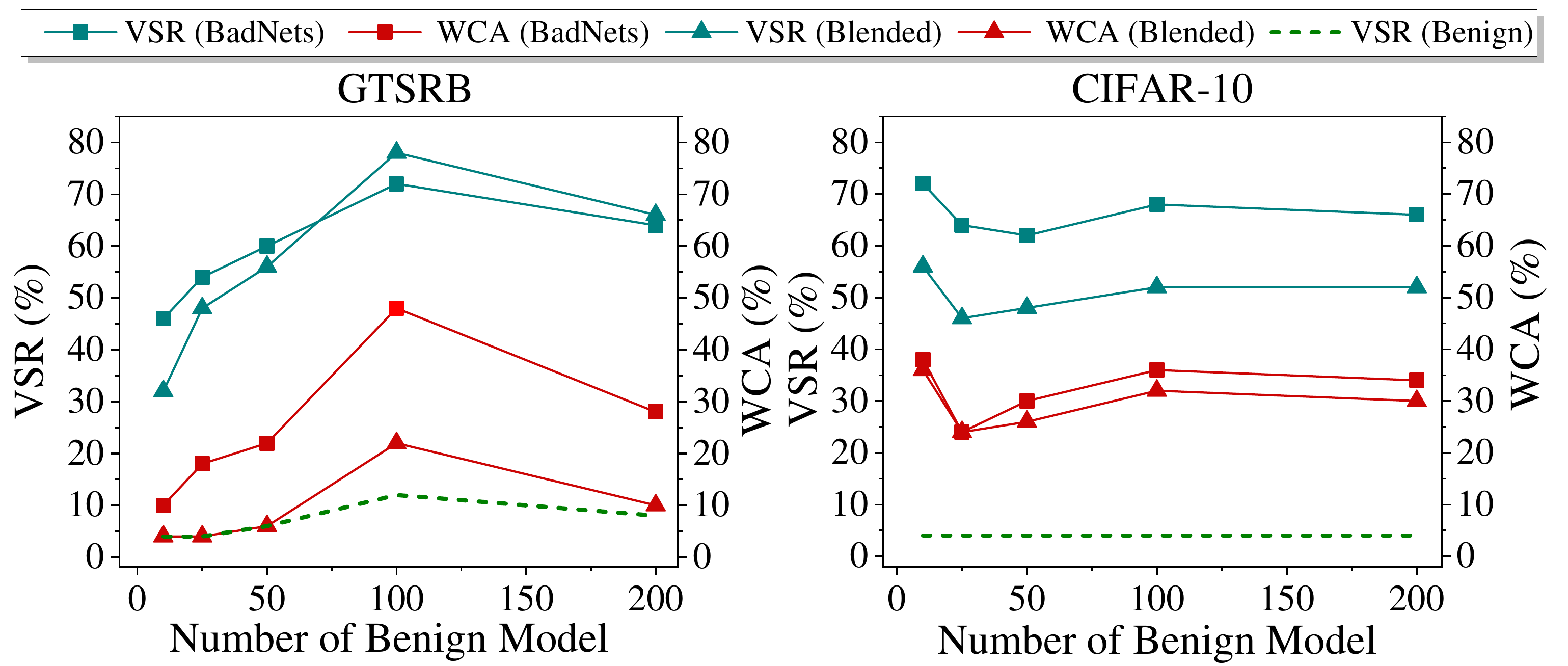}}\hspace{0.5em}
        \vspace{-0.8em}
	\caption{Effects of the number of benign models.}
     \label{fig:validation}
    \end{minipage}\hspace{0.5em}
    \begin{minipage}[t]{0.48\linewidth}
        \centering
	\subfigure[]{
		\includegraphics[width=0.95\linewidth]{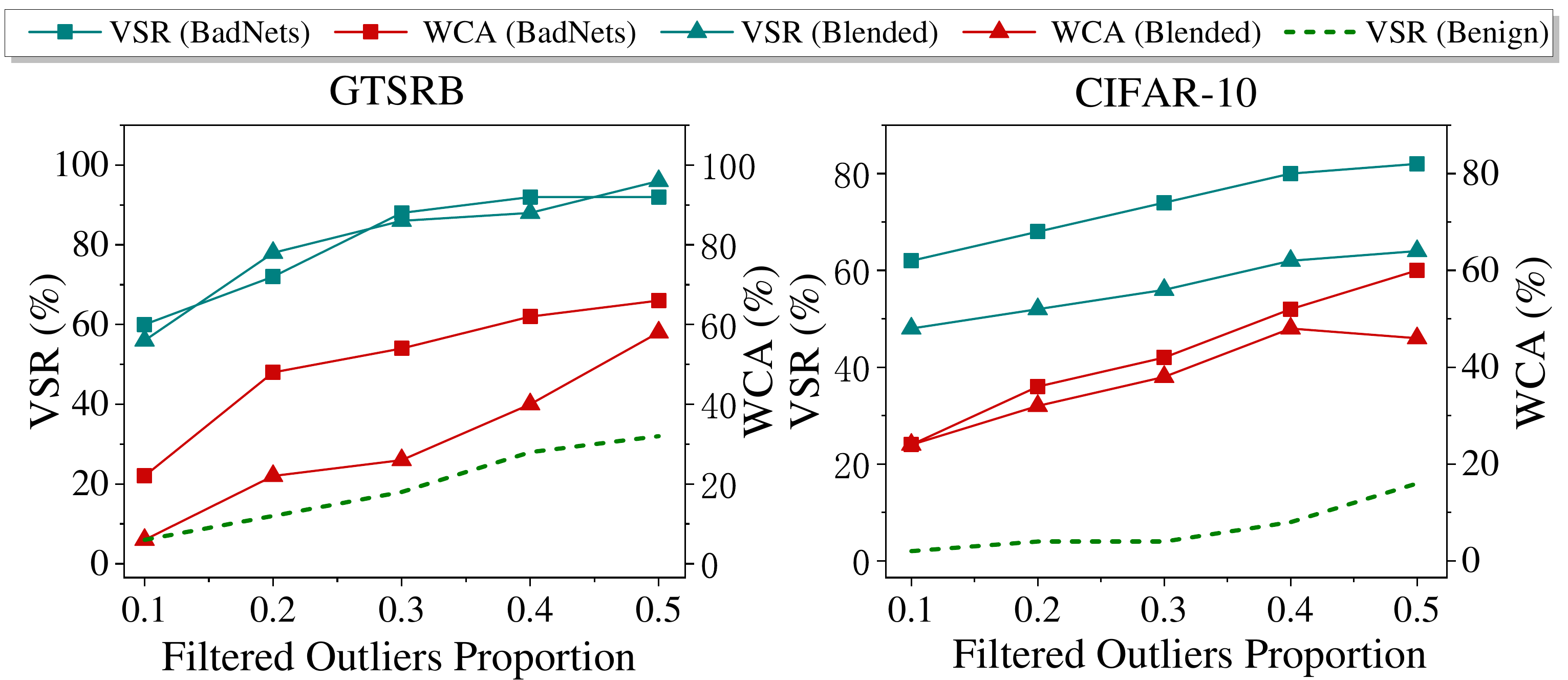}}\hspace{0.5em}
        \vspace{-0.8em}
	\caption{Effects of the proportion of filtered outliers.}
 \label{fig:kappa}
    \end{minipage}%
\end{figure*}

\begin{figure*}[!t]
    \begin{minipage}[t]{0.48\linewidth}
        \centering
	\subfigure[]{
		\includegraphics[width=0.95\linewidth]{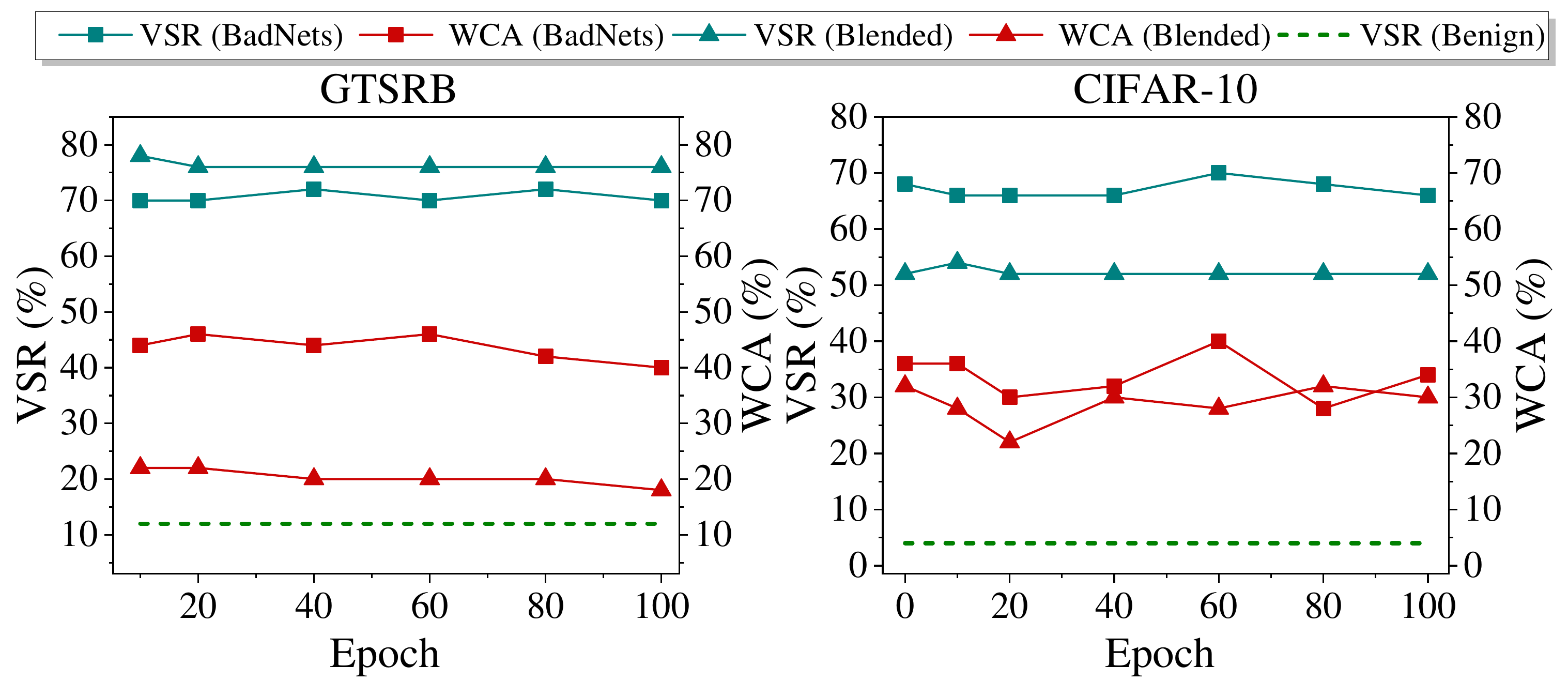}}\hspace{0.5em}
        \vspace{-0.8em}
	\caption{The resistance to fine-tuning.}
     \label{fig:fine-tuning}
    \end{minipage}\hspace{0.5em}
    \begin{minipage}[t]{0.48\linewidth}
        \centering
	\subfigure[]{
		\includegraphics[width=0.95\linewidth]{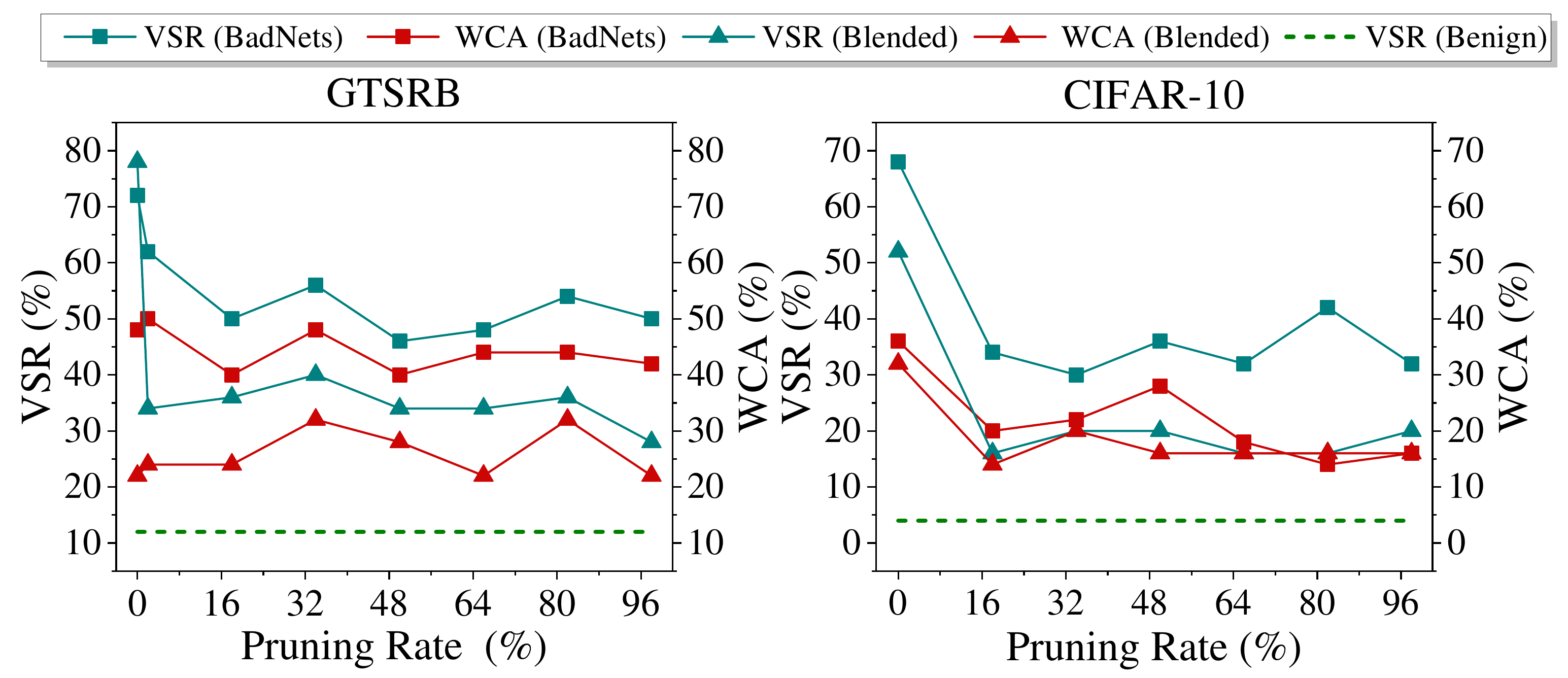}}\hspace{0.5em}
        \vspace{-0.8em}
	\caption{The resistance to model pruning.}
 \label{fig:pruning}
    \end{minipage}%
    \vspace{-1em}
\end{figure*}

\subsection{Ablation Study}

We hereby discuss the impact of key hyper-parameters. For simplicity, we discuss each dataset by using a fixed noise level (\eg, 2.5 on GTSRB and 1.2 on CIFAR-10). Additional ablation studies are provided in Appendix~\ref{app:Ablation}.

\subsubsection{Impact of the Number of Benign Models}

As shown in Figure \ref{fig:validation}, both the VSR and WCA increase with the number of benign models $J$. These results indicate that defenders can enhance the confidence in verification by increasing $J$. Particularly, when the number of benign models reaches 100, both VSR and WCA reach relatively high values. However, in most cases, excessive increases in the number of benign models lead to decreased VSR and WCA. This may be due to the fact that our benign models are trained using a dataset instead of the actual data distribution, and the data is not sufficient. As the number of models increases, this insufficiency becomes more pronounced, resulting in anomalously high PP values. In other words, dataset owners should determine the number of benign models based on their specific requirements.

\subsubsection{Impact of the Proportion of Filtered Outliers}

As shown in Figure \ref{fig:kappa}, both the VSR and the WCA increase as the outlier filtering ratio $\kappa$ increases. However, in most cases, the VSR of independent models (\ie, FPR) also increases, meaning that the likelihood of incorrectly identifying an independent model as being trained on the protected dataset also rises. In other words, there is a trade-off between precision and recall here to some extent. In practice, dataset owners should also determine the value of $\kappa$ based on their specific needs.

\subsection{The Resistance to Potential Adaptive Attacks}

In this section, we discuss the resistance of our method against two potential watermark-removal attacks, including fine-tuning \cite{liu2017neural} and model pruning \cite{liu2018fine}. Additional analyses on STRIP-based detection~\cite{gao2021design} and NAD-based model repair~\cite{li2021neural} are provided in Appendix~\ref{app:adaptive_countermeasures}.

\vspace{0.3em}
\noindent \textbf{The Resistance to Fine-tuning.}  
Following the previous work \cite{liu2017neural}, we fine-tune all layers of the CertDW-watermarked model use 10\% of the benign samples from the original training set, with a learning rate of 0.001. The model is fine-tuning for a total of 100 epochs. As shown in Figure \ref{fig:fine-tuning}, throughout the fine-tuning process, both VSR and WCA remain stable to a large extent. These results indicate that fine-tuning only has a minor impact on our method.

\vspace{0.3em}
\noindent \textbf{The Resistance to Model Pruning.} 
Following the previous work \cite{liu2018fine}, we use 10\% of the benign samples from the original training set to prune the latent representations of our watermarked model (\ie, the inputs to the fully connected layers). The pruning rates vary from 0\% to 98\% in each case. As shown in Figure \ref{fig:pruning}, pruning initially leads to a significant drop in the VSR and WCA. However, subsequent changes become relatively stable, especially as the pruning rate approaches 98\%, where the metrics show minimal further variation. Thus, even under high pruning rates, our method maintains certain levels of VSR and WCA, demonstrating its resilience to model pruning to some extent.

\begin{figure*}[!t]
    \centering
    \subfigure[GTSRB (BadNets)]{
		\includegraphics[width=0.238\textwidth]{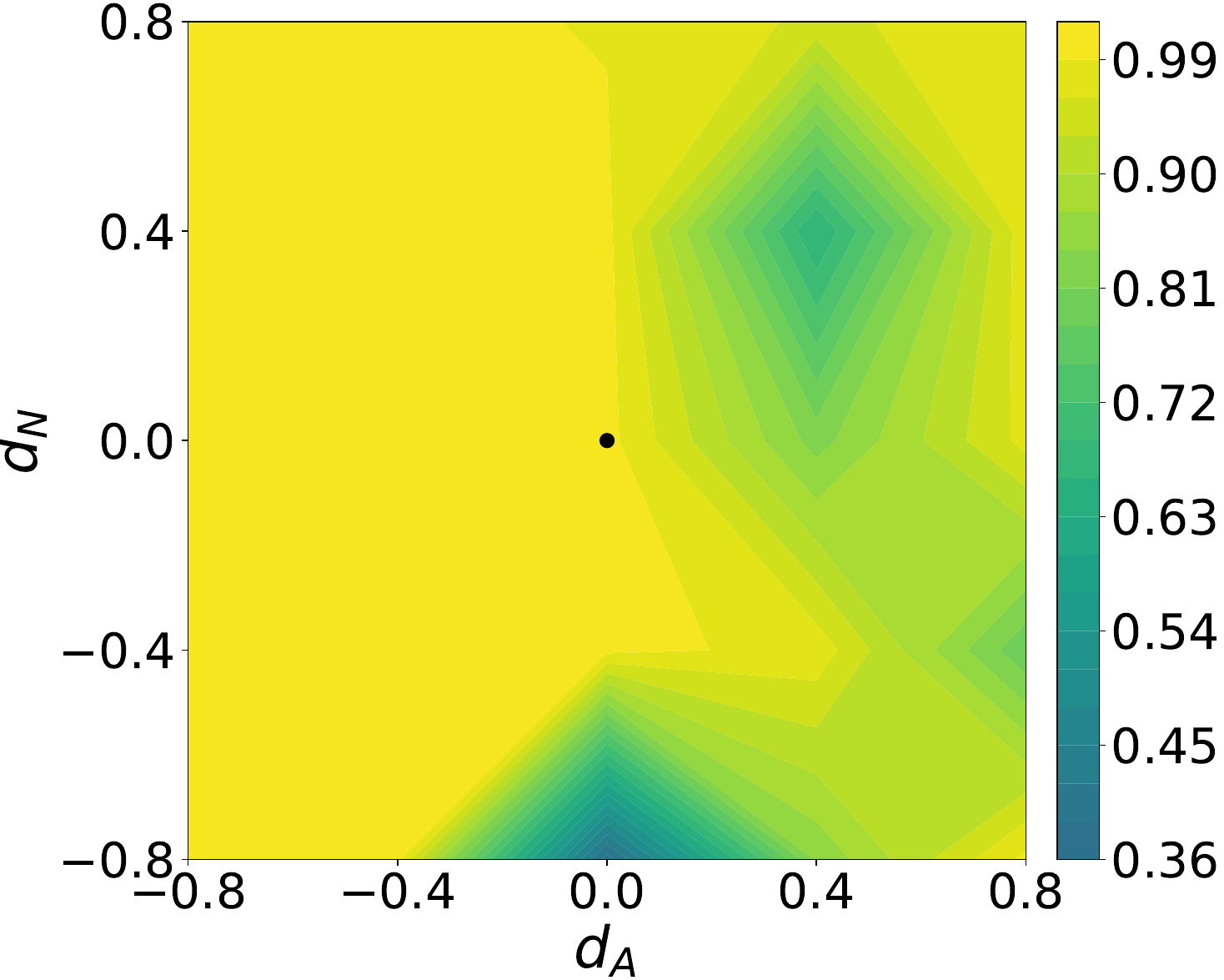}}
	\hspace{0.1em}
    \subfigure[CIFAR-10 (BadNets)]{
		\includegraphics[width=0.24\textwidth]{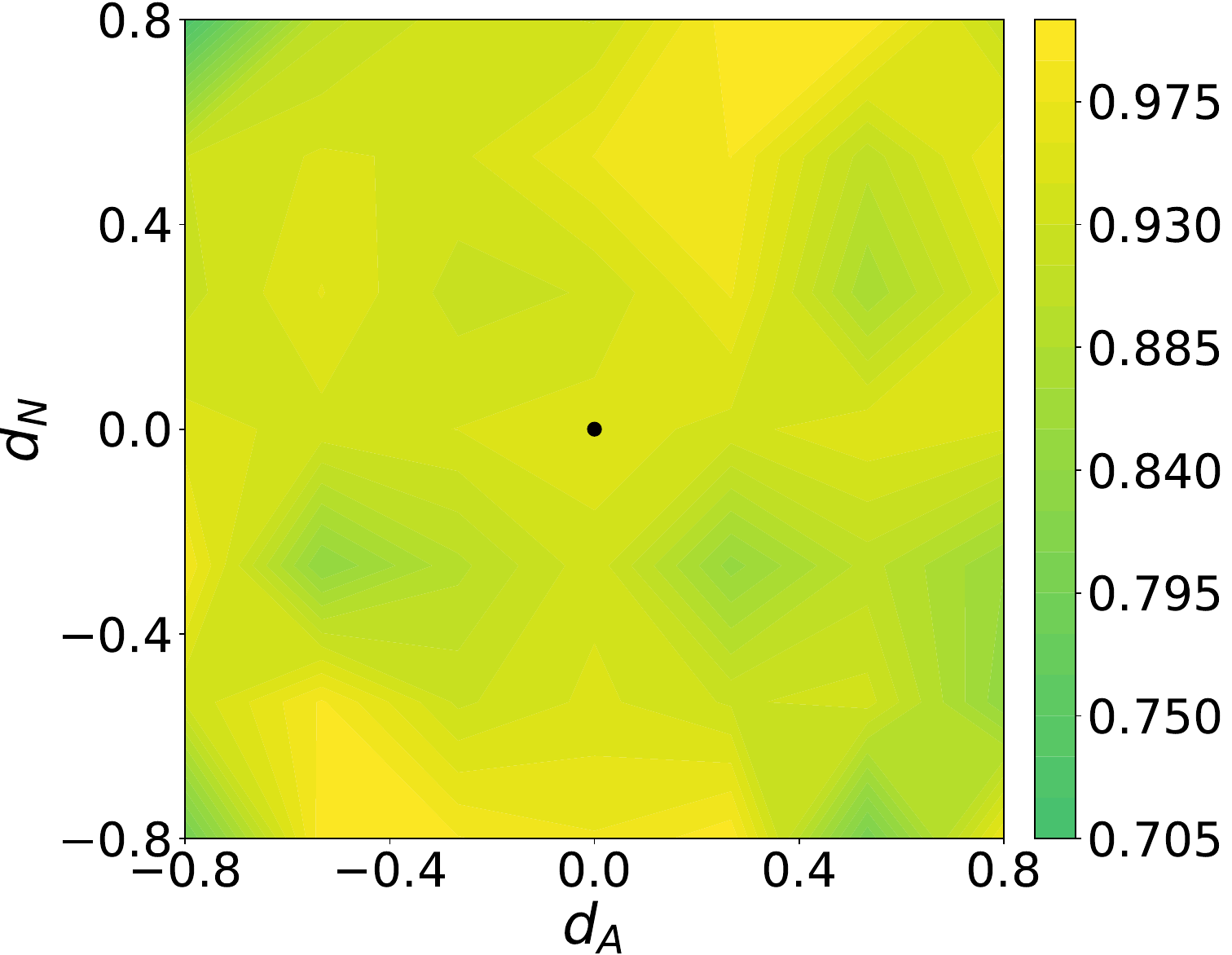}}
    \hspace{0.1em}
	\subfigure[GTSRB (Blended)]{
		\includegraphics[width=0.235\textwidth]{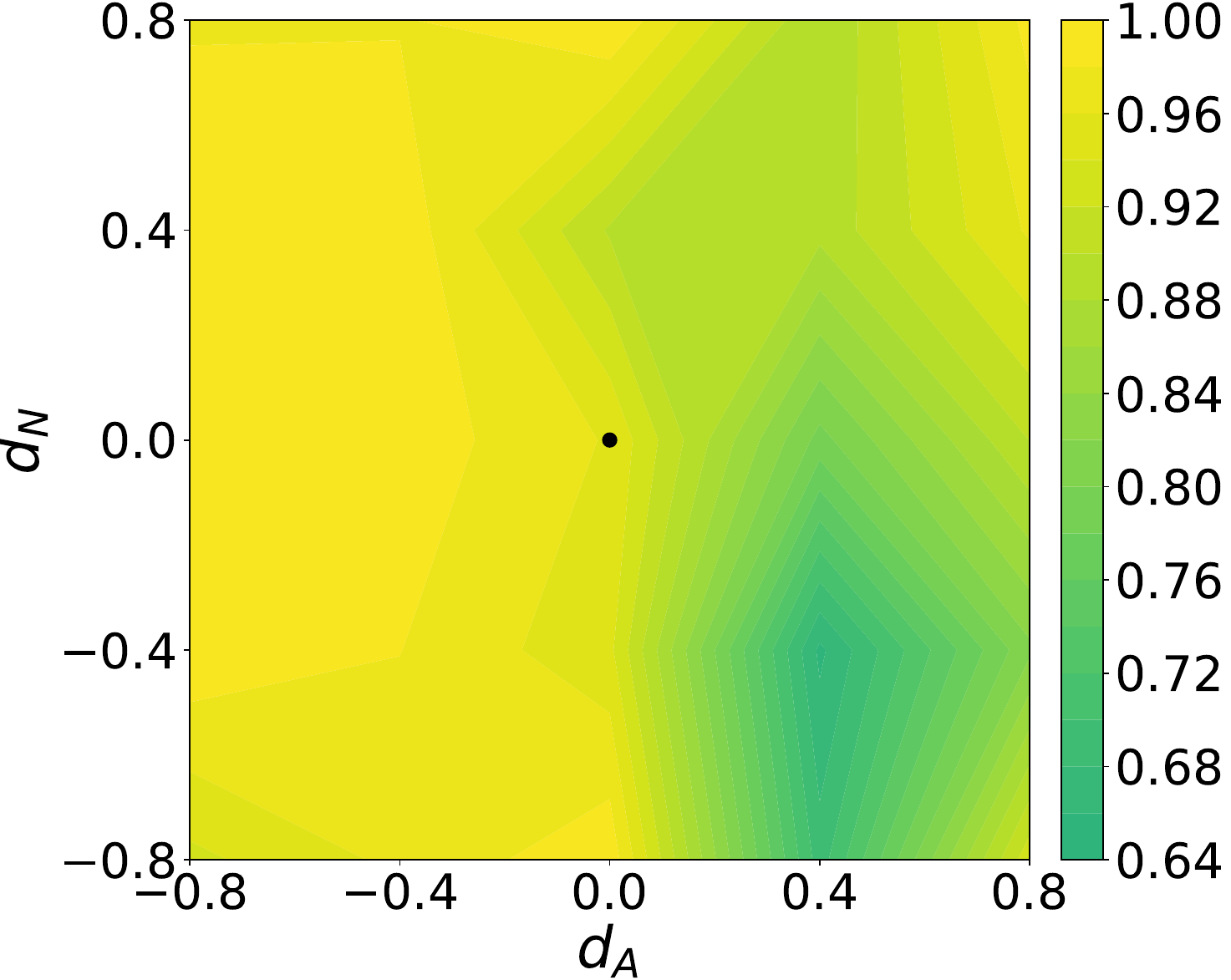}}
        \subfigure[CIFAR-10 (Blended)]{
		\includegraphics[width=0.235\textwidth]{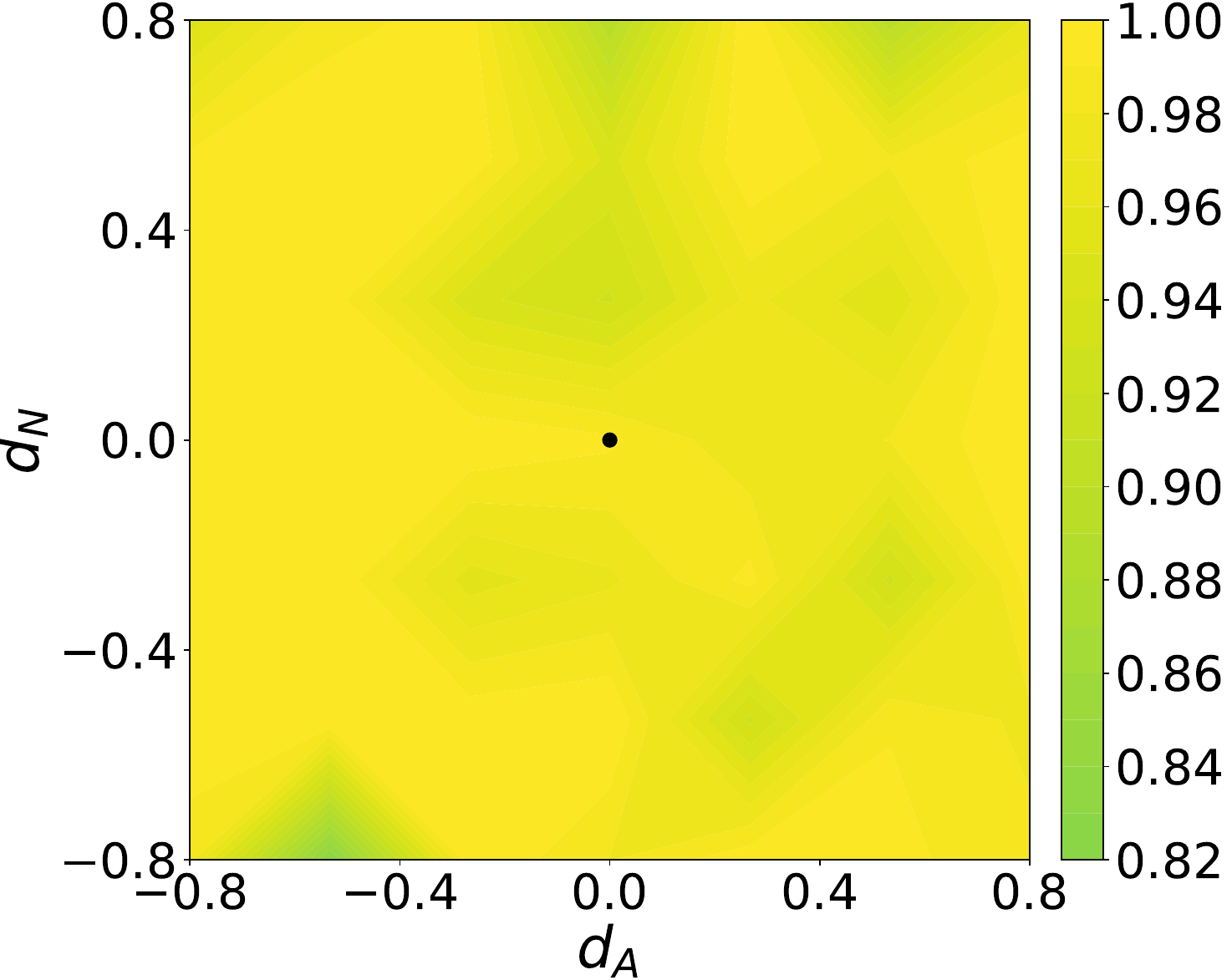}}
	\hspace{0.1em}
    \caption{The performance (WSR) of our method's watermarked samples in the sample space. $d_N$ is the random noise direction, and $d_{A}$ is the adversarial direction. `$\bullet$' denotes the original watermarked sample. The first two columns show results for BadNets-based watermarks, and the last two columns show results for Blended-based watermarks.
    }
    \label{fig:noise_adv_ourmodel}
\end{figure*}

\begin{figure}[!t]
    \centering
    \vspace{-1em}
    \subfigure[VSR]{
		\includegraphics[width=0.238\textwidth]{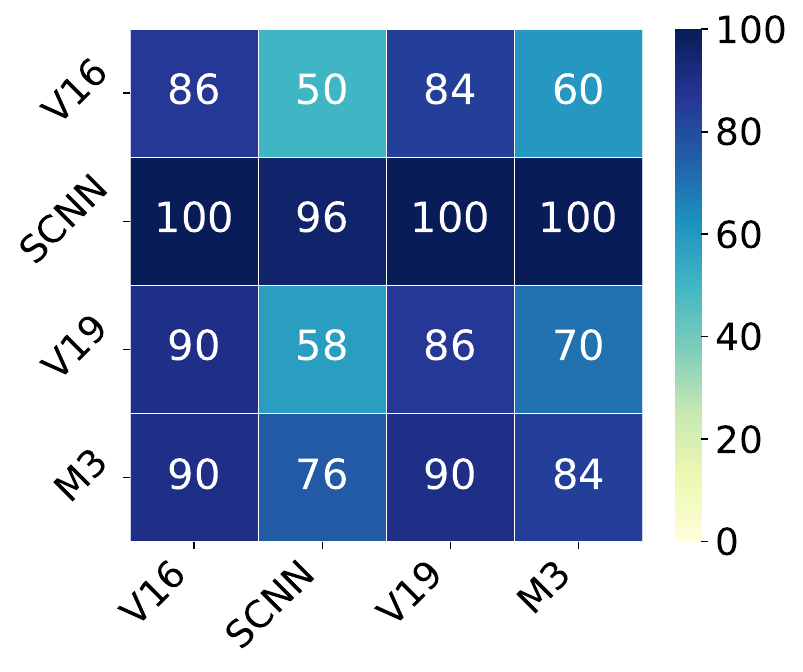}
        \label{model_VSR}}
    \hspace{-1.0em}
    \subfigure[WCA]{
		\includegraphics[width=0.238\textwidth]{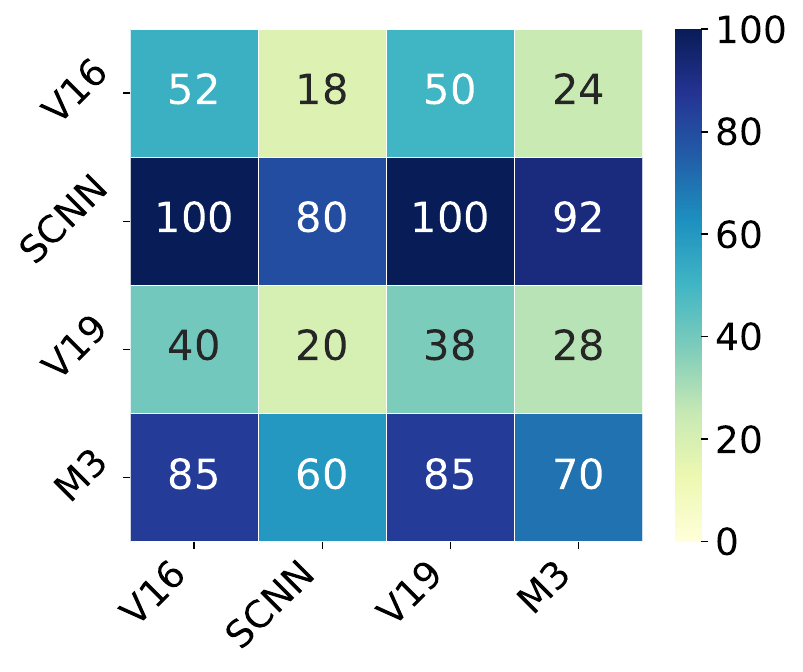}
        \label{model_WCA}}
	\vspace{-0.8em}
    \caption{The performance of our method with different structures of benign and watermarked models on GTSRB. Row: benign models; Column: watermarked models.}
    \label{fig:model_transfer}
    	\vspace{-0.8em}
    \end{figure}

\subsection{Model-level Transferability of CertDW}

As described in Section \ref{sec:com_PP}, we use a pre-trained surrogate model to serve as the benign model. The experiments in Section \ref{sec:results} are conducted based on the setting that the benign model and the watermarked model share the same architecture. However, this assumption may not hold in practice, as dataset owners are typically unaware of the specific architectures exploited by dataset users (including adversaries). Accordingly, in this section, we analyze the effectiveness of our CertDW when benign and watermarked models have different architectures.

Specifically, we select four representative network architectures, including VGG-16 (dubbed `V16'), spacial-enhanced CNN (dubbed `SCNN'), VGG-19 (dubbed `V19'), and MobileNetV3 (dubbed `M3'), on the GTSRB dataset for discussion. All other experimental configurations strictly follow those described in Sections \ref{Exper_setup} and Section \ref{sec:results}, including 50 watermarked models and 100 benign models for each architecture combination. As shown in Figure \ref{fig:model_transfer}, our method remains effective across different model architectures, despite some performance fluctuations due to different model capacity. These results demonstrate that our CertDW does not rely on prior knowledge of the benign and watermarked model, making it a robust and practical approach for dataset ownership verification.

 \subsection{A Closer Look to the Effectiveness of CertDW}

In this section, we intend to further explore the mechanisms behind the effectiveness of our CertDW. Specifically, we visualize the region around the watermarked samples in the sample space and feature space for in-depth discussion.

\begin{figure*}[!t]
    \begin{minipage}[t]{0.48\linewidth}
        \centering
	\subfigure[]{
		\includegraphics[width=0.45\linewidth]{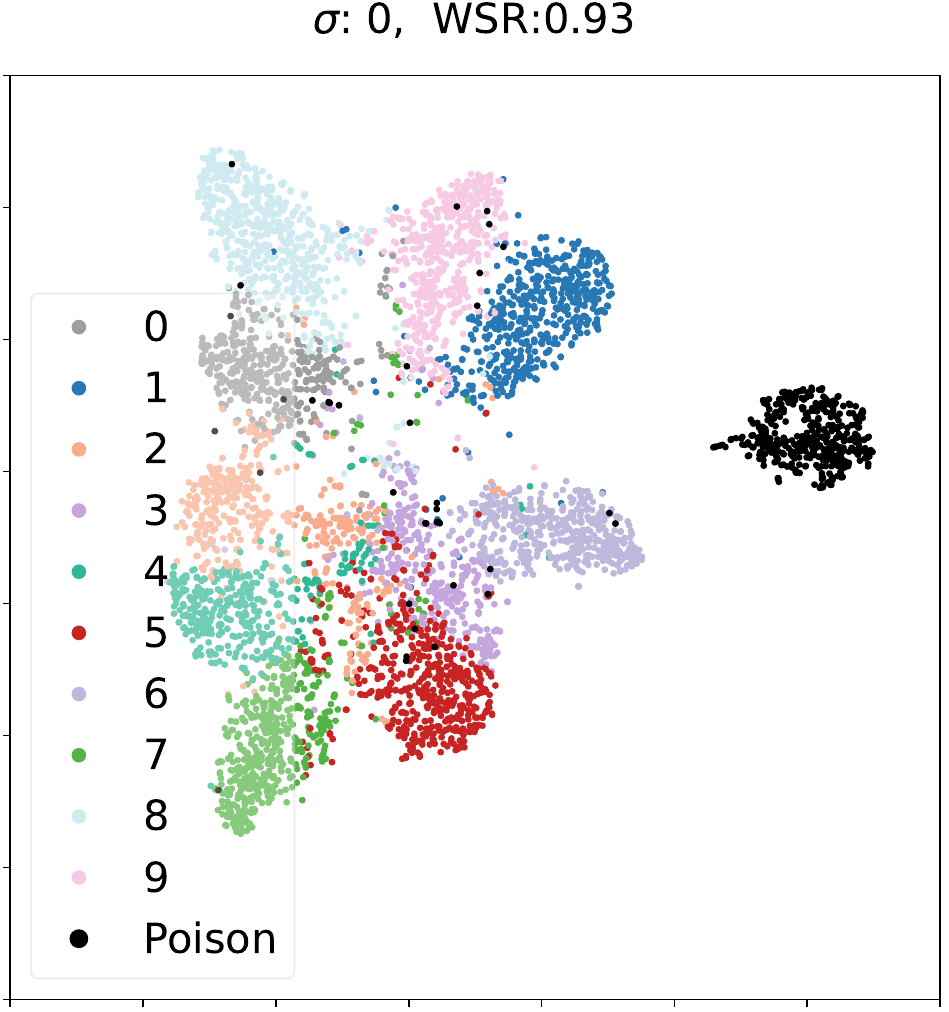}}\hspace{1em}
        \vspace{0.1em}
        \subfigure[]{
		\includegraphics[width=0.45\linewidth]{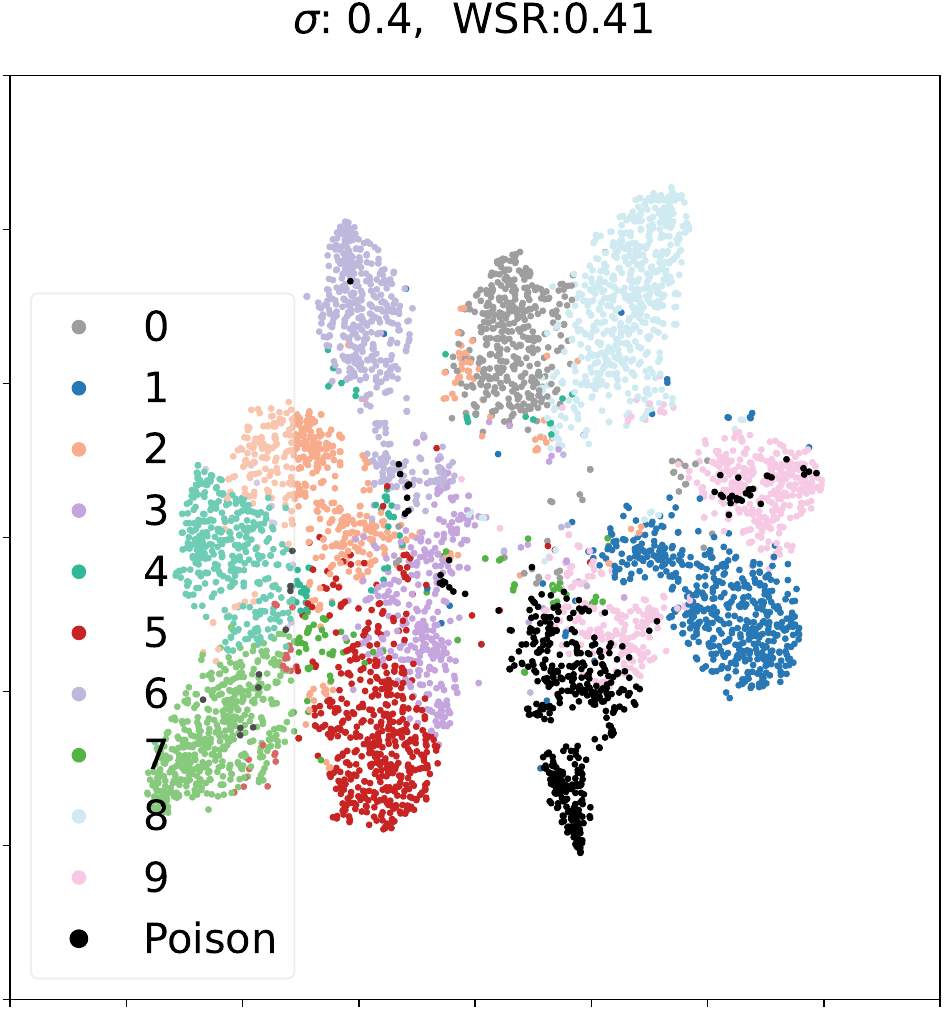}}
        \vspace{0.1em}
	\caption{t-SNE visualization of feature distribution in the vanilla watermarked model with random noises.}
     \label{fig:noise_tsne_vanillia}
    \end{minipage}\hspace{2em}
    \begin{minipage}[t]{0.48\linewidth}
        \centering
	\subfigure[]{
		\includegraphics[width=0.45\linewidth]{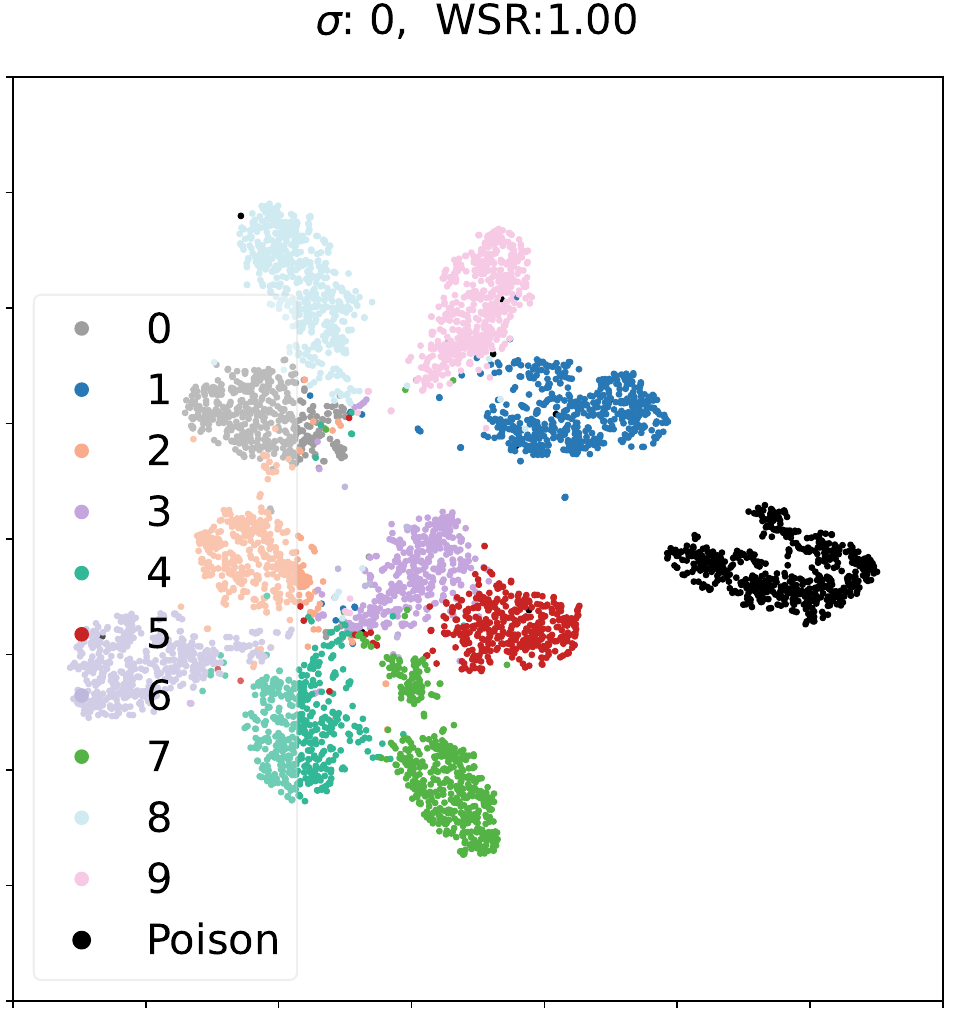}}\hspace{1em}
        \vspace{0.1em}
        \subfigure[]{
		\includegraphics[width=0.45\linewidth]{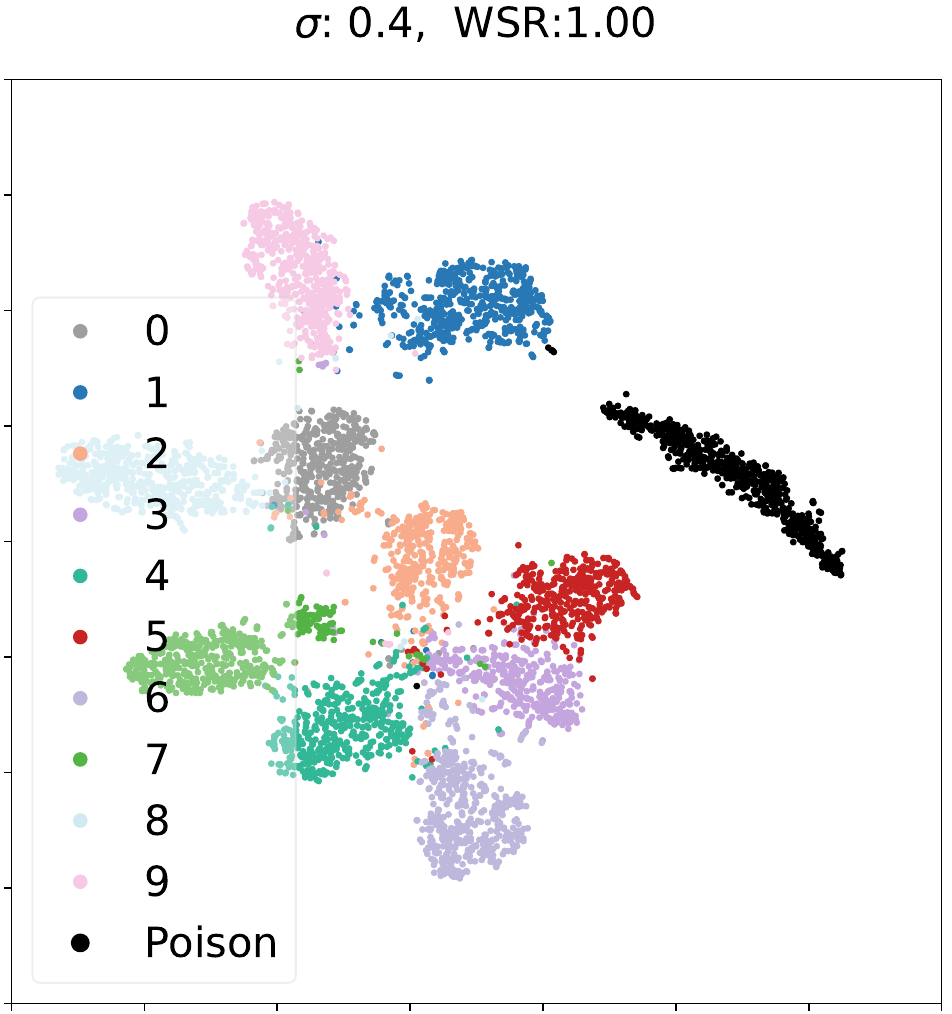}}
        \vspace{0.1em}
	\caption{t-SNE visualization of feature distribution in the our watermarked model with random noises.}
 \label{fig:noise_tsne_our}
    \end{minipage}%
\end{figure*}

\begin{figure*}[!t]
    \begin{minipage}[t]{0.48\linewidth}
        \centering
	\subfigure[]{
		\includegraphics[width=0.45\linewidth]{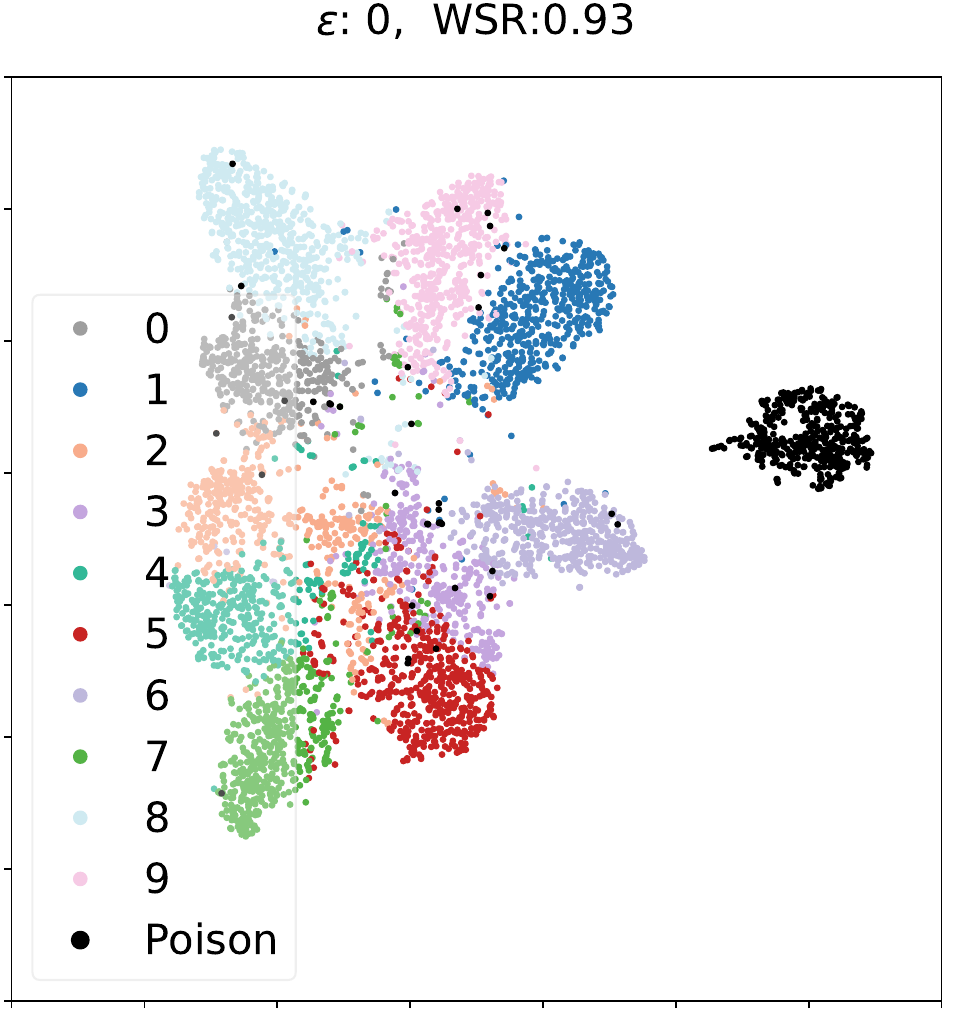}}\hspace{0.5em}
        \vspace{0.1em}
        \subfigure[]{
		\includegraphics[width=0.45\linewidth]{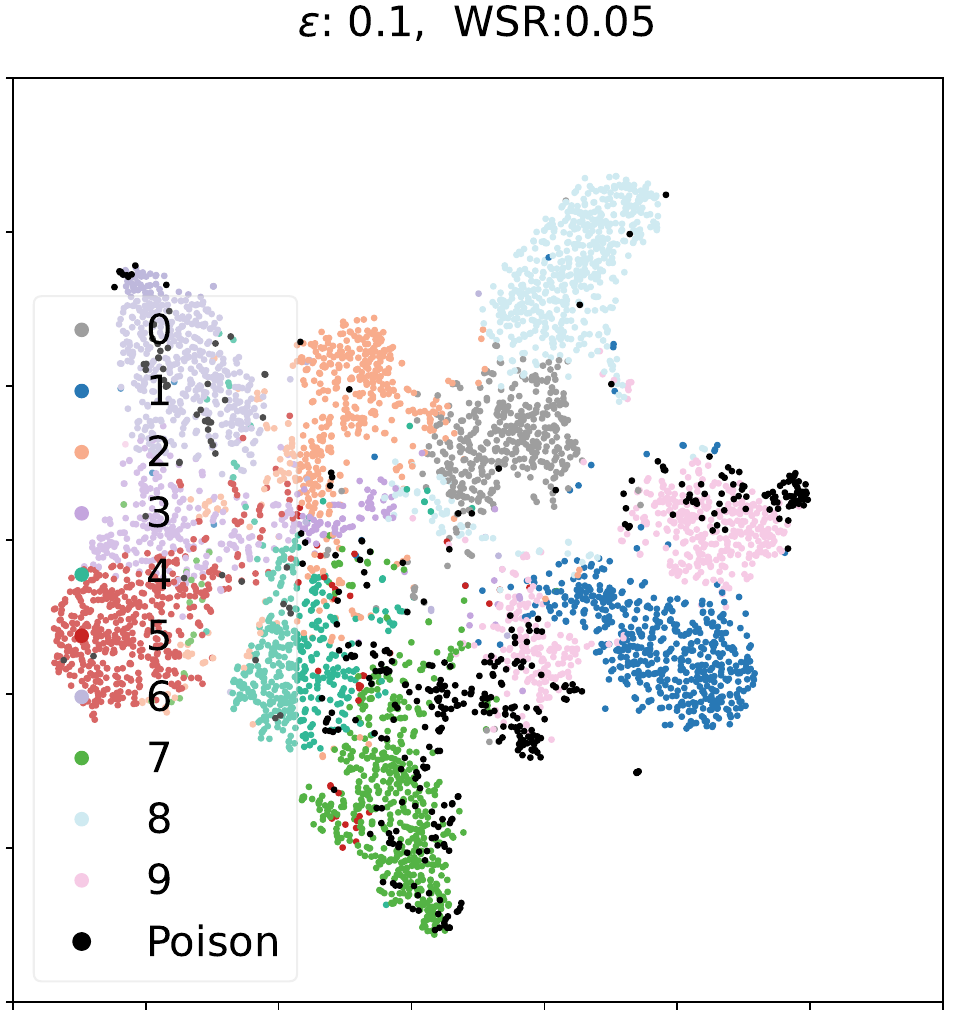}}
        \vspace{0.1em}
        \looseness=-1
	\caption{t-SNE visualization of feature distribution in  vanilla watermarked model with adversarial perturbations.}
     \label{fig:adv_tsne_vanillia}
    \end{minipage}\hspace{2em}
    \begin{minipage}[t]{0.48\linewidth}
        \centering
	\subfigure[]{
		\includegraphics[width=0.45\linewidth]{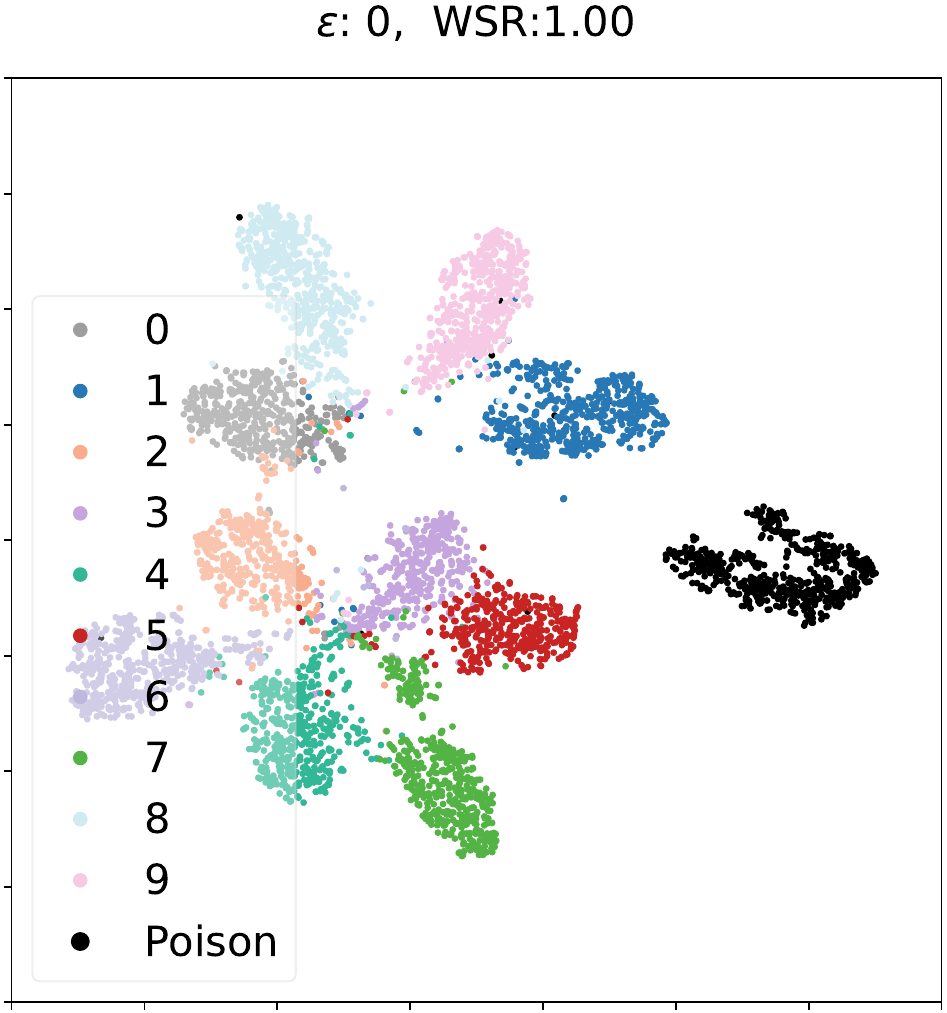}}\hspace{0.5em}
        \vspace{0.1em}
        \subfigure[]{
		\includegraphics[width=0.45\linewidth]{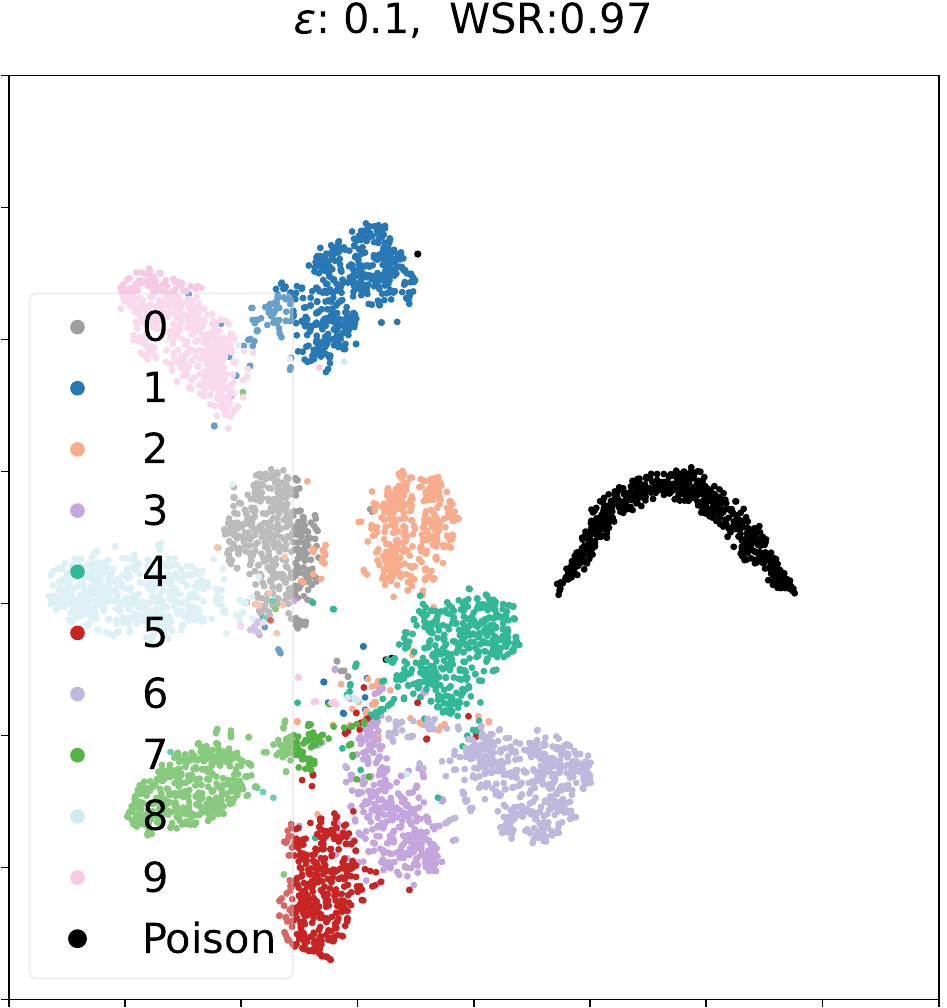}}
        \vspace{0.1em}
	\caption{t-SNE visualization of feature distribution in the our watermarked model with adversarial perturbations.}
 \label{fig:adv_tsne_our}
    \end{minipage}%
\end{figure*}

\subsubsection{Visualizing in the Sample Space}

In this section, we use the same method as in Section \ref{sec:impact_adv} to demonstrate the effectiveness of our approach. As shown in Figure  \ref{fig:noise_adv_ourmodel}, we find that our watermarked samples can maintain a high WSR (over 80\%) in the direction of unintentional random noise. Although there is some performance degradation in the direction of intentional adversarial perturbations, our method still exhibits strong robustness and effectively prevents verification failure. For example, on the CIFAR-10 dataset, even with perturbation magnitudes of 0.8 in both directions, the WSR remains above 70\%. This result significantly increases the difficulty for an adversary to intentionally add perturbations to verification/watermarked samples to completely remove the watermark.

 \subsubsection{Visualizing the Feature Space}

 To better understand our method's effectiveness, we adopt t-SNE \cite{van2008visualizing} to visualize the feature distribution of watermarked samples evolves with unintentional noise and intentional adversarial perturbations.

\vspace{0.3em}
\noindent  \textbf{Features along with the Unintentional Random Noises.} 
We visualized the impact of different random noise magnitudes at the early stages. As shown in Figure \ref{fig:noise_tsne_vanillia}, at the initial stage of adding random noise, the representations of the watermarked samples quickly become very similar to those of the benign samples, leading to a significant reduction in the watermarking success rate. However, our method effectively maintains the watermarked samples within an independent cluster, ensuring a clear separation from the non-target clusters, as depicted in Figure \ref{fig:noise_tsne_our}.

\vspace{0.3em}
\noindent \textbf{Features along with the Intentional  Adversarial Perturbations.} 
To further demonstrate how hidden representations evolve along adversarial directions, we add adversarial perturbations of varying magnitudes to the watermarked samples. As shown in Figure \ref{fig:adv_tsne_vanillia}, with minor perturbations, the representations of the watermarked samples quickly blend with those of the benign samples. In contrast, our method successfully maintains the watermarked samples in a distinct cluster, which remains distinctly separate from non-target clusters, as depicted in Figure \ref{fig:adv_tsne_our}.

\subsection{The Analysis of Computational Complexity}
\label{sec:Com_Complexity}
In this section, we analyze the computational complexity of our CertDW, especially its dataset watermarking and ownership verification process, respectively.

\vspace{0.3em}
\noindent  \textbf{The Complexity of Dataset Watermarking.} Let $N$ denote the number of all training samples, and $\gamma$ is the poisoning rate. Our computational complexity is $\mathcal{O}(\gamma \cdot N)$, since CertDW only needs to watermark a small subset of selected samples in this step. In general, the trigger size of these watermarks must satisfy the $\ell_2$ norm constraint by adjusting pixel-level perturbation size, which is highly efficient.

\vspace{0.3em}
\noindent  \textbf{The Complexity of Dataset Verification.} Let $J$, $W_a$, $I_n$ denote the number of benign models, watermarked models, and independent models, respectively. Dataset owner trains these models with computational complexities of $\mathcal{O}(J)$, $\mathcal{O}(W_a)$, and $\mathcal{O}(I_n)$, respectively. This process supports parallel processing. Additionally, we use the trained multiple benign models and watermarked models to calculate the PP value and WR value, which is highly efficient. For example, training benign models and watermarked models takes approximately 120 seconds and 180 seconds on the CIFAR-10 dataset, respectively. Computing PP and WR values requires only 3 seconds and 2 seconds, respectively. As such, calculating PP values and WR values is almost cost-free. Arguably, although training each benign model is time-consuming, it is generally acceptable, not to mention that we can use parallel computing to further accelerate it.

\section{Potential Limitations and Future Work}
\label{sec:limitations}

As the pioneering work on certified dataset ownership verification, we acknowledge several potential limitations that deserve future investigation.

Firstly, the theoretical analysis of CertDW is primarily based on $\ell_2$-norm bounded perturbations. In future work, we will further generalize our certified framework to broader perturbation norms (\eg, $\ell_\infty$, $\ell_0$) and to perceptual or semantic distances that better reflect realistic threats.

Secondly, CertDW requires training a set of benign models (\ie, $J=100$) to construct the calibration set, which incurs a certain amount of computational overhead. As detailed in Section~\ref{sec:Com_Complexity}, although this cost is largely controllable relative to the achieved certified performance gains, it is not negligible. We will further explore more efficient calibration strategies to further reduce the calibration cost.

Thirdly, the current certified bounds of CertDW are derived under worst-case assumptions, which may lead to overly conservative verification thresholds in some configurations. In future work, we plan to tighten the certified bounds through a more fine-grained theoretical analysis.

Finally, CertDW is currently evaluated primarily on image classification tasks and has not yet been extended to other tasks and modalities, such as object detection, semantic segmentation, text, speech, 3D point clouds, and generative models. In future work, we plan to extend CertDW to a wider range of tasks and modalities, and evaluate its effectiveness under more complex scenarios.

\section{Conclusion}

In this paper, we revisited existing dataset ownership verification (DOV) methods and revealed that their performance degrades sharply under both unintentional random noise and intentional adversarial perturbations. Based on our analysis, we proposed a certified dataset watermark (\ie, CertDW) to provide robustness guarantees for dataset ownership verification. Inspired by conformal prediction, we introduced two statistical measures, \ie, principal probability (PP) and watermark robustness (WR),  
to characterize the prediction behaviors of benign and watermarked
samples under noise perturbations, respectively. 
We further derived sufficient certification conditions that relate WR to the calibration threshold constructed from benign-model PP statistics.
We calculated the PP and WR values by introducing random noise to multiple benign and watermarked samples. As long as a suspicious model's WR is sufficiently larger than a proportion of multiple PP values calculated by several benign models, we can conclude that the suspicious model is trained on the protected dataset, with a provable high-probability bound on the false positive rate. 
Extensive experiments on benchmark datasets validate CertDW's effectiveness and its resilience to potential adaptive attacks. We hope our paper can provide a new perspective on reliable dataset ownership verification, to facilitate more trustworthy dataset sharing and circulation.


\bibliographystyle{IEEEtran}

\bibliography{main}

\newpage

 \appendix 

 \setcounter{theorem}{0}
 \setcounter{corollary}{0}
\setcounter{example}{0}
 
	\appendices

\section{Algorithmic Description of CertDW}
\label{app:certdw}

In this section, we provide the algorithmic description of our CertDW.
Algorithm~\ref{alg:certdw} summarizes the overall verification process, and
Algorithm~\ref{alg:certdw-cert} summarizes the certified process under the
Gaussian smoothing distribution.
The smoothing probabilities in these algorithms are estimated using $M$ Monte Carlo samples. The corresponding finite-$M$ certification procedure based on simultaneous one-sided lower confidence bounds is provided in
Appendix~\ref{app:finite_mc}.

\begin{algorithm}[!t]
\caption{The main process of our CertDW.}
\label{alg:certdw}
\begin{algorithmic}[1]
\State \textbf{Input:} 
The suspicious model $f(\cdot; \vtheta)$ (to be verified); 
$J$ benign models $\{g_j(\cdot; \vw_j)\}_{j=1}^{J}$ independently trained on watermark-free datasets; 
trigger pattern $\vdelta$ and target label $\hat{y}$; 
noise distribution $\pno$; number of Monte Carlo samples $M$; 
outlier ratio $\kappa$; calibration parameter $\alpha_0$
\State \textbf{Output:} Ownership verification result (True / False)
\State \textbf{Step I: Computing Principal Probability (PP)}
\For{$j = 1$ \textbf{to} $J$}
    \State Sample $K$ correctly-predicted benign samples $\{\vx_k\}_{k=1}^K$, one per class, using $g_j$
    \For{$k = 1$ \textbf{to} $K$}
        \State Add $M$ random noises $\{\vepsilon_i\}_{i=1}^M \sim \pno$ to $\vx_k$, query $g_j$, and estimate PD $p(\vx_k | g_j, \pno)$ via Eq.~(\ref{eq:DP_MC})
    \EndFor
     \State Compute the principal probability (PP) of $g_j$ by averaging PDs and taking the $\ell_\infty$ norm following Eq.~(\ref{eq:PP})
\EndFor
    
\State Construct the calibration set $P_C = \{P_C^j\}_{j=1}^{J}$ and filter the largest $\lfloor \kappa \cdot J \rfloor$ outliers.
\State \textbf{Step II: Calculating Watermark Robustness (WR)}
\State Sample $K$ correctly-predicted benign samples $\{\vx_k\}_{k=1}^K$, one per class
\State Construct watermarked samples $\hat{\vx}_k = \vx_k + \vdelta$ for $k = 1, \cdots, K$
\For{$k = 1$ \textbf{to} $K$}
    \State Add $M$ random noises $\{\vepsilon_i\}_{i=1}^M \sim \pno$ to $\hat{\vx}_k$, query $f$, and estimate $p_{\hat{y}}(\hat{\vx}_k | f_{\vtheta}, \pno)$ via Eq.~(\ref{eq:DP_MC})
\EndFor
\State Compute the watermark robustness (WR) of the suspicious model $f(\cdot; \vtheta)$ on watermarked samples by taking the minimum target-label probability across $K$ watermarked samples, one per class, following Eq.~(\ref{eq:WR})
\State \textbf{Step III: Ownership Verification via Conformal Calibration}
\State Compute the conformal calibration statistic $\rho$ based on WR and the calibration set $\mathcal{P}_C$ following Eq.~(\ref{eq:p-value})
\State Verify dataset ownership via Proposition~\ref{pro:conformal_prediction}: 
    \textbf{return} \texttt{True} if $\rho \geq 1 - \alpha_0$,   
    else \textbf{return} \texttt{False}
\end{algorithmic}
\end{algorithm}

\begin{algorithm}[!t]
\caption{The certified process of our CertDW (under Gaussian smoothing).}
\label{alg:certdw-cert}
\begin{algorithmic}[1]
\State \textbf{Input:} 
The watermarked model $f(\cdot; \vtheta)$; 
trigger pattern $\vdelta$ with perturbation magnitude $R = \max_k \|\vr_k\|_2$ and target label $\hat{y}$; 
the calibration set $\mathcal{P}_C$ obtained from $J$ benign models (\textbf{Step~I} of Algorithm~\ref{alg:certdw}); 
$K$ correctly-predicted benign samples $\{\vx_k\}_{k=1}^K$, one per class; 
Gaussian smoothing distribution $\pno = \mathcal{N}(0, \sigma^2 I)$ and number of Monte Carlo samples $M$; 
outlier ratio $\kappa$; calibration parameter $\alpha_0$; standard Gaussian CDF $\Phi$
\State \textbf{Output:} certified result (Certified / Not Certified)
\State Construct watermarked samples $\hat{\vx}_k = \vx_k + \vdelta$ for $k = 1, \cdots, K$
\For{$k = 1$ \textbf{to} $K$}
    \State Add $M$ random noises $\{\vepsilon_i\}_{i=1}^M \sim \pno$ to $\hat{\vx}_k$, query $f$, and estimate the target-label probability $p_{\hat{y}}(\hat{\vx}_k \mid f_{\vtheta}, \pno) = \frac{1}{M}\sum_{i=1}^{M}\mathbb{I}\{\arg\max f(\hat{\vx}_k + \vepsilon_i; \vtheta) = \hat{y}\}$ via Eq.~(\ref{eq:DP_MC})
\EndFor
\State Compute the watermark robustness $\mathrm{WR} = \min_{k=1,\cdots,K} p_{\hat{y}}(\hat{\vx}_k \mid f_{\vtheta}, \pno)$ following Eq.~(\ref{eq:WR})
\State Filter the largest $m = \lfloor \kappa \cdot J \rfloor$ outliers in $\mathcal{P}_C$ and set the calibration threshold $\tau = \mathcal{P}_C^{(J - m - \lfloor \alpha_0 (J - m + 1) \rfloor)}$ (see Remark~\ref{remark:2})
\State Verify dataset ownership certification via Example~\ref{example:GS_a}: 
    \textbf{return} \texttt{Certified} if $\mathrm{WR} > \Phi\!\left(\frac{R}{\sigma} + \Phi^{-1}(\tau)\right)$, 
    else \textbf{return} \texttt{Not Certified}
\end{algorithmic}
\end{algorithm}

\section{Proofs}
\label{app:proof_theorem}

Here we provide the proofs for the results stated in the main part of the paper. We write $\beta_1(\phi)=\beta_1(\phi; H_0)$ and  $\beta_2(\phi)=\beta_2(\phi; H_1)$ for type-I and type-II error probabilities.

\subsection{Proof of Theorem \ref{thm:General_condition_a}}

\vspace{0.3em}
\noindent \textbf{Preliminaries and Auxiliary Lemmas:} Central to our theoretical results are likelihood ratio tests which are statistical hypothesis tests for testing whether a sample $\vx$ originates from a distribution $ X_0 $  or $ X_1 $. These tests are defined as
	\begin{equation}
       \label{eq:tests}
		\begin{array}{ll}
			\phi (\vx) =\begin{cases}
				1 & \text{ if } \Lambda (\vx)>l, \\
				q & \text{ if } \Lambda (\vx)=l , \quad with \quad  \Lambda (\vx)= \frac{f_{X_1(\vx)}}{f_{X_0(\vx)}}, \\
				0 & \text{ if } \Lambda (\vx)<l,
			\end{cases}  
		\end{array}
	\end{equation}
where $q$ and $l$ are chosen such that $\phi$ has significance $\alpha _1$, \ie, $\beta_1 (\phi ) = \mathbb{P}_{H_0} (\Lambda (X)>l) + q\cdot \mathbb{P}_{H_0} (\Lambda (X)= l) =\alpha _1$.
	
 \begin{lemma}[\cite{weber2023rab}]\label{lemma:A1}
		 Let $X_0$ and $X_1$ be two random variables with densities $f_0$ and $f_1$ with respect to a measure $\mu$ and denote by $\Lambda$ the likelihood ratio $\Lambda (\vx)= f_1(\vx)/f_0(\vx)$. For  $b \in [ 0,1 ] $, let $l_b := \inf \left \{ l\ge 0: \mathbb{P} (\Lambda (X_0)\le l) \ge b \right \}. $ Then it holds that
		\begin{equation}\label{eq:min_t_p}
			\begin{array}{ll}
				\mathbb{P}(\Lambda (X_0)<  l_b) \le  b \le \mathbb{P} (\Lambda (X_0)\le l_b). 
			\end{array}
		\end{equation}
		
	\end{lemma}

	\begin{lemma}[\cite{weber2023rab}]\label{lemma:A2}
Let $X_0$ and $X_1$ be random variables taking values in $\gZ$ and with probability density functions $f_0$ and $f_1$ with respect to a measure $\mu$. Let $\phi ^*$ be a likelihood ratio test for testing the null $X_0$ against the alternative $X_1$. Then for any deterministic function $\phi: \gZ \to [0,1]$ the following implications hold:

		\begin{equation}\label{eq:ratio1}
			\begin{array}{ll}
			\quad \beta_1 (\phi ) \le \beta_1 (\phi ^*) \Rightarrow \beta_2 (\phi ) \ge  \beta_2 (\phi^* ).
			\end{array}
		\end{equation}

	\end{lemma}

\begin{theorem}[\textbf{General Condition of Certified Dataset Watermarking}]
\label{thm:General_condition_a}
For a watermarked model, let $W(f_{\vtheta},\pno)$ and
$S(f_{\vtheta},\pno)$ be defined as in Eq. (\ref{eq:noise_robust}) and (\ref{eq:R_functionality}) of Definition~\ref{def:properties_cdw}, respectively.
 Dataset ownership is guaranteed to be verified if the optimal type-II error, for testing the null $ \pno+\vr \sim H _0$ against the alternative $\pno \sim H_1$, satisfies the following condition:
\begin{equation}
   \label{eq:general_condition_a}
   \beta_2^* (1-W(f_{\vtheta},\pno), H_1 )>\pcl^{(J-m-\left \lfloor \alpha_0 (J-m+1) \right \rfloor )}(g_{\vw},\pno), 
\end{equation}
where $P_C^{(j)}(g_{\vw},\pno)$ denotes the $j$-th smallest element in $P_{\blue{C}}(g_{\vw}, \pno )$, $\alpha_0$, $J$ and $m$ are defined as in Proposition \ref{pro:conformal_prediction}.	
\end{theorem}

\begin{proof}
We first show the existence of a likelihood ratio test $\phi_{W(f_{\vtheta},\pno)}$ with type-I error probability $1-W(f_{\vtheta},\pno)$. Let $Z \sim \pno + \vr$ and $Z' \sim \pno$, and recall that the likelihood ratio $\Lambda$ between the densities of $Z$ and $Z'$ is given by $\Lambda(\vx) = \frac{f_{Z'}(\vx)}{f_{Z}(\vx)}$. Furthermore, for any $b \in [0,1]$, let $l_b := \inf \left\{ l \ge 0 : \mathbb{P}(\Lambda(Z) \le l) \ge b \right\}$ and

\begin{equation}\label{eq:ratio-1}
\begin{array}{ll}
q_b=\begin{cases}
0 & \text{ if } \mathbb{P}(\Lambda(Z) = l_b) = 0, \\
\frac{\mathbb{P}(\Lambda(Z) \le l_b) - b}{\mathbb{P}(\Lambda(Z) = l_b)} & \mathrm{otherwise}.
\end{cases}  
\end{array}
\end{equation}

Note that by Lemma \ref{lemma:A1}, we have $\mathbb{P}(\Lambda(Z) \le l_b) \ge b$ and

\begin{equation}\label{eq:ratio-2}
\begin{array}{ll}
\mathbb{P}(\Lambda(Z) \le l_b) &= \mathbb{P}(\Lambda(Z) < l_b) + \mathbb{P}(\Lambda(Z) = l_b) \\
&\le b + \mathbb{P}(\Lambda(Z) = l_b),
\end{array}
\end{equation}
and hence $q_b \in [0, 1]$. For $b \in [0, 1]$, let $\phi_b$ be the likelihood ratio test defined in (\ref{eq:tests}) with $q \triangleq q_b$ and $l \triangleq l_b$. Note that $\phi_b$ has a type-I error probability $\beta_1(\phi_b) = 1 - b$. Thus, the test $\phi_{W(f_{\vtheta},\pno)}$ satisfies $\beta_1(\phi_{W(f_{\vtheta},\pno)}) = 1 - W(f_{\vtheta},\pno)$.

Next, we connect the above likelihood ratio test with 
$W(f_{\vtheta},\pno)$ and $S(f_{\vtheta},\pno)$. To avoid 
introducing a separate hypothesis test for each selected sample, 
we consider a global mixture distribution. Let 
$I\sim \mathrm{Unif}\{1,\ldots,K\}$. Under the null hypothesis 
$H_0$, we observe $Z=x_I+\vr_I+\vepsilon,\quad \vepsilon\sim\pno$, whereas under the alternative hypothesis $H_1$, we observe $Z'=x_I+\vepsilon,\quad \vepsilon\sim\pno$. For simplicity, we still denote these two distributions by 
$\pno+\vr\sim H_0$ and $\pno\sim H_1$, respectively.
Let $A=\{\vx:\arg\max f_{\vtheta}(\vx)=\hat y\}$
be the target-label decision region. From Eq. 
(\ref{eq:noise_robust}) in Definition \ref{def:properties_cdw}, 
for every $k\in\{1,\ldots,K\}$, we have $\mathbb{P}_{\vepsilon\sim\pno}
\left(\arg\max f_{\vtheta}(\vx_k+\vr_k+\vepsilon)=\hat y
\right)\ge
W(f_{\vtheta},\pno)$. Taking the average over the uniformly sampled index $I$, we obtain

\begin{equation}\label{eq:minSTR2}
\begin{aligned}
\mathbb{P}_{H_0}(Z\in A)
&=
\frac{1}{K}\sum_{k=1}^{K}
\mathbb{P}_{\vepsilon\sim\pno}
\left(
\arg\max f_{\vtheta}(\vx_k+\vr_k+\vepsilon)=\hat y
\right) \\
&\ge
W(f_{\vtheta},\pno).
\end{aligned}
\end{equation}

Define the test function $\phi(\vx)=1-\mathbb{I}_{\{\vx\in A\}}$. That is, $\phi(\vx)=1$ means that the input is not predicted 
as the defender-specified target label $\hat y$. Therefore, under 
$H_0$, the type-I error of $\phi$ satisfies

\begin{equation}\label{eq:typeI_phi}
\begin{aligned}
\beta_1(\phi;H_0)
&=\mathbb{P}_{H_0}(\phi(Z)=1)=1-\mathbb{P}_{H_0}(Z\in A) \\
&\le 1-W(f_{\vtheta},\pno) \\
&=\beta_1(\phi_{W(f_{\vtheta},\pno)};H_0).
\end{aligned}
\end{equation}

By applying Lemma \ref{lemma:A2} to the function $\phi$ and $\phi^*=\phi_{W(f_{\vtheta},\pno)}$, it follows that

\begin{equation}\label{eq:minSTR3}
\beta_2(\phi;H_1) \ge
\beta_2(\phi_{W(f_{\vtheta},\pno)};H_1).
\end{equation}

Moreover, by the Neyman-Pearson Lemma, the likelihood ratio test
$\phi_{W(f_{\vtheta},\pno)}$ attains the minimum type-II error among
all tests with type-I error at most $1-W(f_{\vtheta},\pno)$. Therefore,

\begin{equation}
  \beta_2 (\phi_{W(f_{\vtheta},\pno)} ;H_1)=\beta_2^*(1-W(f_{\vtheta},\pno),H_1).  
\end{equation}

On the other hand, under $H_1$, we have $Z'=\vx_I+\vepsilon$. 
Thus,

\begin{equation}\label{eq:S_equals_beta}
\begin{aligned}
\beta_2(\phi;H_1)
&=
\mathbb{P}_{H_1}(\phi(Z')=0)=
\mathbb{P}_{H_1}(Z'\in A) \\
&=
\frac{1}{K}\sum_{k=1}^{K}
\mathbb{P}_{\vepsilon\sim\pno}
\left(
\arg\max f_{\vtheta}(\vx_k+\vepsilon)=\hat y
\right) \\
&=
S(f_{\vtheta},\pno),
\end{aligned}
\end{equation}
where the last equality follows from Eq. (\ref{eq:R_functionality}) 
in Definition \ref{def:properties_cdw}. Combining Eqs. 
(\ref{eq:minSTR3})--(\ref{eq:S_equals_beta}), we have

\begin{equation}\label{eq:S_lower_beta}
S(f_{\vtheta},\pno)
\ge
\beta_2^*(1-W(f_{\vtheta},\pno),H_1).
\end{equation}

If Eq.~(\ref{eq:general_condition_a}) holds, then

\begin{equation}\label{eq:S_thrse}
S(f_{\vtheta},\pno)>\tau,
\end{equation}
where $\tau=
\pcl^{(J-m-\lfloor \alpha_0(J-m+1)\rfloor)}(g_{\vw},\pno)$. Moreover, Eq.~(\ref{eq:general_condition_a}) also implies
$W(f_{\vtheta},\pno)>\tau$. Indeed, a trivial randomized test that
rejects $H_0$ with probability $1-W(f_{\vtheta},\pno)$ is feasible
under the type-I error constraint $1-W(f_{\vtheta},\pno)$ and has
type-II error $W(f_{\vtheta},\pno)$. Hence, $\beta_2^*(1-W(f_{\vtheta},\pno),H_1)\le W(f_{\vtheta},\pno)$. Together with Eq.~(\ref{eq:general_condition_a}), we have
\begin{equation}\label{eq:W_thrse}
W(f_{\vtheta},\pno)>\tau.
\end{equation}

Therefore, Eqs.~(\ref{eq:S_thrse}) -- (\ref{eq:W_thrse}) imply 
\begin{equation}
   \min\{W(f_{\vtheta},\pno),S(f_{\vtheta},\pno)\}>\tau. 
\end{equation}

Based on Definition~\ref{def:CDW_formal}, dataset ownership
verification is guaranteed.
\end{proof}

\subsection{Proof of Example \ref{example:GS_a}}

\begin{example}[Robustness Conditions under Gaussian Distribution]
\label{example:GS_a}
Let the noise $\vepsilon \sim \mathcal{N}(0, \sigma^2 I)$. Given $W(f_{\vtheta},\pno)$ 
 as defined in Eq. (\ref{eq:noise_robust}) in Definition \ref{def:properties_cdw} for the watermarked model's (transformation-based) WR.  Let $R$ denote the maximum perturbation magnitude of the dataset watermark, as defined in Definition \ref{def:lpnorm}. Dataset ownership verification is guaranteed if $W(f_{\vtheta},\pno)$ satisfies the following condition:

\begin{equation}
\label{eq:gs_cp_a}
 W(f_{\vtheta},\pno)> \Phi(\frac{R}{\sigma } + \Phi^{-1}(\pcl^{(J - m - \left\lfloor \alpha_0 (J - m + 1) \right\rfloor )}(g_{\vw}, \pno)) ),
\end{equation}
where $\Phi$ is the cumulative distribution function (CDF) of the standard Gaussian distribution.
\end{example}
	\begin{proof}
		We prove this statement by direct application of Theorem  \ref{thm:General_condition_a}. For each $k \in \{1,\ldots,K\}$, let $\tilde{Z}_k\sim \mathcal{N}(\hat{\vx}_k,\sigma^2 I)$ and
$\tilde{Z}'_k\sim \mathcal{N}(\hat{\vx}_k-\vr_k,\sigma^2 I)$, where $\hat{\vx}_k=\vx_k+\vr_k$ denotes the watermarked version of $\vx_k$. By Theorem \ref{thm:General_condition_a}, there exist likelihood ratio tests $\phi^{(k)}_{W(f_{\vtheta},\pno)}$ for testing the null $\tilde{Z}_k$  against the alternative $\tilde{Z} '_k$ such that, if 
		\begin{equation}\label{eq:theorem1-0}
			\begin{array}{ll}
				\beta_2 (\phi^{(k)}_{W(f_{\vtheta},\pno)} )> P_C^{(J-m-\left \lfloor \alpha_0 (J-m+1) \right \rfloor )}(g_{\vw},\pno),\quad \forall k.
			\end{array}
		\end{equation}
		then it is guaranteed that a dataset watermark with a trigger $\vdelta$ and a target class $\hat{y}$ will be verified. We will now construct the corresponding likelihood ratio tests and show that (\ref{eq:theorem1-0}) has the form (\ref{eq:gs_cp_a}). From these definitions, the likelihood ratio between $\tilde{Z}_k$ and $\tilde{Z} '_k$ can be derived as follows:
        \begin{equation}
\begin{array}{ll}
\Lambda_k(z)=\exp\left\{\left\langle z-\hat{\vx}_k,-\vr_k\right\rangle_\Sigma
-\frac{1}{2}\left\langle \vr_k,\vr_k\right\rangle_\Sigma\right\},
\end{array}
\end{equation}
		where $\Sigma =\sigma ^2\mathrm {I} _d $ and $\left \langle \zeta,\xi  \right \rangle_\Sigma =\zeta^\top\Sigma^{-1}\xi=\frac{\zeta^\top\xi}{\sigma^2}$. Thus, since singletons have probability 0 under the Gaussian distribution, any likelihood ratio test for testing $\tilde{Z}_k$ and $\tilde{Z} '_k$ has the form
		\begin{equation}
			\begin{array}{ll}
				\phi_l^{(k)}(z)=\left\{\begin{matrix}
					1, &  \Lambda_k(z)\ge l,\\
					0, &  \Lambda_k(z)< l.
				\end{matrix}\right.
			\end{array}
		\end{equation}

		For $b \in [0,1] $	, let  
         \begin{equation}
        l_b^{(k)} \triangleq \exp\left(
\Phi^{-1}(b)\sqrt{\left\langle \vr_k,\vr_k\right\rangle_\Sigma}-\frac{1}{2}
\left\langle \vr_k,\vr_k\right\rangle_\Sigma
\right) ,   
       \end{equation} 
       and note that  $\beta_1(\phi^{(k)}_{l_b^{(k)}}) =1-b $ since 
        \begin{equation}
\begin{array}{ll}
\beta_1(\phi_{l_b^{(k)}}^{(k)})=1-\Phi\left(
\frac{\log(l_b^{(k)})+\frac{1}{2}\left\langle \vr_k,\vr_k\right\rangle_\Sigma}{
\sqrt{\left\langle \vr_k,\vr_k\right\rangle_\Sigma}}\right),
\end{array}
\end{equation}
		where $\Phi $ is the CDF of the standard normal distribution. Thus, the test $\phi^{(k)}_{W(f_{\vtheta},\pno)}$, satisfies $\beta_1(\phi^{(k)}_{W(f_{\vtheta},\pno)})=1-W(f_{\vtheta},\pno) $. Thus, computing the type-II error probability of $\phi^{(k)}_{W(f_{\vtheta},\pno)}$ yields 
\begin{equation}
\begin{array}{ll}
\beta_2(\phi^{(k)}_{W(f_{\vtheta},\pno)})
&=
\Phi\left(
\Phi^{-1}(W(f_{\vtheta},\pno))
-
\sqrt{\left\langle \vr_k,\vr_k\right\rangle_\Sigma}
\right)\\
&=
\Phi\left(
\Phi^{-1}(W(f_{\vtheta},\pno))
-
\frac{\|\vr_k\|_2}{\sigma}
\right).
\end{array}
\end{equation}

       Using $\|\vr_k\|_2\le R$ from Definition~\ref{def:lpnorm}, we have $\frac{\|\vr_k\|_2}{\sigma}\le \frac{R}{\sigma}$. By the monotonicity of $\Phi$, this yields, for every $k$, $\beta_2(\phi^{(k)}_{W(f_{\vtheta},\pno)}) \ge \Phi\left(\Phi^{-1}(W(f_{\vtheta},\pno))-\frac{R}{\sigma}\right)$. 
Since the above lower bound holds for every $k$, it also lower-bounds
the average transformation-removed stability used in Theorem~\ref{thm:General_condition_a}.
Therefore, it is sufficient to require this lower bound to be larger
than the calibration threshold.

Finally, Eq.~(\ref{eq:theorem1-0}) is satisfied if

\begin{equation}
\Phi\left(
\Phi^{-1}(W(f_{\vtheta},\pno))-\frac{R}{\sigma}\right)>P_C^{(J-m-\left\lfloor \alpha_0 (J-m+1) \right\rfloor)}(g_{\vw},\pno).
\end{equation}

Since $\Phi$ is strictly increasing, the above condition is equivalent to
 \begin{equation*}
 W(f_{\vtheta},\pno)> \Phi(\frac{R}{\sigma } + \Phi^{-1}(\pcl^{(J - m - \left\lfloor \alpha_0 (J - m + 1) \right\rfloor )}(g_{\vw}, \pno)) ).
	\end{equation*}

This is exactly Eq.~(\ref{eq:gs_cp_a}). Moreover, since $R/\sigma\ge0$,
Eq.~(\ref{eq:gs_cp_a}) also implies $W(f_{\vtheta},\pno)>\tau$, where $\tau=P_C^{(J-m-\left\lfloor \alpha_0 (J-m+1) \right\rfloor)}
(g_{\vw},\pno)$. Together with the guaranteed transformation-removed stability condition, dataset ownership verification is guaranteed by Definition~\ref{def:CDW_formal}.

\end{proof}

\subsection{Proof of Example \ref{example:US_b}}

\begin{example}[Robustness Conditions under Uniform Distribution]
\label{example:US_b}
Let the one-dimensional smoothing noise $\vepsilon \sim \mathcal{U}([e, h])$. Given $W(f_{\vtheta},\pno)$  
 as defined in Eq. (\ref{eq:noise_robust}) in Definition \ref{def:properties_cdw} for the watermarked model's (transformation-based) WR. Let $R$ denote the maximum absolute watermark perturbation magnitude, as defined in Definition \ref{def:lpnorm}. Dataset ownership verification is guaranteed if $W(f_{\vtheta},\pno)$ satisfies the following condition:
\begin{equation}
\label{eq:US_cp_b}
W(f_{\vtheta},\pno) > P_C^{(J-m-\lfloor\alpha_0(J-m+1)\rfloor)}(g_{\vw},\pno) + \frac{R}{h-e}.
\end{equation}
\end{example}

\begin{proof}
We proceed analogously to the proof of Example~\ref{example:GS_a},
but consider the one-dimensional uniform smoothing case.
For each $k \in \{1, \ldots, K\}$,
let $r_k\in\mathbb{R}$ denote the scalar watermark perturbation
and define $R=\max_{k=1,\ldots,K}|r_k|$.
Let $\hat{Z}_k \sim \mathcal{U}([e, h]) + r_k$
and $\hat{Z}'_k \sim \mathcal{U}([e, h])$,
and let $V'_k := [e+r_k,h+r_k]$ and $V_k := [e,h]$
denote the supports of $\hat{Z}_k$ and $\hat{Z}'_k$, respectively.
For any $z \in V_k \cup V'_k$, the likelihood ratio
between $\hat{Z}_k$ and $\hat{Z}'_k$ is:
\begin{equation}
\Lambda_k(z) = \frac{f_{\hat{Z}'_k}(z)}{f_{\hat{Z}_k}(z)} = \begin{cases}
0, & z \in V'_k \setminus V_k, \\
1, & z \in V'_k \cap V_k, \\
\infty, & z \in V_k \setminus V'_k.
\end{cases}
\end{equation}

Any likelihood ratio test for testing $\hat{Z}_k$ against $\hat{Z}'_k$ has the form (\ref{eq:tests}). We now construct such likelihood ratio tests $\phi^{(k)}_{W(f_{\vtheta},\pno)}$ with $\beta_1(\phi^{(k)}_{W(f_{\vtheta},\pno)})=1-W(f_{\vtheta},\pno)$ by following the construction in the proof of Theorem~\ref{thm:General_condition}. Specifically, we compute $q^{(k)}_{W(f_{\vtheta},\pno)}$ and $l^{(k)}_{W(f_{\vtheta},\pno)}$ such that the type-I error probability is satisfied. Notice that
\begin{equation}
\begin{array}{ll}
b_{0,k} &\triangleq \mathbb{P}_{\hat{Z}_k}(V'_k \setminus V_k)
= 1-\mathbb{P}_{\hat{Z}_k}(V'_k\cap V_k)\\
&= 1-\left(1-\frac{|\vr_k|}{h-e}\right)_+
\le \frac{|\vr_k|}{h-e}.
\end{array}
\end{equation}

For $l \geq 0$, we have:
\begin{equation}
\begin{array}{ll}
\mathbb{P}(\Lambda_k(\hat{Z}_k) \le l)
&= \left\{\begin{matrix}
\mathbb{P}_{\hat{Z}_k}(V'_k \setminus V_k), & l < 1,\\
\mathbb{P}_{\hat{Z}_k}(V'_k), & \mathrm{otherwise},
\end{matrix}\right.\\
&= \left\{\begin{matrix}
b_{0,k}, & l < 1,\\
1, & \mathrm{otherwise}.
\end{matrix}\right.
\end{array}
\end{equation}

Recall that $l_b^{(k)} := \inf\{l \geq 0 : \mathbb{P}(\Lambda_k(\hat{Z}_k) \leq l) \geq b\}$ for $b \in [0, 1]$, and hence:
\begin{equation}
l_b^{(k)} = \begin{cases}
0, & b \leq b_{0,k},\\
1, & \text{otherwise}.
\end{cases}
\end{equation}

We notice that, if $b \le b_{0,k}$, then $l^{(k)}_{W(f_{\vtheta},\pno)}=0$. This implies that the type-II error probability of the corresponding test $\phi^{(k)}_{W(f_{\vtheta},\pno)}$ is 0, since in this case $\Lambda_k(\hat{Z}'_k)>0$ almost surely under $\hat{Z}'_k$:
\begin{equation}
\begin{array}{ll}
\beta_2(\phi^{(k)}_{W(f_{\vtheta},\pno)})
&= 1-\mathbb{P}(\Lambda_k(\hat{Z}'_k)>0)\\
&-q^{(k)}_{W(f_{\vtheta},\pno)}\mathbb{P}(\Lambda_k(\hat{Z}'_k)=0)\\
&=0.
\end{array}
\end{equation}

Thus, in this case, $\beta_2(\phi^{(k)}_{W(f_{\vtheta},\pno)}) >
P_C^{(J-m-\lfloor\alpha_0(J-m+1)\rfloor)}(g_{\vw},\pno)$ can never be satisfied, and we find that $b>b_{0,k}$ (\ie, $W(f_{\vtheta},\pno)>b_{0,k}$) is necessary. In this case, we have $l^{(k)}_{W(f_{\vtheta},\pno)}=1$. Let $q^{(k)}_{W(f_{\vtheta},\pno)}$ be defined as in the proof of Theorem~\ref{thm:General_condition}, \ie,
\begin{equation}
q^{(k)}_{W(f_{\vtheta},\pno)}
=\frac{\mathbb{P}(\Lambda_k(\hat{Z}_k)\le 1)-W(f_{\vtheta},\pno)}
{\mathbb{P}(\Lambda_k(\hat{Z}_k)=1)}
=\frac{1-W(f_{\vtheta},\pno)}{1-b_{0,k}}.
\end{equation}

Clearly, the corresponding likelihood ratio test $\phi^{(k)}_{W(f_{\vtheta},\pno)}$ has type-I error probability $1-W(f_{\vtheta},\pno)$. Furthermore, notice that
\begin{equation}
\begin{array}{ll}
\mathbb{P}_{\hat{Z}_k}(\hat{Z}_k\in V'_k\setminus V_k)=\mathbb{P}_{\hat{Z}'_k}(\hat{Z}'_k\in V_k\setminus V'_k)
=b_{0,k},\\
\mathbb{P}_{\hat{Z}_k}(\hat{Z}_k\in V'_k\cap V_k)
=\mathbb{P}_{\hat{Z}'_k}(\hat{Z}'_k\in V'_k\cap V_k)
=1-b_{0,k}.
\end{array}
\end{equation}
Hence its type-II error $\beta_2(\phi^{(k)}_{W(f_{\vtheta},\pno)})$ is given by
\begin{equation}
\begin{array}{ll}
\beta_2(\phi^{(k)}_{W(f_{\vtheta},\pno)})
&= 1-\mathbb{P}(\Lambda_k(\hat{Z}'_k)>1)\\
&-q^{(k)}_{W(f_{\vtheta},\pno)}\cdot \mathbb{P}(\Lambda_k(\hat{Z}'_k)=1)\\
&= 1-b_{0,k}-\dfrac{1-W(f_{\vtheta},\pno)}{1-b_{0,k}}\cdot(1-b_{0,k})\\
&= W(f_{\vtheta},\pno)-b_{0,k}.
\end{array}
\end{equation}

Using $|r_k|\le R$, we have
$b_{0,k}\le \frac{|\vr_k|}{h-e}\le \frac{R}{h-e}$, which yields
\begin{equation}
\beta_2(\phi^{(k)}_{W(f_{\vtheta},\pno)})
=
W(f_{\vtheta},\pno)-b_{0,k}
\ge
W(f_{\vtheta},\pno)-\frac{R}{h-e}.
\end{equation}
Since the above lower bound holds for every $k$, it also guarantees
the average transformation-removed stability condition used in
Definition~\ref{def:properties_cdw}.

Finally, the statement follows, since $\beta_2(\phi^{(k)}_{W(f_{\vtheta},\pno)}) >
P_C^{(J-m-\lfloor\alpha_0(J-m+1)\rfloor)}(g_{\vw},\pno)$ is satisfied if
\begin{equation}
W(f_{\vtheta},\pno)
>
P_C^{(J-m-\lfloor\alpha_0(J-m+1)\rfloor)}(g_{\vw},\pno)
+
\frac{R}{h-e}.
\end{equation}
This is exactly Eq.~(\ref{eq:US_cp_b}). Moreover, since $R/(h-e)\ge0$,
Eq.~(\ref{eq:US_cp_b}) also implies $W(f_{\vtheta},\pno)>\tau$, where
$\tau=P_C^{(J-m-\lfloor\alpha_0(J-m+1)\rfloor)}(g_{\vw},\pno)$. Together
with the guaranteed transformation-removed stability condition, dataset
ownership verification is guaranteed by Definition~\ref{def:CDW_formal}.
\end{proof}

  \begin{figure*}[!t]
\vspace{-1em}
\centering
\subfigure[GTSRB]{\includegraphics[width=3.3in]{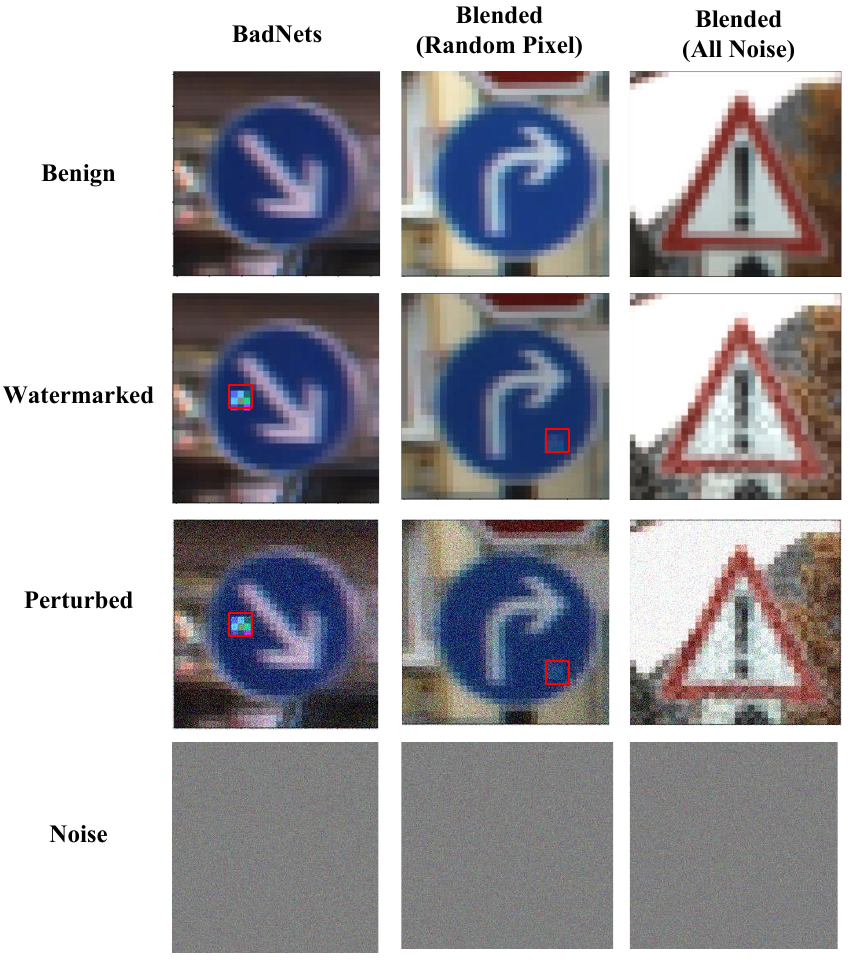}%
\label{watermark1}}
\hfil
\subfigure[CIFAR-10]{\includegraphics[width=3.3in]{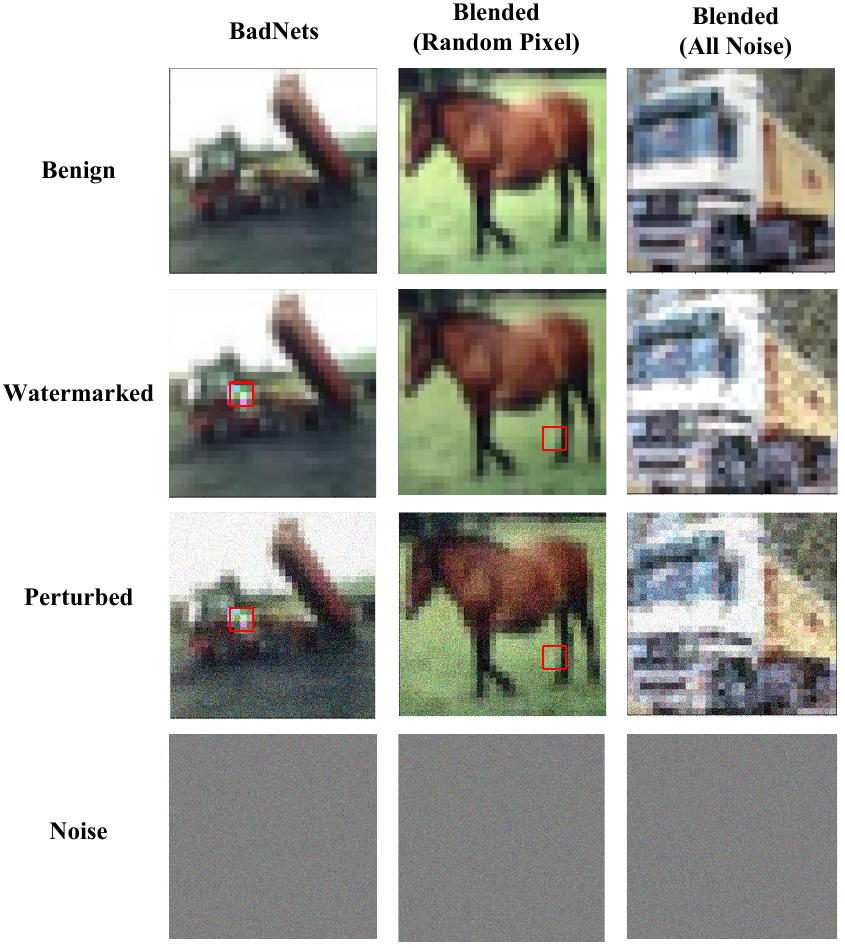}%
\label{watermark2}}
\vspace{-0.4em}
\caption{The examples of benign, watermarked, perturbed images, and the corresponding noise generated by BadNets and the Blended watermark with different trigger patterns on the GTSRB and CIFAR-10 dataset. The randomly positioned trigger areas are indicated by red boxes.}
\label{fig:examples_imgs}

\end{figure*}

\subsection{Proof of Theorem \ref{thm:certdw_fpr_0}}

\begin{theorem}[\textbf{High-Probability FPR Control of CertDW}]
\label{thm:certdw_fpr_0}
Let $C_j=P(g_{\vw}^{(j)},\pno)$, $j=1,\ldots,J$, be the empirical PP
calibration statistics computed using $M$ Monte Carlo samples for $J$
benign models, and assume that
$C_1,\ldots,C_J$ are \textit{i.i.d.} samples from an arbitrary (possibly discrete) distribution
$F_C$. Let $\mathsf{T}=\mathsf{T}(f_{\mathrm{s}},\pno)$ denote a scalar statistic used for
false-positive analysis of a suspicious model $f_{\mathrm{s}}$ that is not
trained on the protected dataset, and let $F_{\mathsf{T}}$ be the distribution of $\mathsf{T}$.
Assume that $\mathsf{T}$ is independent of the calibration set and
\begin{equation}
\label{eq:certdw_fpr_domination}
F_{\mathsf{T}}(t)\ge F_C(t),\quad \forall t\in\mathbb{R}.
\end{equation}
Let $C_{(1)}\le\cdots\le C_{(J)}$ be the order statistics of the
calibration set. Following Proposition~\ref{pro:conformal_prediction}, define
the calibration threshold as
$\tau_J=C_{\left(J-m-\lfloor\alpha_0(J-m+1)\rfloor\right)}$, where $J$, $m$,
and $\alpha_0$ are defined in Proposition~\ref{pro:conformal_prediction}.
We assume that the order-statistic index
$J-m-\lfloor\alpha_0(J-m+1)\rfloor$ lies in $\{1,\ldots,J\}$. The FPR with
respect to the statistic $\mathsf{T}$ is given by
$Z_J=\mathbb{P}\left(\mathsf{T}>\tau_J\mid C_1,\ldots,C_J\right)$. Then, for any
$\zeta\in(0,1)$, with probability at least $1-\zeta$ over the random draw
of the calibration set,
\begin{equation}
\label{eq:certdw_fpr_bound}
Z_J\le \frac{m+\lfloor\alpha_0(J-m+1)\rfloor}{J} +\sqrt{\frac{\log(2/\zeta)}{2J}}.
\end{equation}
\end{theorem}

\begin{figure*}[!t]
    \begin{minipage}[t]{0.48\linewidth}
        \centering
	\subfigure[]{
		\includegraphics[width=0.95\linewidth]{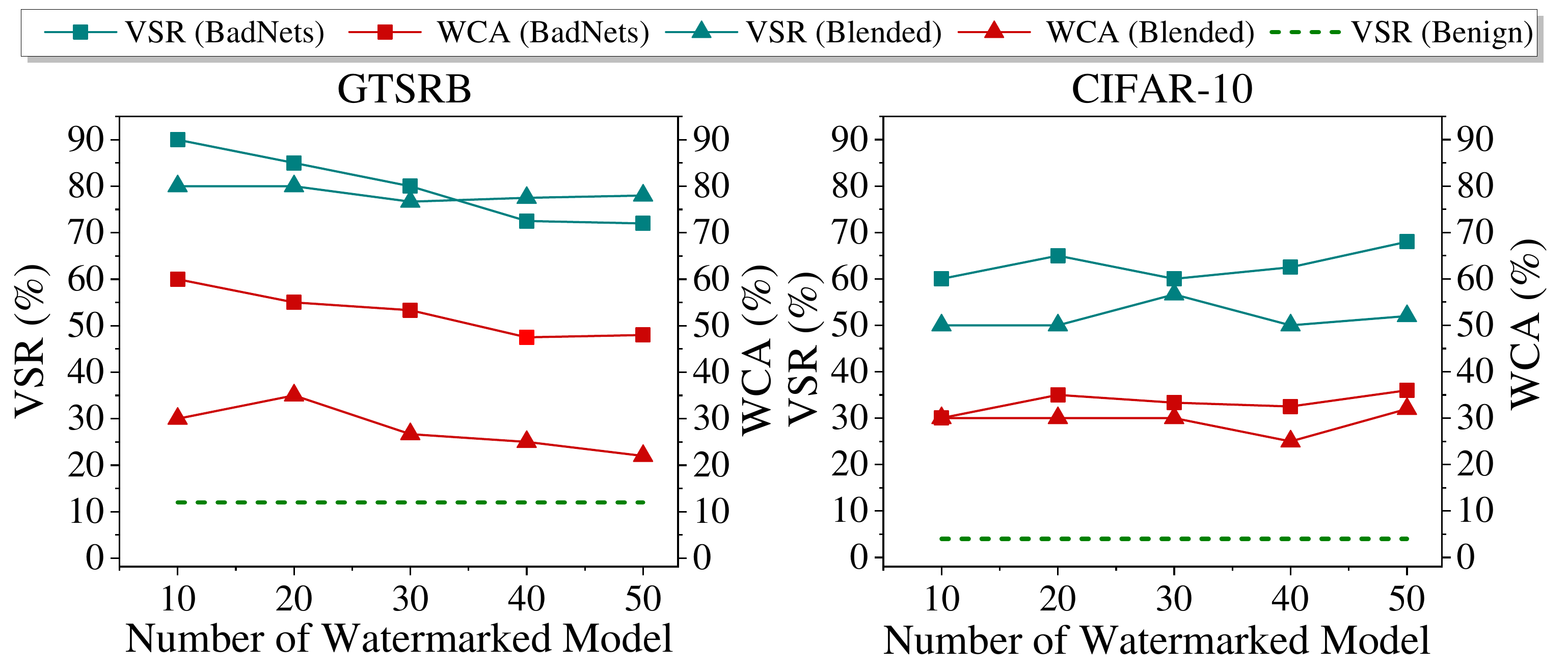}}\hspace{0.5em}
        \vspace{-0.8em}
	\caption{Effects of the number of watermarked models.}
     \label{fig:water_model}
    \end{minipage}\hspace{0.5em}
    \begin{minipage}[t]{0.48\linewidth}
        \centering
	\subfigure[]{
		\includegraphics[width=0.95\linewidth]{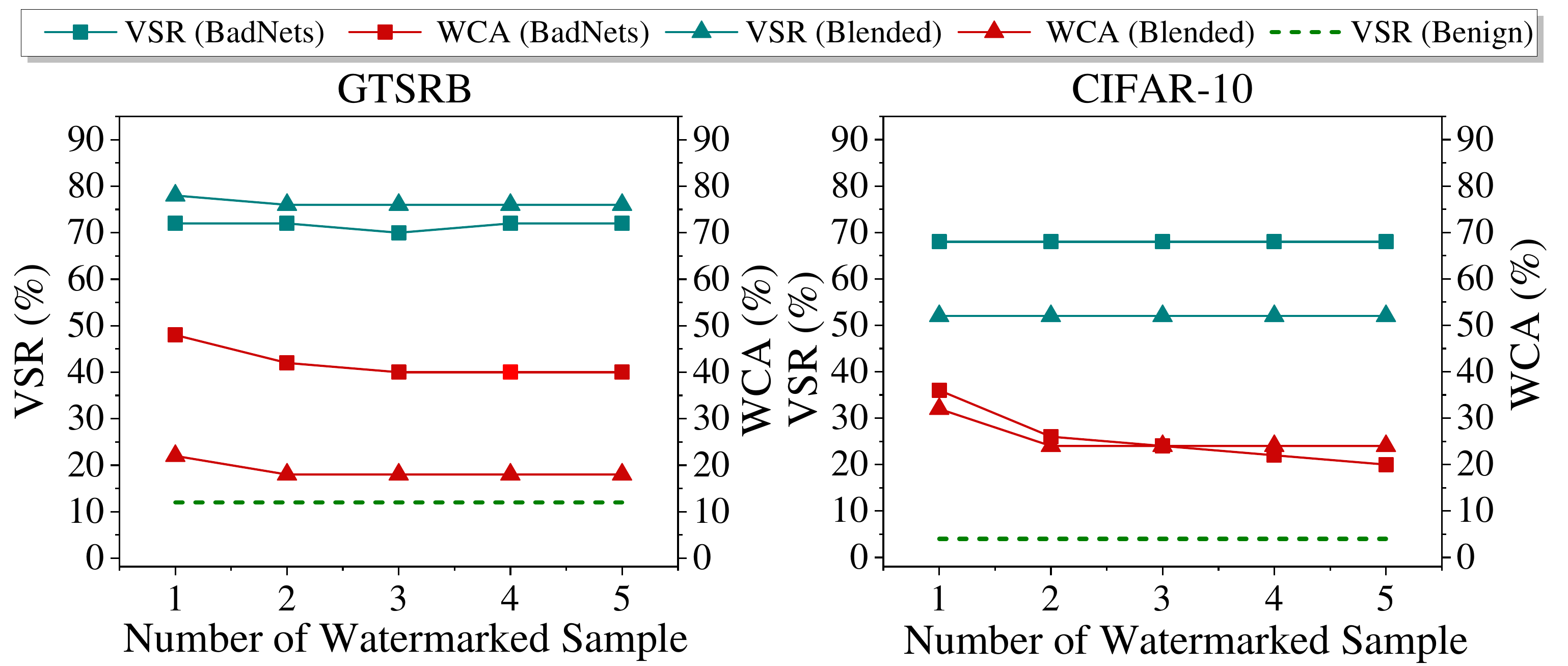}}\hspace{0.5em}
        \vspace{-0.8em}
	\caption{Effects of the number of watermarked samples.}
 \label{fig:water_sample}
    \end{minipage}%
\end{figure*}

\begin{figure}
    \centering
    \includegraphics[width=0.95\linewidth]{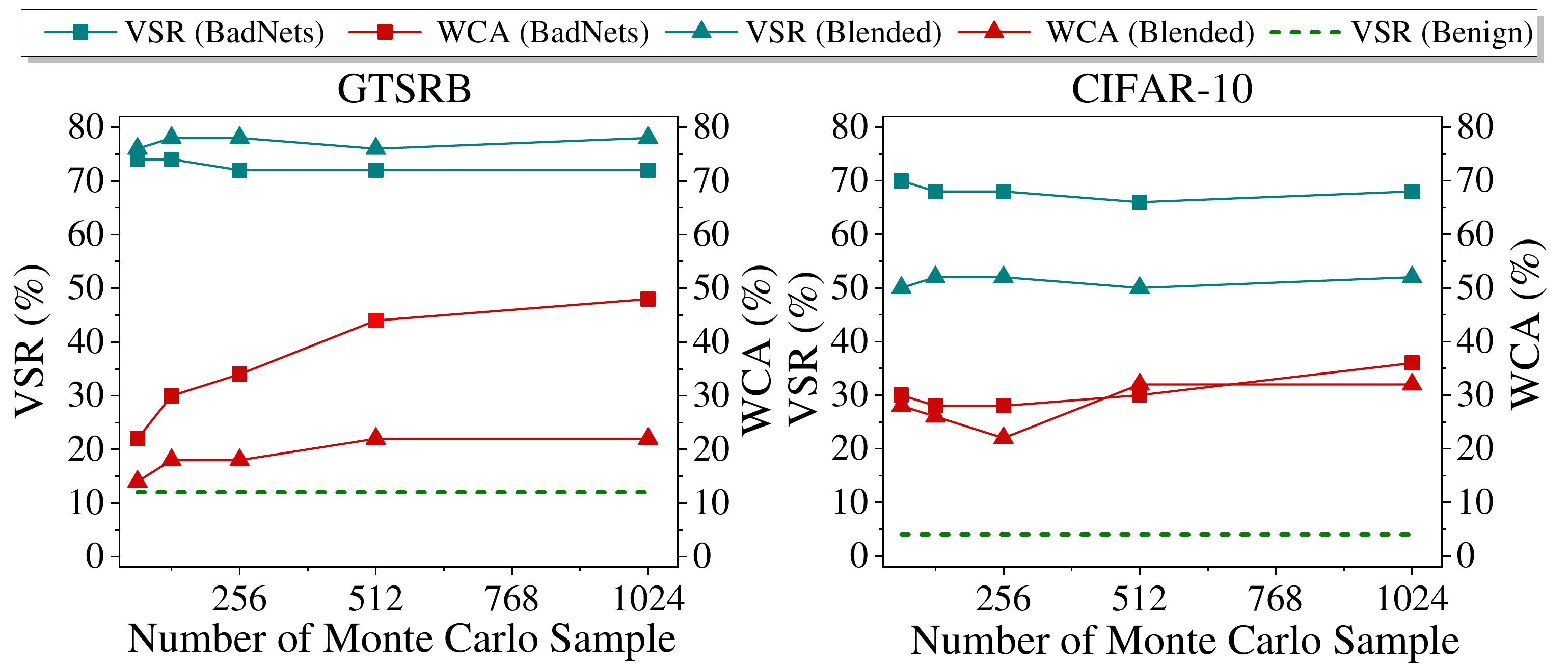}
    \caption{Effects of the number of Monte Carlo samples.}
    \label{fig:MC}
\end{figure}

\begin{proof}
Let $\iota_J=J-m-\lfloor\alpha_0(J-m+1)\rfloor$. Then the
calibration threshold can be written as $\tau_J=C_{(\iota_J)}$. By definition,
the conditional false positive rate is
$Z_J=\mathbb{P}\left(\mathsf{T}>\tau_J\mid C_1,\ldots,C_J\right)$. Since $\mathsf{T}$
follows the distribution $F_{\mathsf{T}}$ and is independent of the calibration set,
we have
\begin{equation}
\label{eq:fpr_as_tail}
Z_J=1-F_{\mathsf{T}}(\tau_J).
\end{equation}

By the stochastic dominance assumption in
Eq.~(\ref{eq:certdw_fpr_domination}), we have
$F_{\mathsf{T}}(t)\ge F_C(t)$ for all $t$. Taking $t=\tau_J$ gives
$F_{\mathsf{T}}(\tau_J)\ge F_C(\tau_J)$. Therefore, from
Eq.~(\ref{eq:fpr_as_tail}), we obtain
\begin{equation}
\label{eq:fpr_reduce_to_calibration}
Z_J=1-F_{\mathsf{T}}(\tau_J)\le 1-F_C(\tau_J).
\end{equation}

Next, we relate the population CDF $F_C(\tau_J)$ to the empirical CDF of
the calibration statistics via the Dvoretzky-Kiefer-Wolfowitz inequality.
Define the empirical CDF as
$\widehat F_J(t)=\frac{1}{J}\sum_{j=1}^{J}\mathbb{I}\{C_j\le t\}$.
By the Dvoretzky-Kiefer-Wolfowitz inequality with Massart's tight
constant~\cite{dvoretzky1956asymptotic,massart1990tight}, for any
$\epsilon>0$,
\begin{equation}
\mathbb{P}\left(\sup_{t\in\mathbb{R}}\left|\widehat F_J(t)-F_C(t)\right|>\epsilon\right)\le 2e^{-2J\epsilon^2}.
\end{equation}
Set $\epsilon_J(\zeta)=\sqrt{\frac{\log(2/\zeta)}{2J}}$. Then, with
probability at least $1-\zeta$ over the calibration set,
\begin{equation}
\label{eq:dkw_event}
\sup_{t\in\mathbb{R}}\left|\widehat F_J(t)-F_C(t)\right|\le \epsilon_J(\zeta).
\end{equation}

On the event in Eq.~(\ref{eq:dkw_event}), for any $t$ we have
$F_C(t)\ge \widehat F_J(t)-\epsilon_J(\zeta)$. Taking $t=\tau_J$ gives
\begin{equation}
\label{eq:fc_tau_lower}
F_C(\tau_J)\ge\widehat F_J(\tau_J)-\epsilon_J(\zeta).
\end{equation}

Since $\tau_J=C_{(\iota_J)}$ is the $\iota_J$-th order statistic of
the calibration set, we always have
$\widehat F_J(\tau_J)=\frac{1}{J}\sum_{j=1}^{J}\mathbb{I}\{C_j\le C_{(\iota_J)}\}\geq \frac{\iota_J}{J}$.
This inequality also holds when $F_C$ is discrete and  equal values occur;
therefore, no continuity assumption on $F_C$ is required.
Substituting it into Eq.~(\ref{eq:fc_tau_lower}), we obtain
\begin{equation}
F_C(\tau_J)\geq\frac{\iota_J}{J}-\epsilon_J(\zeta).
\label{eq:fc_tau_final}
\end{equation}

Combining Eq.~(\ref{eq:fpr_reduce_to_calibration}) and
Eq.~(\ref{eq:fc_tau_final}), we have
$Z_J\le 1-F_C(\tau_J)\le 1-\frac{\iota_J}{J}+\epsilon_J(\zeta)$.
Using $\iota_J=J-m-\lfloor\alpha_0(J-m+1)\rfloor$, we obtain
$1-\frac{\iota_J}{J}=\frac{m+\left\lfloor\alpha_0(J-m+1)\right\rfloor}{J}$.
Therefore,
\begin{equation}
Z_J\le\frac{m+\left\lfloor\alpha_0(J-m+1)\right\rfloor}{J}+\sqrt{\frac{\log(2/\zeta)}{2J}}.
\end{equation}
This completes the proof.
\end{proof}

\section{Examples of Benign, Watermarked, and Perturbed Samples}
\label{app:sample_examples}

We hereby provide illustrative examples of benign samples, their corresponding watermarked and perturbed samples, and the added noise generated by different watermarking methods on the GTSRB and CIFAR-10 datasets in Figure~\ref{fig:examples_imgs}.

\section{Additional Ablation Studies}
\label{app:Ablation}

In this section, we provide additional ablation studies to further investigate the behavior and stability of CertDW under different experimental settings. Specifically, we analyze the impacts of four key factors: the number of watermarked models, the number of watermarked samples per class, the Monte Carlo sample size, and the calibration model choice.

\subsection{Impact of the Number of Watermarked Models}

As shown in Figure~\ref{fig:water_model}, both VSR and WCA remain highly stable as the number of watermarked models varies, with only moderate fluctuations across both datasets and watermark types. Meanwhile, the VSR on independent models (\ie, FPR) consistently stays at a low level, further validating the reliability of CertDW. These results demonstrate that CertDW is robust to the choice of the number of watermarked models. Even with as few as 10 watermarked models, CertDW can provide stable and reliable ownership verification. This property offers practical flexibility, allowing dataset owners to adjust the number of watermarked models based on their available computational resources.

\begin{table*}[!t]
\captionsetup{font=small}
\caption{The performance (\%) of CertDW with different calibration 
models on the GTSRB dataset: \textbf{Benign Models} (100 models trained on 5,000 reserved samples) vs. \textbf{Independent Models} (100 models trained on the full training dataset). We evaluate all methods under three different noise levels: 1.5, 2.5, and 3.5.}
\vspace{-0.5em}
\centering
\scalebox{0.95}{
\begin{tabular}{c|c|cc|cc|cc}
\toprule
\multirow{2}{*}{Watermark$\downarrow$} & \multicolumn{1}{c|}{$\sigma$$\rightarrow$} & \multicolumn{2}{c|}{1.5} & \multicolumn{2}{c|}{2.5} & \multicolumn{2}{c}{3.5} \\ \cline{2-8}
 & Calibration Models$\downarrow$, Metric$\rightarrow$ & VSR & WCA & VSR & WCA & VSR & WCA \\
\hline
\multirow{2}{*}{BadNets}
 & Benign Models      & 88 & 28 & 72 & 48 & 72 & 54 \\
 & Independent Models & 96 & 62 & 90 & 58 & 80 & 54 \\
\hline
\multirow{2}{*}{Blended (patch)}
 & Benign Models      & 82 & 6 & 78 & 22 & 76 & 36 \\
 & Independent Models & 94 & 12 & 86 & 34 & 84 & 46 \\
\bottomrule
\end{tabular}
}
\label{tab:calibration_compare_gtsrb}
\end{table*}

\begin{table*}[!t]
\captionsetup{font=small}
\caption{The performance (\%) of CertDW with different calibration 
models on the CIFAR-10 dataset: \textbf{Benign Models} (100 models trained on 5,000 reserved samples) vs. \textbf{Independent Models} (100 models trained on the full training dataset). We evaluate all methods under three different noise levels: 0.6, 1.2, and 1.8.}
\vspace{-0.5em}
\centering
\scalebox{0.95}{
\begin{tabular}{c|c|cc|cc|cc}
\toprule
\multirow{2}{*}{Watermark$\downarrow$} & \multicolumn{1}{c|}{$\sigma$$\rightarrow$} & \multicolumn{2}{c|}{0.6} & \multicolumn{2}{c|}{1.2} & \multicolumn{2}{c}{1.8} \\ \cline{2-8}
 & Calibration Models$\downarrow$, Metric$\rightarrow$ & VSR & WCA & VSR & WCA & VSR & WCA \\
\hline
\multirow{2}{*}{BadNets}
 & Benign Models      & 70 & 24 & 68 & 36 & 64 & 40 \\
 & Independent Models & 78 & 54 & 90 & 62 & 90 & 60 \\
\hline
\multirow{2}{*}{Blended (patch)}
 & Benign Models      & 48 & 20 & 52 & 32 & 54 & 32 \\
 & Independent Models & 66 & 34 & 74 & 50 & 72 & 50 \\
\bottomrule
\end{tabular}
}
\label{tab:calibration_compare_cifar}
\end{table*}

\subsection{Impact of the Number of Watermarked Samples per Class}

As shown in Figure~\ref{fig:water_sample}, both VSR and WCA remain highly stable as the number of watermarked samples per class varies from 1 to 5, with only small fluctuations across both datasets and watermark types. Meanwhile, the VSR on independent models (\ie, FPR) consistently stays at a low level. In practice, using one sample per class already enables CertDW to perform reliable ownership verification at low computational cost, making it a practical and efficient choice for real-world dataset copyright protection.

\subsection{Impact of the Monte Carlo Sample Size}

As shown in Figure~\ref{fig:MC}, both VSR and WCA gradually improve and stabilize as the Monte Carlo sample size $M$ increases. Specifically, when $M$ is small (\eg, 256), the Monte Carlo estimation of WR and PP exhibits noticeable variance, leading to lower and less stable verification performance. As $M$ increases to 768 or beyond, the estimation becomes more accurate, and both VSR and WCA converge to stable values. Meanwhile, the VSR on independent models (\ie, FPR) consistently stays at a low level, further validating the reliability of CertDW. When $M$ reaches our default setting of 1024, the verification performance is sufficiently stable, and further increasing $M$ yields only marginal gains at the cost of additional computation.

\subsection{Impact of Calibration Model Choice}

We hereby compare two calibration sets: 100 benign models trained on 5,000 reserved samples and 100 independent models trained on the full benign training dataset. As shown in Tables~\ref{tab:calibration_compare_gtsrb}-\ref{tab:calibration_compare_cifar}, CertDW remains effective under both calibration model sources across two datasets, two watermark types, and three noise levels. Using independent models generally achieves higher VSR and WCA, likely because these models are trained on the full benign dataset and thus produce more concentrated PP distributions, leading to more stable calibration thresholds. We use benign models as the default setting to maintain role separation between calibration and FPR estimation. Overall, these results show that CertDW is robust to the choice of calibration model.

\begin{table}[!t]
\caption{The average prediction entropy under STRIP detection for models watermarked by Vanilla BadNets and our CertDW on GTSRB and CIFAR-10 datasets. A higher entropy indicates stronger resistance to STRIP.}
\centering
\begin{tabular}{c|cc}
\toprule
Dataset, Method$\rightarrow$ & Vanilla BadNets & CertDW \\
\hline
GTSRB     & 0.16 & 0.46  ($3\times$)\\
CIFAR-10  & 0.10 & 1.15 ($12\times$)\\
\bottomrule
\end{tabular}
\label{tab:strip}
\end{table}

\begin{table}[!t]
\caption{The performance (\%) of CertDW on watermarked models repaired by NAD on GTSRB and CIFAR-10 datasets. Higher VSR and WCA indicate stronger resistance to NAD.}
\centering
\begin{tabular}{c|cc}
\toprule
Dataset$\downarrow$, Metric$\rightarrow$ & VSR & WCA \\
\hline
    GTSRB  & 68 & 22 \\
  CIFAR-10  & 56  & 32  \\
\bottomrule
\end{tabular}
\label{tab:nad}
\end{table}

\section{Additional Analysis on Adaptive Countermeasures}
\label{app:adaptive_countermeasures}

We hereby further analyze the resistance of CertDW against two representative adaptive countermeasures, including STRIP-based detection and NAD-based model repair.

\vspace{0.3em}
\noindent \textbf{The Resistance to STRIP.}  Following the previous work~\cite{gao2021design}, STRIP distinguishes poisoned samples (low entropy) from benign samples (high entropy) by superimposing random clean images onto the suspicious image and computing the prediction entropy. On GTSRB and CIFAR-10 datasets, we randomly select 2,000 test samples, generate watermarked versions using Vanilla BadNets and CertDW respectively, and compute the average entropy using the open-source implementation of STRIP with default settings\footnote{https://github.com/yjkim721/STRIP-ViTA} \cite{gao2021design}. As shown in Table~\ref{tab:strip}, CertDW's entropy is about $3$-$12$ times larger than Vanilla BadNets's (CIFAR-10: 1.15 vs 0.10; GTSRB: 0.46 vs 0.16). This is because CertDW's triggers are constrained by an $\ell_2$-norm budget, leading to smaller perturbations that are easily disrupted when STRIP superimposes random images. These results indicate that CertDW is more resistant to STRIP than Vanilla BadNets.

\vspace{0.3em}
\noindent \textbf{The Resistance to NAD.}  Following the previous work~\cite{li2021neural}, NAD is a repairing-based backdoor defense based on neural attention distillation, which uses a small amount of clean data to align the intermediate-layer attention maps of a watermarked model with those of a benign teacher model 
to remove the watermark. Specifically, we implement NAD using the open-source code from BackdoorBox\footnote{https://github.com/THUYimingLi/BackdoorBox} \cite{li2023backdoorbox}, repair each watermarked model with 5\% of clean training data, and re-evaluate CertDW on the repaired models. As shown in Table~\ref{tab:nad}, CertDW remains resistant to NAD to some extent, achieving 56\% VSR and 32\% WCA on CIFAR-10, and 68\% VSR and 22\% WCA on GTSRB. These results demonstrate that CertDW is resilient to NAD-based repair to some extent.

\begin{figure}[!t]
    \centering
    \subfigure[GTSRB]{
		\includegraphics[width=0.48\textwidth]{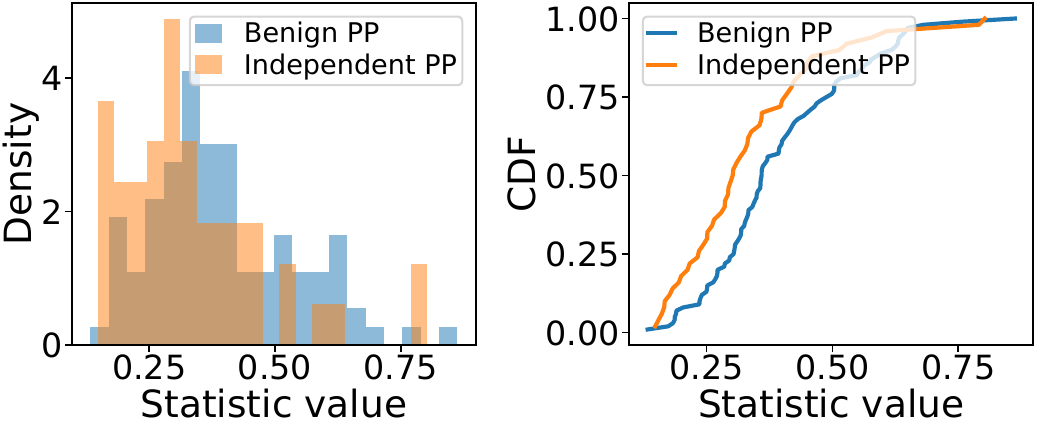}
        \label{stochastic_dominance_gtsrb}}
    \hspace{-1.0em}
    \subfigure[CIFAR-10]{
		\includegraphics[width=0.48\textwidth]{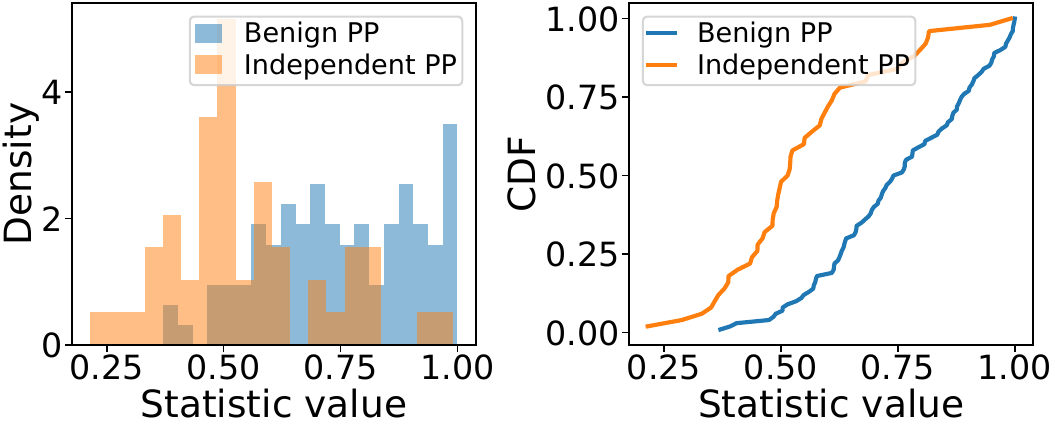}
        \label{stochastic_dominance_cifar10}}
	\vspace{-0.8em}
    \caption{Empirical validation of the stochastic dominance assumption in
Theorem~\ref{thm:certdw_fpr}. We compare the PP distributions of benign
calibration models and independent models using histograms and empirical CDFs.}
    \label{fig:stochastic_dominance}
    	\vspace{-0.8em}
    \end{figure}

\begin{figure*}[!t]
    \centering
    \includegraphics[width=0.95\textwidth]{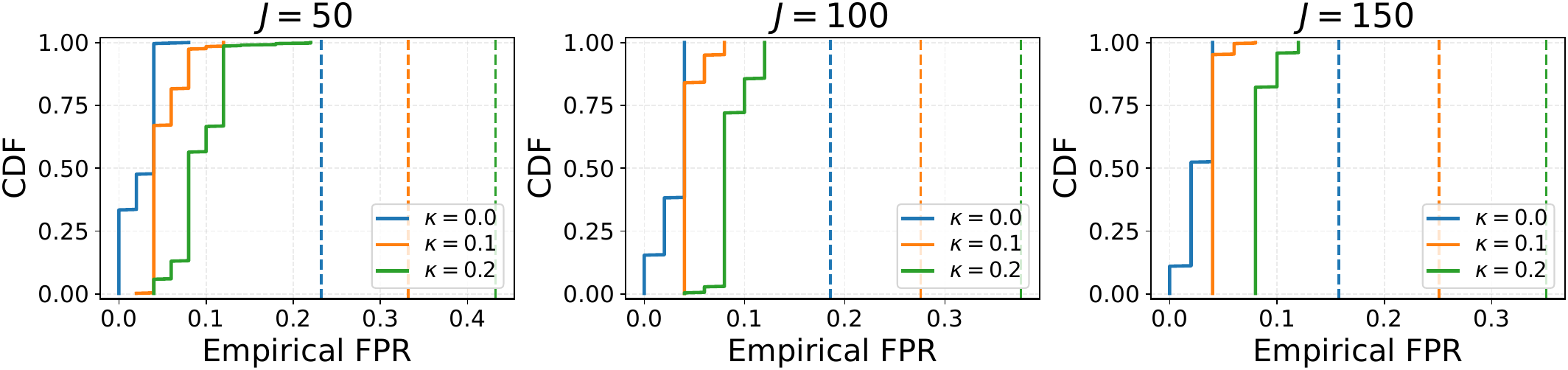}
    \vspace{-0.6em}
    \caption{ Empirical validation of the high-probability FPR bound in Theorem~\ref{thm:certdw_fpr} on GTSRB. Solid curves
denote empirical FPR CDFs, and dashed vertical lines denote the theoretical
upper bounds. }
    \label{fig:fpr_bound_validation}
    \vspace{-0.8em}
\end{figure*}

\begin{table}[t]
\centering
\caption{
Empirical validation of the high-probability FPR bound on GTSRB. We report the mean FPR, 95\% FPR, and theoretical bound for different $J$ and $\kappa$.}
\label{tab:fpr_bound_validation}
\setlength{\tabcolsep}{4pt}
\begin{tabular}{c c c c c}
\toprule
$J$ & $\kappa$ & Mean FPR & 95\% FPR & Bound \\
\midrule
50  & 0.0 & 0.0239 & 0.0400 & 0.2321 \\
50  & 0.1 & 0.0511 & 0.0800 & 0.3321 \\
50  & 0.2 & 0.0925 & 0.1200 & 0.4321 \\
\midrule
100 & 0.0 & 0.0293 & 0.0400 & 0.1858 \\
100 & 0.1 & 0.0442 & 0.0610 & 0.2758 \\
100 & 0.2 & 0.0878 & 0.1200 & 0.3758 \\
\midrule
150 & 0.0 & 0.0273 & 0.0400 & 0.1576 \\
150 & 0.1 & 0.0400 & 0.0400 & 0.2509 \\
150 & 0.2 & 0.0844 & 0.1000 & 0.3509 \\
\bottomrule
\end{tabular}
\end{table}

\section{Empirical Validation of FPR Control}
\label{app:fpr_validation}

We hereby provide additional empirical evidence for the FPR control analysis in Theorem~\ref{thm:certdw_fpr}. Specifically, we conduct two complementary experiments. First, we empirically examine the stochastic dominance relation between the PP statistics of benign calibration models and those of independent models. Second, we validate whether the empirical FPR is controlled by the derived high-probability upper bound under different calibration sizes and trimming ratios.

\vspace{0.3em}
\noindent \textbf{Empirical Validation of the Stochastic Dominance Assumption.} To support the
stochastic dominance assumption in Theorem~\ref{thm:certdw_fpr}, we compare
the PP statistics of benign calibration models and independent models not
trained on the protected dataset. As shown in
Figure~\ref{fig:stochastic_dominance}, the empirical CDF of independent PP
generally lies above that of benign calibration PP on CIFAR-10 and GTSRB.
Specifically, the average PP values of benign calibration models and
independent models are $0.7515$ and $0.5481$ on CIFAR-10, and $0.3960$ and
$0.3349$ on GTSRB, respectively. These results indicate that the PP
calibration statistics of benign models are empirically more conservative
than the PP statistics of independent models, supporting the stochastic
dominance assumption used in Theorem~\ref{thm:certdw_fpr}. In addition, the
WR statistics of independent models are much smaller than their PP
statistics, with average values of $0.0504$ on CIFAR-10 and $0.0004$ on
GTSRB. This further confirms that models not trained on the protected
dataset do not learn the watermark-target association. Therefore, using
independent PP for empirical FPR evaluation provides a stricter assessment
than directly using independent WR.

\begin{table*}[!t]
\vspace{0.5em}
\captionsetup{font=small}
\caption{The performance (\%) of CertDW on the SubImageNet dataset with three modern backbones (ResNet-34, ResNet-50, and ResNet-101). The watermarking technique is BadNets with a $5\times 5$ trigger.}
\vspace{-0.5em}
\centering
\scalebox{1.0}{
\begin{tabular}{c|cc|ccc}
\toprule
\multirow{2}{*}{Architecture$\downarrow$} & \multicolumn{2}{c|}{Dataset Watermarking} & \multicolumn{3}{c}{Dataset Ownership Verification} \\
\cline{2-6}
 & BA & WSR & VSR & WCA & WSR \\
\hline
ResNet-34  & 82.40 & 97.16 & 54 & 16 & 62.21 \\
ResNet-50  & 81.40 & 97.16 & 26 & 18 & 54.44 \\
ResNet-101 & 80.90 & 98.00 & 26 & 16 & 51.37 \\
\bottomrule
\end{tabular}
}
\label{table:subimagenet}
\end{table*}

\vspace{0.3em}
\noindent \textbf{Empirical Validation of the FPR Bound.} We further validate the high-probability FPR bound in Theorem~\ref{thm:certdw_fpr} on GTSRB. We use
200 benign calibration models to form the calibration pool and 50 independent
models to evaluate the empirical FPR. For each calibration size $J\in\{50,100,150\}$ and trimming ratio $\kappa\in\{0,0.1,0.2\}$, we repeatedly sample PP calibration statistics from benign models, compute the corresponding calibration threshold, and evaluate the fraction of independent models whose PP statistics exceed the threshold. We set the confidence parameter as \(\zeta=0.05\) when computing the theoretical bound. As shown in Figure~\ref{fig:fpr_bound_validation}, the empirical FPR
distributions are consistently located to the left of the corresponding
theoretical upper bounds, indicating that the derived bound provides
conservative FPR control. Table~\ref{tab:fpr_bound_validation} further
reports the mean FPR, the 95\% quantile of empirical FPR, and the theoretical
bound. Across all settings, the 95\% empirical FPR is much smaller than the
theoretical bound. For example, when $J=150$ and $\kappa=0.2$, the 95\% quantile of empirical FPR is $0.1000$, while the theoretical bound is $0.3509$. In addition, the
theoretical bound decreases as $J$ increases, reflecting that a larger
calibration set leads to tighter FPR control. Meanwhile, increasing
$\kappa$ raises the bound, which is consistent with the explicit trimming
cost characterized in Theorem~\ref{thm:certdw_fpr}.

\section{Preliminary Experiments in Larger-Scale Settings}
\label{sec:subimagenet_appendix}

To provide preliminary evidence of CertDW under larger-scale settings, we conduct additional experiments on (Sub-)ImageNet (a 20-class subset of ImageNet \cite{deng2009imagenet} with about 10,000 training and 1,000 testing images at $224\times 224$ resolution) using three deep architectures: ResNet-34, ResNet-50, and ResNet-101, all initialized with ImageNet pre-trained weights. We adopt the same BadNets-based watermarking as in our main experiments, with a randomly-positioned $5\times 5$ trigger, and all other settings (\eg, $J=100$, $\kappa=0.2$, $\alpha_0=0.05$, $M=1024$) follow those in Sections \ref{Exper_setup}\&\ref{sec:results}. As shown in Table~\ref{table:subimagenet}, the results provide preliminary evidence of the applicability of CertDW under this larger-scale setting across different deep architectures, with BA ranging from 80.90\% to 82.40\%, WSR exceeding 97\%, and VSR reaching 54\% on ResNet-34.

\section{The Relation to Conformal Prediction}
\label{sec:cp}

Conformal Prediction (CP)~\cite{vovk2005algorithmic,shafer2008tutorial,angelopoulos2021gentle} is a distribution-free framework for uncertainty quantification. It constructs quantile-based thresholds from non-conformity scores on a calibration set and provides finite-sample coverage guarantees under the exchangeability assumption. Due to these favorable properties, CP has been widely applied to classification~\cite{romano2020classification}, regression~\cite{lei2014distribution}, and several non-standard settings such as label noise~\cite{feldman2023conformal}, ambiguous ground truth~\cite{stutz2023conformal}, and distribution shift~\cite{tibshirani2019conformal}. More recently, CP has been further combined with randomized smoothing to provide certified robustness against sample-level perturbations. For example, RSCP~\cite{gendler2022adversarially} inflates the calibration threshold using a Lipschitz-bound-derived margin, and subsequent works~\cite{yan2024provably,jeary2024verifiably,jeary2026verifiably,zargarbashi2024robust} further improve the efficiency and applicability of robust CP.

Inspired by these works, we consider adapting such a ``calibration + smoothing'' paradigm from sample-level prediction to the DOV setting: on the one hand, we construct a data-driven calibration reference distribution from benign-model statistics and compare the suspicious model's verification statistic against this reference; on the other hand, we combine randomized smoothing to certify robustness against sample-level perturbations, thereby enabling robust ownership verification of the protected dataset. However, existing CP-based robustness methods mainly focus on coverage guarantees for individual test samples and cannot be directly transferred to DOV. Moreover, in classical CP, the validity of the coverage guarantee relies on the exchangeability between calibration and test points; this assumption is difficult to satisfy in the DOV setting, since the calibration statistics are computed from benign models, whereas the test statistic comes from the suspicious model. Therefore, adapting these two ingredients to certified dataset watermarking is non-trivial, and the unique challenges it poses go beyond a simple combination of existing CP and randomized-smoothing methods.

\section{Finite-M Certification and Validation}
\label{app:finite_mc}

The certification conditions in Theorem~\ref{thm:General_condition}
and Examples~\ref{example:GS}--\ref{example:US} are stated in terms of the true smoothing probability $W(f_{\vtheta},\pno)$. In practice, however, this quantity is estimated using a finite number $M$ of Monte Carlo samples. In this appendix, we explicitly account for the
resulting estimation error. We first derive simultaneous one-sided lower confidence bounds for the $K$ selected samples, then provide the corresponding finite-$M$ certification procedure, and finally evaluate its effects on WCA and the certified region.

\subsection{Simultaneous Lower Confidence Bound for WR}

For each selected watermarked sample $\hat{\bm{x}}_k=\bm{x}_k+\mathbf{r}_k$,
$k=1,\ldots,K$, let $q_k
\triangleq
p_{\hat{y}}(\hat{\bm{x}}_k\mid f_{\vtheta},\pno)
=
\mathbb{P}_{\vepsilon\sim\pno}
\left(
\arg\max f(\hat{\bm{x}}_k+\vepsilon;\vtheta)=\hat{y}
\right)$ denote its true target-label smoothing probability. According to
Definition~\ref{def:properties_cdw}, we have $W(f_{\vtheta},\pno)
=
\min_{k=1,\ldots,K}q_k$.
For each $\hat{\vx}_k$, we sample $M$ independent noises
$\{\vepsilon_i\}_{i=1}^{M}$ from $\pno$ and define $n_k=
\sum_{i=1}^{M}
\mathbb{I}
\left\{
\arg\max f(\hat{\vx}_k+\vepsilon_i;\vtheta)=\hat{y}
\right\}$. Then $n_k\sim\mathrm{Binomial}(M,q_k)$. For a prescribed failure
probability $\delta_M\in(0,1)$, we compute a one-sided
Clopper--Pearson lower confidence bound for each $q_k$ as

\begin{equation}
\label{eq:finiteM_pk_lower}
\underline{q}_k=
\begin{cases}
0, & n_k=0,\\[1mm]
F^{-1}_{\mathrm{Beta}(n_k,M-n_k+1)}
\left(\dfrac{\delta_M}{K}\right),
& n_k>0,
\end{cases}
\end{equation}
where $F^{-1}_{\mathrm{Beta}(a,b)}(\cdot)$ denotes the quantile
function of the $\mathrm{Beta}(a,b)$ distribution. Thus, the one-sided lower bound $\underline{q}_k$ has coverage
probability at least $1-\delta_M/K$. We further define the simultaneous lower confidence
bound for WR as
\begin{equation}
\label{eq:finiteM_W_lower}
\underline{W}
=
\min_{k=1,\ldots,K}\underline{q}_k.
\end{equation}

\begin{proposition}[Finite-$M$ Lower Confidence Bound for WR]
\label{pro:finiteM_WR}

For a fixed model and the $K$ selected samples, the lower bound
$\underline{W}$ defined in Eq.~(\ref{eq:finiteM_W_lower}) satisfies
\begin{equation}
\label{eq:finiteM_simultaneous}
\mathbb{P}
\left(
W(f_{\vtheta},\pno)\geq\underline{W}
\right)
\geq
1-\delta_M,
\end{equation}
where the probability is taken over the Monte Carlo noise samples.
\end{proposition}
\begin{proof}
For each $k$, the one-sided Clopper--Pearson lower confidence bound
in Eq.~(\ref{eq:finiteM_pk_lower}) satisfies $\mathbb{P}
\left(
q_k<\underline{q}_k
\right)
\leq
\frac{\delta_M}{K}$. Therefore, by the union bound,
\begin{equation}
\mathbb{P}
\left(
\exists k\in\{1,\ldots,K\}:
q_k<\underline{q}_k
\right)
\leq
\sum_{k=1}^{K}\frac{\delta_M}{K}
=
\delta_M.
\end{equation}

Hence, with probability at least $1-\delta_M$,
$q_k\geq\underline{q}_k$ holds simultaneously for all
$k=1,\ldots,K$. Taking the minimum over the $K$ samples gives
$W(f_{\vtheta},\pno)\geq\underline{W}$.

\end{proof}

\subsection{Finite-M Certification Condition}

We next connect $\underline{W}$ with the certification condition in
Theorem~\ref{thm:General_condition}. Since
$W(f_{\vtheta},\pno)\geq\underline{W}$ holds with probability at
least $1-\delta_M$, we have
$1-W(f_{\vtheta},\pno)\leq1-\underline{W}$.
Moreover, the optimal type-II error
$\beta_2^*(\alpha_1,H_1)$ is non-increasing with respect to $\alpha_1$.
Therefore,
\begin{equation}
\label{eq:finiteM_beta}
\beta_2^*
\left(
1-W(f_{\vtheta},\pno),H_1
\right)
\geq
\beta_2^*
\left(
1-\underline{W},H_1
\right).
\end{equation}

Consequently, a finite-$M$ sufficient condition for
Theorem~\ref{thm:General_condition} can be obtained by using
$\underline{W}$ in place of the unknown
$W(f_{\vtheta},\pno)$. Under Gaussian smoothing, the corresponding finite-$M$ condition in Example~\ref{example:GS} becomes
\begin{equation}
\label{eq:finiteM_gaussian}
\underline{W}
>
\Phi
\left(
\frac{R}{\sigma}
+
\Phi^{-1}
\left(
P_C^{(J-m-\lfloor\alpha_0(J-m+1)\rfloor)}
(g_{\vw},\pno)
\right)
\right).
\end{equation}

Similarly, for the one-dimensional uniform smoothing case in
Example~\ref{example:US}, the corresponding finite-$M$ condition is
\begin{equation}
\label{eq:finiteM_uniform}
\underline{W}
>
P_C^{(J-m-\lfloor\alpha_0(J-m+1)\rfloor)}
(g_{\vw},\pno)
+
\frac{R}{h-e}.
\end{equation}

Therefore, under the corresponding smoothing distribution, satisfying
Eq.~(\ref{eq:finiteM_gaussian}) or Eq.~(\ref{eq:finiteM_uniform})
ensures that the original certification condition holds with
probability at least $1-\delta_M$, where the probability is taken over
the Monte Carlo noise samples used to estimate WR.

The finite-$M$ certification can be directly incorporated into
Algorithm~\ref{alg:certdw-cert} with only a minor modification.
Specifically, for each selected watermarked sample, the $M$ Monte Carlo
samples in Step~5 are used to obtain the target-label count $n_k$ and
the corresponding one-sided Clopper--Pearson lower confidence bound
$\underline{q}_k$ according to Eq.~(\ref{eq:finiteM_pk_lower}).
The WR in Step~7 is then replaced by
$\underline{W}=\min_{k=1,\ldots,K}\underline{q}_k$, and Step~9 uses
the finite-$M$ certification condition in
Eq.~(\ref{eq:finiteM_gaussian}). All other steps remain unchanged.

\subsection{Empirical Validation}

We further evaluate the effect of the finite-$M$ lower confidence
bound from two aspects. First, we compare the WCA obtained using the
original Monte Carlo estimates with that obtained using
$\underline{W}$. Second, we examine the effect of different
confidence levels on the certified region.

\begin{table*}[t]
\centering
\caption{
Comparison of the empirical WCA and the finite-$M$ WCA for
CertDW on GTSRB and CIFAR-10 under different numbers of
Monte Carlo samples $M$. The finite-$M$ WCA is calculated using
the simultaneous lower confidence bound $\underline{W}$ with
$\delta_M=0.001$.
}
\label{tab:finiteM_wca}
\scalebox{0.88}{
\begin{tabular}{c c c cc cc cc}
\toprule
\multirow{2}{*}{Dataset} &
\multirow{2}{*}{Watermark} &
\multirow{2}{*}{$\sigma$} &
\multicolumn{2}{c}{$M=1024$} &
\multicolumn{2}{c}{$M=2048$} &
\multicolumn{2}{c}{$M=4096$} \\
\cmidrule(lr){4-5}
\cmidrule(lr){6-7}
\cmidrule(lr){8-9}
& & &
Emp. & Finite-$M$ &
Emp. & Finite-$M$ &
Emp. & Finite-$M$ \\
\midrule

\multirow{6}{*}{GTSRB}
& \multirow{3}{*}{BadNets}
& 1.5 & 28 & 20 & 26 & 18 & 24 & 22 \\
& & 2.5 & 48 & 34 & 50 & 36 & 46 & 34 \\
& & 3.5 & 54 & 42 & 56 & 42 & 52 & 50 \\
\cline{2-9}

& \multirow{3}{*}{Blended (patch)}
& 1.5 & 6 & 6 & 6 & 6 & 6 & 6 \\
& & 2.5 & 22 & 12 & 20 & 10 & 20 & 16 \\
& & 3.5 & 36 & 30 & 34 & 32 & 34 & 32 \\

\midrule

\multirow{6}{*}{CIFAR-10}
& \multirow{3}{*}{BadNets}
& 0.6 & 24 & 14 & 24 & 14 & 22 & 20 \\
& & 1.2 & 36 & 26 & 34 & 26 & 34 & 28 \\
& & 1.8 & 40 & 30 & 36 & 30 & 36 & 36 \\
\cline{2-9}

& \multirow{3}{*}{Blended (patch)}
& 0.6 & 20 & 8 & 22 & 14 & 20 & 18 \\
& & 1.2 & 32 & 20 & 26 & 20 & 26 & 22 \\
& & 1.8 & 32 & 22 & 34 & 26 & 30 & 26 \\

\bottomrule
\end{tabular}
}
\end{table*}

\vspace{0.3em}
\noindent \textbf{Effects on WCA.} 
We hereby compare the empirical WCA with the finite-$M$ WCA under different
numbers of Monte Carlo samples $M$ using representative settings from
Tables~\ref{table:GTSRB} and~\ref{table:CIFAR10}. The finite-$M$ WCA
is calculated using the simultaneous lower confidence bound
$\underline{W}$ and the certification condition in
Eq.~(\ref{eq:finiteM_gaussian}). Unless otherwise specified, we set
$\delta_M=0.001$, corresponding to a confidence level of
$99.9\%$.
As shown in Table~\ref{tab:finiteM_wca}, the gap between the empirical
WCA and the finite-$M$ WCA generally decreases as the number of Monte
Carlo samples $M$ increases, while the empirical WCA remains relatively
stable across different values of $M$. For example, under the BadNets
setting on GTSRB with $\sigma=3.5$, the gap decreases from 12 percentage
points at $M=1024$ to 2 percentage points at $M=4096$. Similarly, under
the Blended (patch) setting on CIFAR-10 with $\sigma=0.6$, the gap
decreases from 12 to 2 percentage points. Overall, the average gap
decreases from 9.50 percentage points at $M=1024$ to 7.83 at $M=2048$,
and further to 3.50 at $M=4096$. These results indicate that incorporating the finite-$M$ confidence bound leads to more conservative certification, while this additional conservativeness is substantially reduced as the number of Monte Carlo samples increases.

 \begin{figure*}[!t]
	\centering 
  \subfigure[]{
		\includegraphics[width=0.32\textwidth]{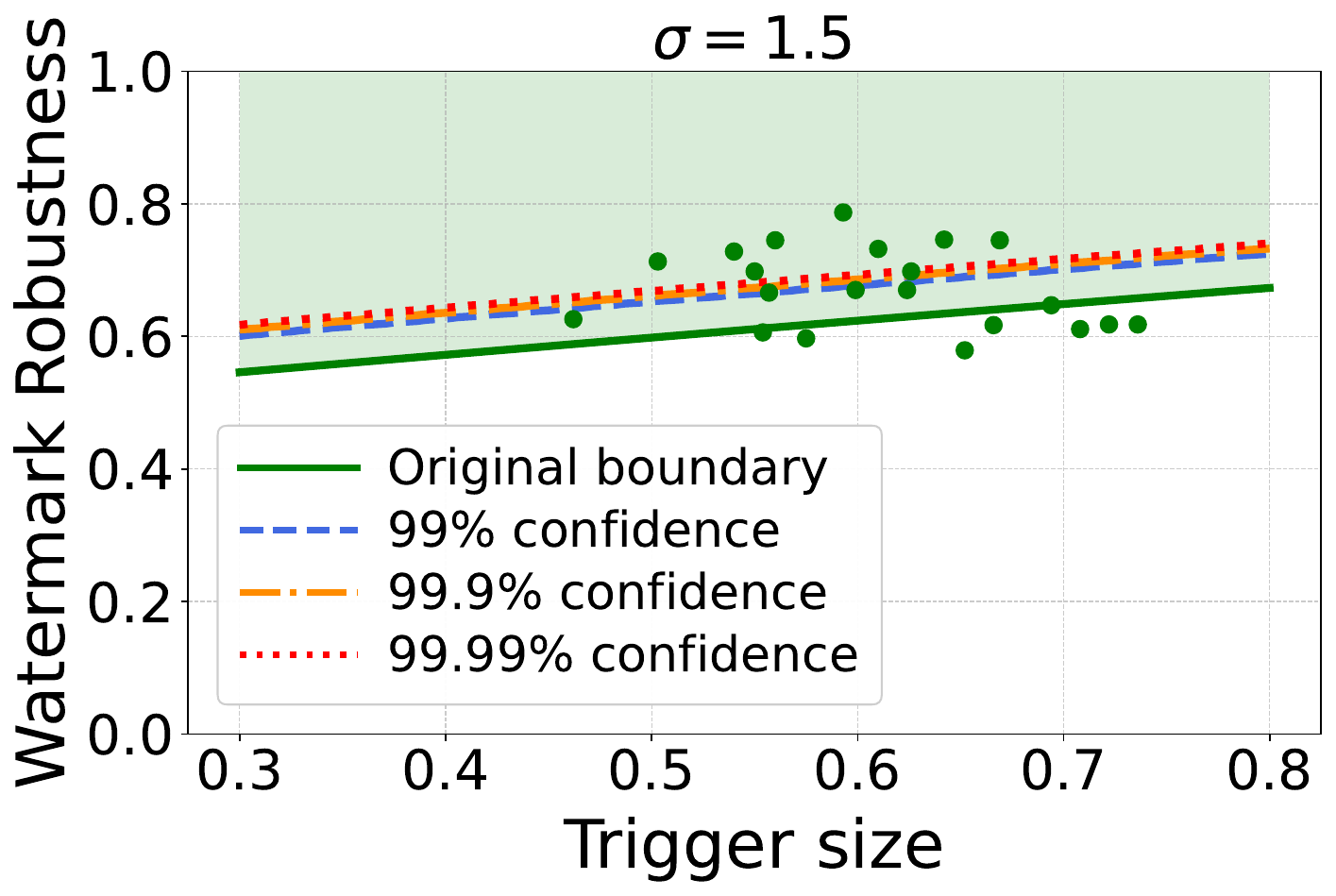}}
	\hspace{0.1em}
      \subfigure[]{
		\includegraphics[width=0.32\textwidth]{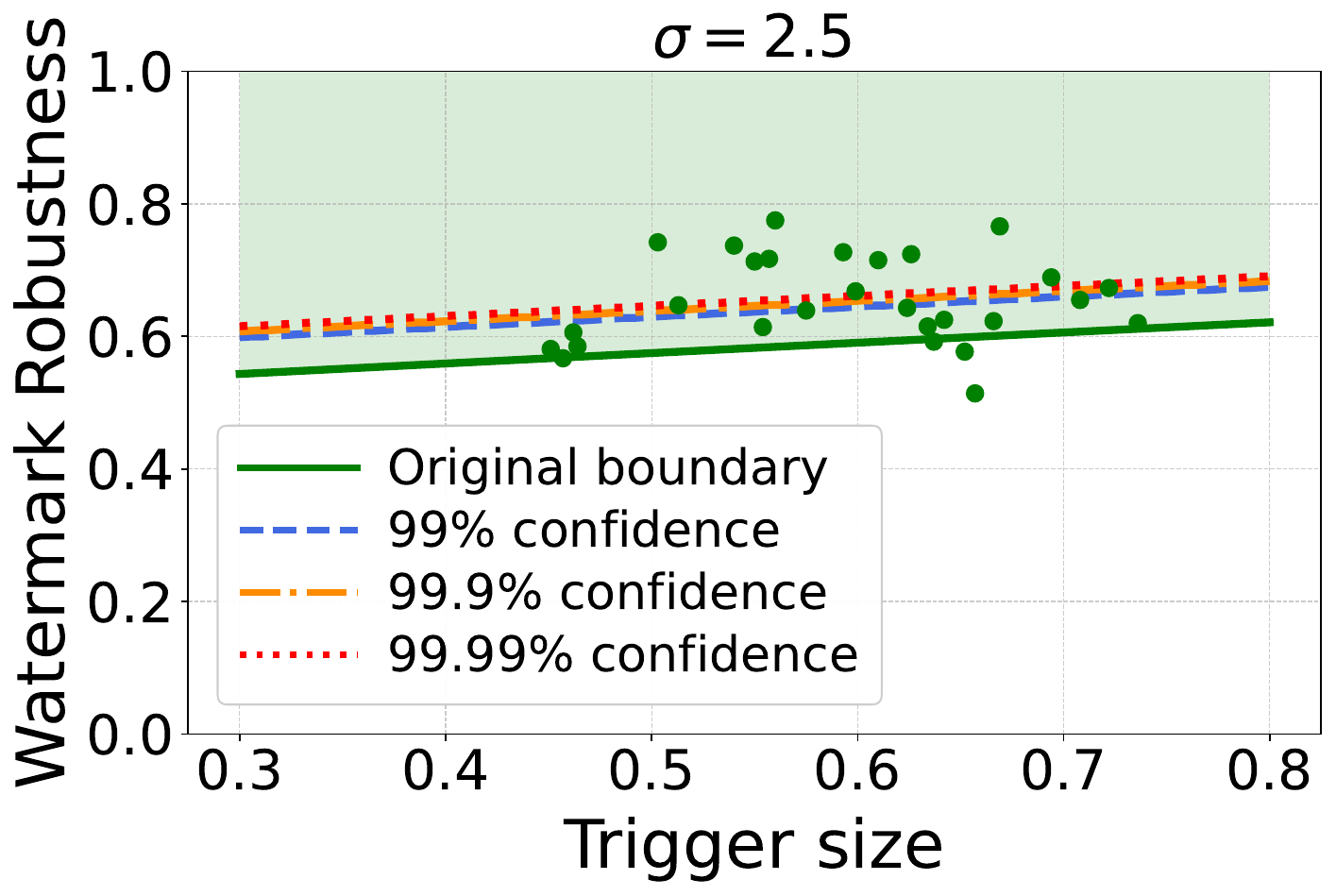}}
	\hspace{0.1em}
     \subfigure[]{
		\includegraphics[width=0.32\textwidth]{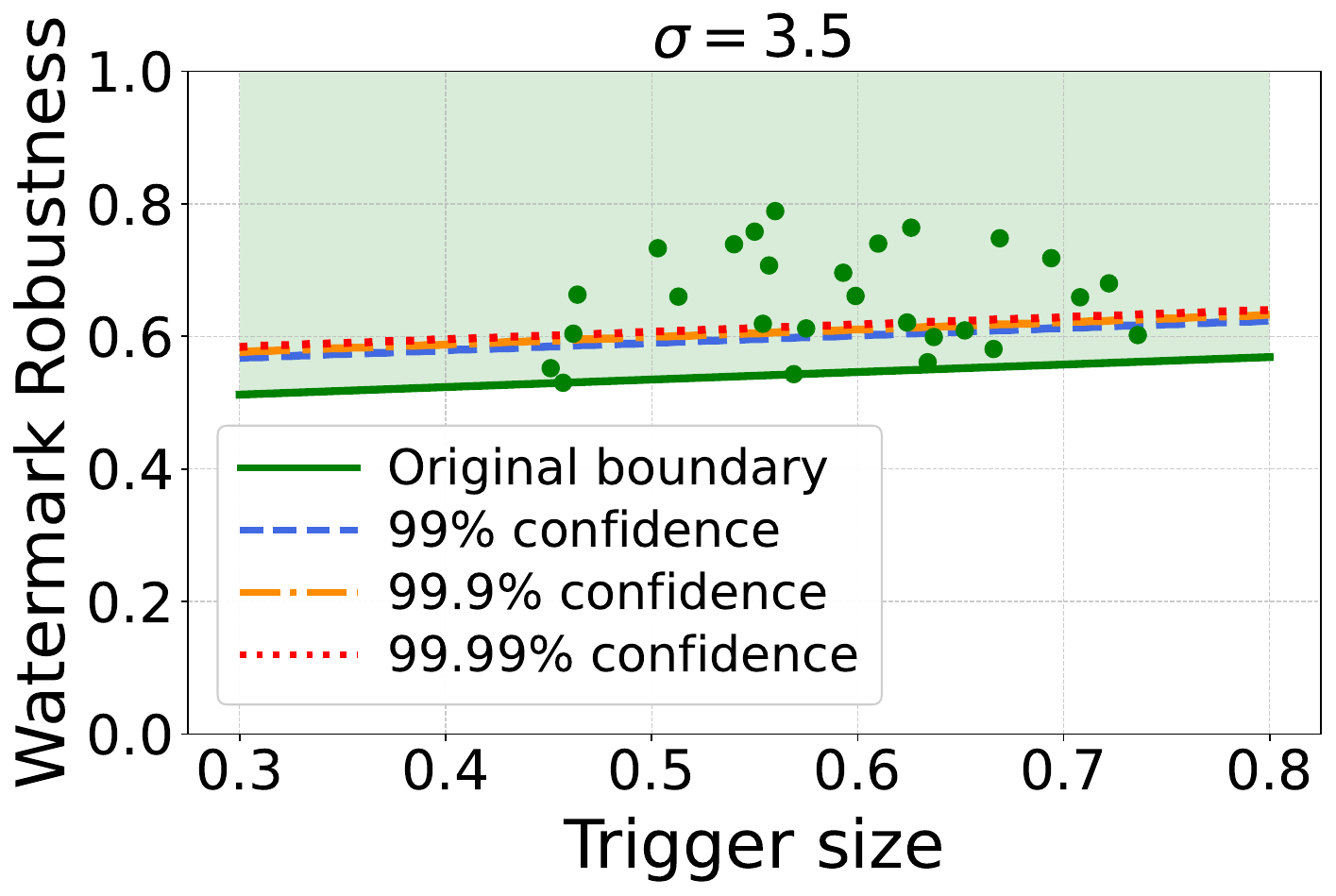}}
	\hspace{0.1em}
     \subfigure[]{
		\includegraphics[width=0.32\textwidth]{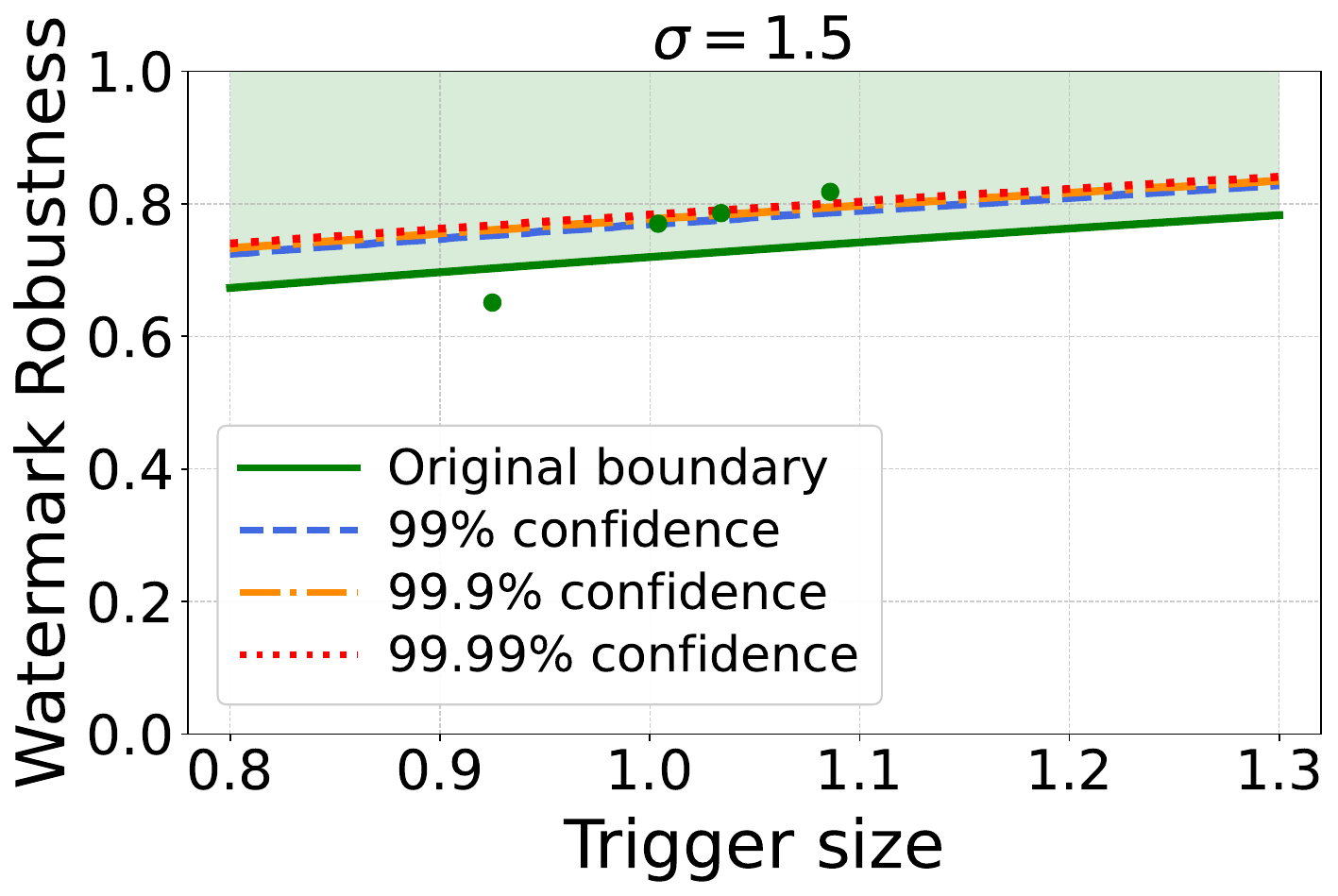}}
	\hspace{0.1em}
	 \subfigure[]{
		\includegraphics[width=0.32\textwidth]{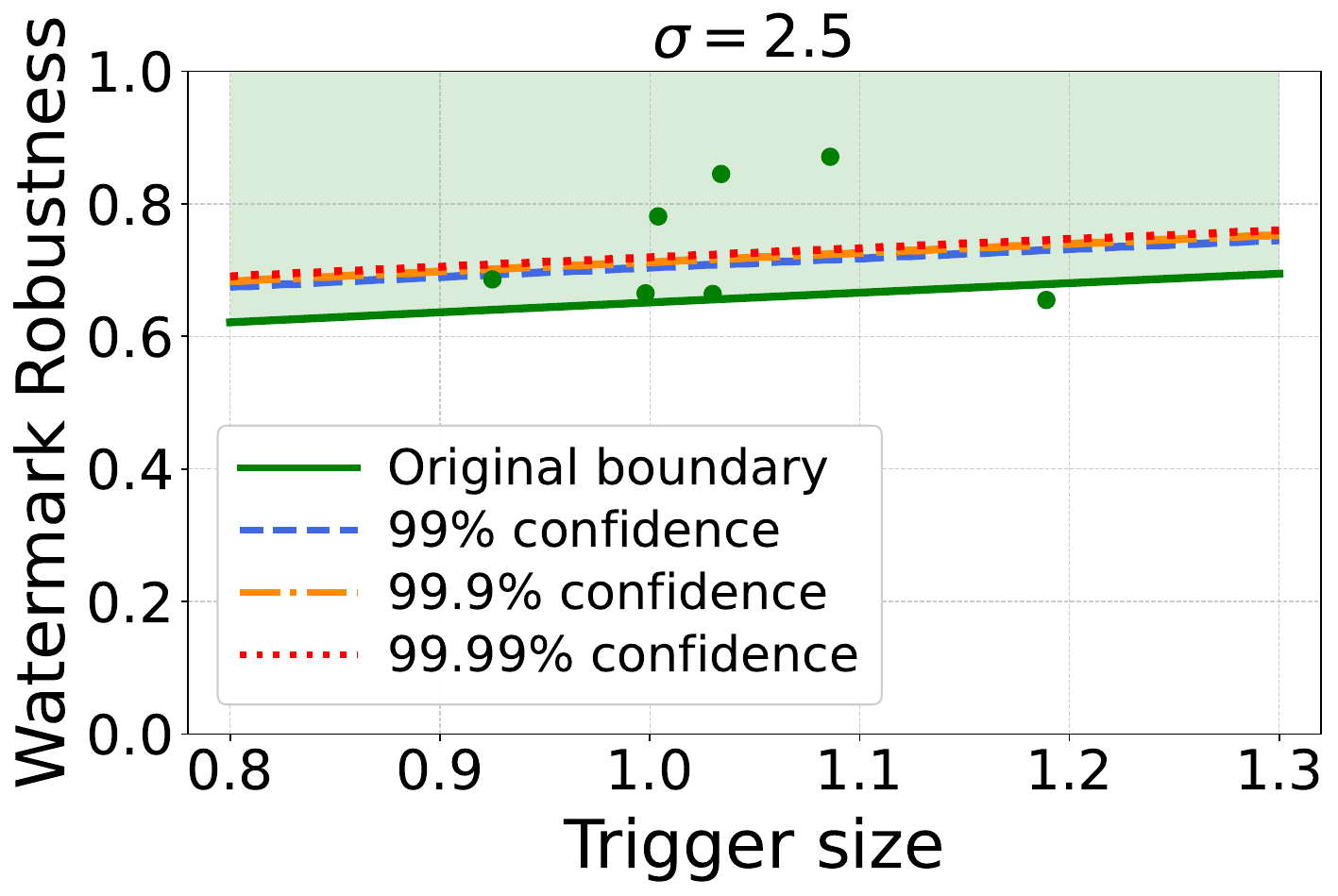}}
	\hspace{0.1em}
	 \subfigure[]{
		\includegraphics[width=0.32\textwidth]{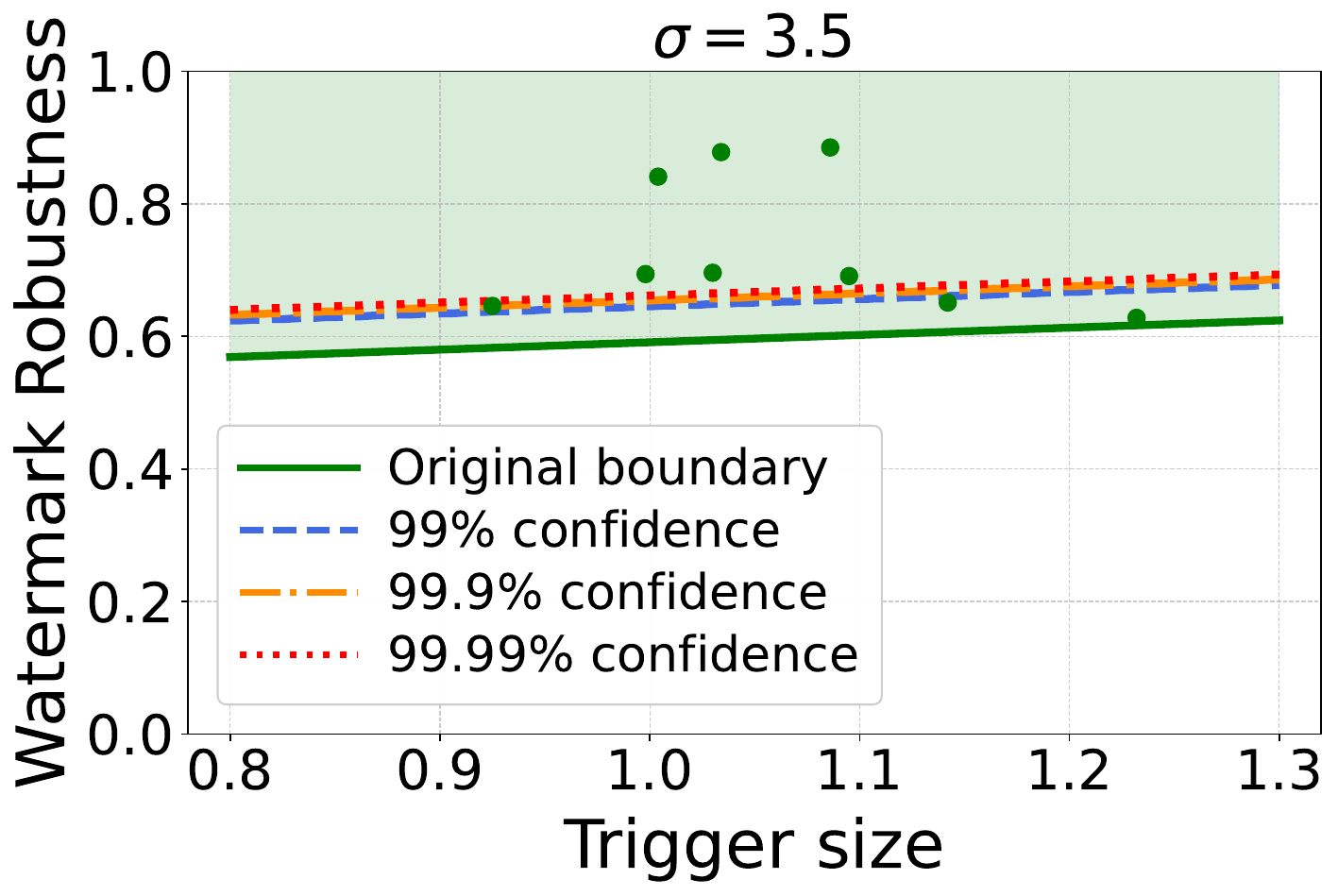}}
        \hspace{0.1em}
         \vspace{-1em}
	\caption{
Certified regions under different finite-$M$ confidence levels $1-\delta_M$. The horizontal axis represents the watermark perturbation magnitude $R$, and the vertical axis represents the empirical WR estimated using $M$ Monte Carlo samples. For each confidence level, the empirical WR is converted to the corresponding simultaneous lower confidence bound $\underline{W}$ before applying Eq.~(\ref{eq:finiteM_gaussian}).
}
\label{fig:finiteM_region}

\end{figure*}

\vspace{0.3em}
\noindent \textbf{Effects of the Confidence Level on the Certified Region.} 
We further analyze the effect of the confidence level on the certified
region. We fix $M=1024$ and consider
$1-\delta_M\in\{99\%,99.9\%,99.99\%\}$ while keeping the other
settings unchanged. For each confidence level, we compute the one-sided
lower confidence bound $\underline{q}_k$ for each selected sample according
to Eq.~(\ref{eq:finiteM_pk_lower}), obtain $\underline{W}$ using
Eq.~(\ref{eq:finiteM_W_lower}), and then apply the certification condition
in Eq.~(\ref{eq:finiteM_gaussian}). As shown in
Figure~\ref{fig:finiteM_region}, the finite-$M$ confidence boundaries
consistently lie above the original certification boundary, resulting in
smaller certified regions. Moreover, as the confidence level increases from
$99\%$ to $99.9\%$ and $99.99\%$, the corresponding boundary gradually
shifts upward and the certified region becomes more conservative.
Nevertheless, the three confidence boundaries remain relatively close
across all settings, indicating that increasing the confidence level
introduces only a moderate additional reduction in the certified region.

\section{Formal Definitions of Evaluation Metrics}
\label{app:evaluation_metrics}

We hereby provide the formal model-level definitions of VSR, FPR, and WCA used
in our experiments. Let $N_{\mathrm{wm}}$ and $N_{\mathrm{ind}}$ denote
the numbers of evaluated watermarked and independent models, respectively.
For an evaluated model $f$, let $D(f)\in\{0,1\}$ denote the ownership
verification decision, where $D(f)=1$ indicates that the model is verified
as having been trained on the protected dataset.

For the $N_{\mathrm{wm}}$ watermarked models, VSR is defined as
\begin{equation}
\mathrm{VSR}=\frac{1}{N_{\mathrm{wm}}}\sum_{i=1}^{N_{\mathrm{wm}}}D(f_i^{\mathrm{wm}})\times 100\%.
\end{equation}

For the $N_{\mathrm{ind}}$ independent models, we use their PP statistics
for a conservative empirical false-positive evaluation, as described in
Remark~\ref{remark:fpr_statistic} and Appendix~\ref{app:fpr_validation}.
Let $D(f_j^{\mathrm{ind}})=1$ indicate that the corresponding independent
model is incorrectly verified as having been trained on the protected
dataset. The FPR is then defined as:
\begin{equation}
\mathrm{FPR}=\frac{1}{N_{\mathrm{ind}}}\sum_{j=1}^{N_{\mathrm{ind}}}
D(f_j^{\mathrm{ind}})\times 100\%.
\end{equation}

For WCA, let $C(f_i^{\mathrm{wm}})\in\{0,1\}$ indicate whether the
watermarked model satisfies the certification condition. Under Gaussian
smoothing, this is equivalent to determining whether its $(R_i,W_i)$ pair
falls into the certified region:
\begin{equation}
C(f_i^{\mathrm{wm}})=\mathbb{I}\left\{W_i>\Phi\left(
\frac{R_i}{\sigma}+\Phi^{-1}(\tau)\right)\right\},
\end{equation}
where $W_i$ and $R_i$ denote the WR and watermark perturbation magnitude
of the $i$-th watermarked model, respectively, and $\tau$ denotes the
calibration threshold. Accordingly,
\begin{equation}
\mathrm{WCA}=\frac{1}{N_{\mathrm{wm}}}\sum_{i=1}^{N_{\mathrm{wm}}}C(f_i^{\mathrm{wm}})\times100\%.
\end{equation}

Thus, VSR and WCA respectively measure the proportions of successfully
verified and certified watermarked models, whereas FPR measures the
proportion of independent models incorrectly verified as having been trained on the protected dataset. The above definition corresponds to the empirical WCA reported in the main experiments. Its finite-$M$ counterpart replaces the empirical WR by the simultaneous lower confidence bound $\underline{W}$ and applies Eq.~(\ref{eq:finiteM_gaussian}), as detailed in Appendix~\ref{app:finite_mc}.

\end{document}